\documentclass{article}

\usepackage[final]{vscale}
\usepackage[utf8]{inputenc}
\usepackage[T1]{fontenc}
\usepackage{hyperref}
\usepackage{url}
\usepackage{booktabs}
\usepackage{colortbl}
\usepackage{amsfonts}
\usepackage{nicefrac}
\usepackage{microtype}
\usepackage{xcolor}
\usepackage{amsmath}
\usepackage{graphicx}
\usepackage{fontawesome5}
\usepackage{xcolor}
\usepackage{wrapfig}
\usepackage{placeins}
\usepackage{tabularx}
\usepackage{array}
\usepackage{multirow}
\usepackage{colortbl}
\usepackage{subcaption}
\usepackage{longtable}

\usetikzlibrary{fadings}
\IfFileExists{needspace.sty}{\usepackage{needspace}}{\providecommand{\Needspace}[1]{}}
\newcommand{\corrauthormark}{%
    \raisebox{0.15ex}{\textcolor{vscaleaccent}{\scriptsize\faEnvelope}}
}

\def\ie{\emph{i.e.}}

\newcommand{\name}{ZimaBlue}

\newcolumntype{Y}{>{\raggedright\arraybackslash}X}
\newcolumntype{C}{>{\centering\arraybackslash}p{0.12\linewidth}}

\definecolor{ZimaBlue}{HTML}{22B7F2}
\definecolor{rowblue}{HTML}{E3F2FD}

\title{\textit{\textbf{Zima\textcolor{ZimaBlue}{Blue}}}: Evolving Generalizable World Action Models through Scalable Video Pre-training}

\author{%
Joy Future Academy\textsuperscript{\hyperref[sec:authors]{1}}
}

\begin{document}

\maketitle

\begin{abstract}
Robotic manipulation faces a fundamental scaling challenge: robust generalization demands broad physical experience, yet action-labeled robot trajectories are expensive to collect and inherently limited in diversity. Egocentric videos offer a far more scalable source of embodied experience, capturing object interactions, contact dynamics, tool use, and long-horizon behaviors across diverse environments. The central challenge is how to convert this abundant but action-free experience into effective robot control. We introduce \textbf{\name{}}, a scalable framework for learning generalizable World Action Models (WAMs) from large-scale video. \name{} follows a three-stage training curriculum: it first performs causal embodied video pre-training on large-scale human and robot egocentric videos, then grounds the learned visual dynamics in heterogeneous robot trajectories through video-action mid-training with a unified action representation, and finally specializes the model to a target robot for deployment. To make generative WAMs practical for real-time control, \name{} further adopts an asynchronous Slow-Fast dual-system architecture, where a high-capacity Slow world model provides generalizable spatiotemporal representations and a lightweight Fast branch enables \textit{\textbf{30 Hz action prediction}} on NVIDIA RTX 4090. On real-robot zero-shot evaluations, scaling from target-robot data alone to over \textit{\textbf{120,000 hours of embodied video improves success from 36.1\% to 77.8\%}}. \name{} further delivers strong performance across multiple benchmarks, with particularly pronounced gains on unseen tasks.

\vspace{3pt}
{
  \small
  \fontfamily{SourceSansPro-TLF}\fontseries{sb}\selectfont
  \checkdata[Project]{\url{https://zimablue-wam.github.io/}}\par
  \checkdata[Code]{\url{https://github.com/ZimaBlue-WAM/ZimaBlue}}
}
\end{abstract}

\section{Introduction}

A central obstacle in general-purpose robotics is not the absence of capable policy architectures, but the lack of a scalable source of embodied experience. A useful robot must follow open-ended language instructions, and execute new skills under changes in scene layout, viewpoint, dynamics, and embodiment. Recent vision-language-action (VLA) models~\cite{gemini_robotics,openvla,gr00t,pi0,pi05,xiaomi_robotics_0,internvla_a1,yang2026datascalingrepresentationcentriccontinued} have made important progress on the first requirement. By extending pretrained vision-language models to motor control, they inherit strong semantic priors, can ground instructions in diverse objects and scenes, and usually support efficient feed-forward action prediction. Yet their generalization is still bounded by the action-labeled robot data used for policy learning. In practice, a VLA may understand what an instruction means while still lacking the spatial, dynamic, and motor knowledge required to perform a new manipulation skill~\cite{ye2026world}. 
This gap reflects a deeper bottleneck: reactive policies rely on action-supervised robot data not only to learn control, but also to acquire perceptual and physical knowledge that could otherwise be learned from much larger video corpora.

World Action Models (WAMs)~\cite{kim2026cosmos,ye2026world,lingbotva2026,zhang2026native,team2026motubrain,ye2026gigaworld,dyna2026dyna2} provide a different way to organize the learning problem, predicting how the world will evolve and how the robot should act within that evolving world. This joint video-action objective explicitly links visual understanding to the physical consequences of action. A misaligned grasp, a slipping object, or an occlusion during manipulation is not just visual variation; it is part of the causal structure that the model must explain. This is why recent WAMs such as DreamZero~\cite{ye2026world} and LingBot-VA~\cite{lingbotva2026} are appealing for robot learning: they make control depend less on memorizing labeled trajectories and more on understanding which visual changes are reachable. In this case, video becomes a scalable substrate for learning physical and causal priors before dense robot actions are available.

This distinction matters because the most scalable embodied data today is not robot trajectories, but video. Real-robot demonstrations remain expensive, even with teleoperation and UMI systems~\cite{pi05,chi2024universal,rayyan2026mv,zhaxizhuoma2025fastumi}; simulation provides useful coverage but suffers from a persistent sim-to-real gap and still misses much of the visual, material, and contact diversity of the real world~\cite{rt1,gr00t,tian2026interndata}. By contrast, first-person human videos are abundant and easy to expand~\cite{grauman2024ego,xperience_10m,buildaiegocentric100k2025,hoque2026egodex,zheng2026egoscale}. They contain rich evidence about object affordances, tool use, contact events, failure recovery, and long-horizon task structure, but most of them do not include action labels. This makes them hard to use in standard VLA training. A WAM can exploit them more naturally: it can first learn causal visual dynamics from unlabeled video, then align those dynamics with robot states and actions using a smaller amount of action-labeled data. We investigate this scaling strategy, with controlled comparisons across target-robot data, cross-embodiment robot data, and large-scale egocentric video pre-training. This separation is the key reason we choose the WAM paradigm for scaling embodied generalization.

However, simply initializing a WAM from an off-the-shelf video generator is not enough. Several recent video-action models~\cite{ye2026world,lingbotva2026} inherit powerful spatiotemporal priors from large video diffusion backbones~\cite{wan2025wan}, which were originally designed for generic video synthesis rather than robot control. This creates several mismatches. First, generic video models are usually optimized for visually plausible reconstruction under descriptive prompts, whereas a robot needs instruction-conditioned prediction that is action-relevant, and grounded in the consequences of intervention. Second, many video backbones either attend across an entire clip or generate long chunks at once, whereas a deployed controller receives observations sequentially and can condition only on what has already happened. Finally, web video contains animation, visual effects, edited transitions, and other non-physical content that may attenuate the priors needed for contact-rich manipulation. These limitations suggest that WAMs should be specialized through embodied video pre-training, rather than retrofitted from generic video generation alone.

We present \name{}, a WAM training framework built around this principle. The core hypothesis is simple: scaling causal embodied video pre-training improves downstream robotic generalization. \name{} follows a three-stage recipe. \textbf{\textit{First}}, we perform causal video pre-training on a large corpus of first-person and robot videos, scaling up to over 120,000 hours in our study. The model predicts future visual states from past observations and language instruction, encouraging it to learn semantic, temporal, and physical regularities before it sees target-robot actions. \textbf{\textit{Second}}, we use multi-embodiment robot trajectories to connect the learned video dynamics to executable control. To make data from different embodiments useful in a single model, we introduce a unified action representation that standardizes robot states and actions across heterogeneous platforms. This stage turns broad video priors into action-aware representations, preventing the model from overfitting to a single robot. \textbf{\textit{Third}}, we perform specific-embodiment post-training, which calibrates the model to the deployment embodiment and its action space. The resulting pipeline uses each data source for its natural role: abundant world knowledge from video, action grounding from cross-embodiment robot data, and precise control from target-robot demonstrations.

\begin{figure}[t]
  \centering
  \includegraphics[width=1.0\linewidth]{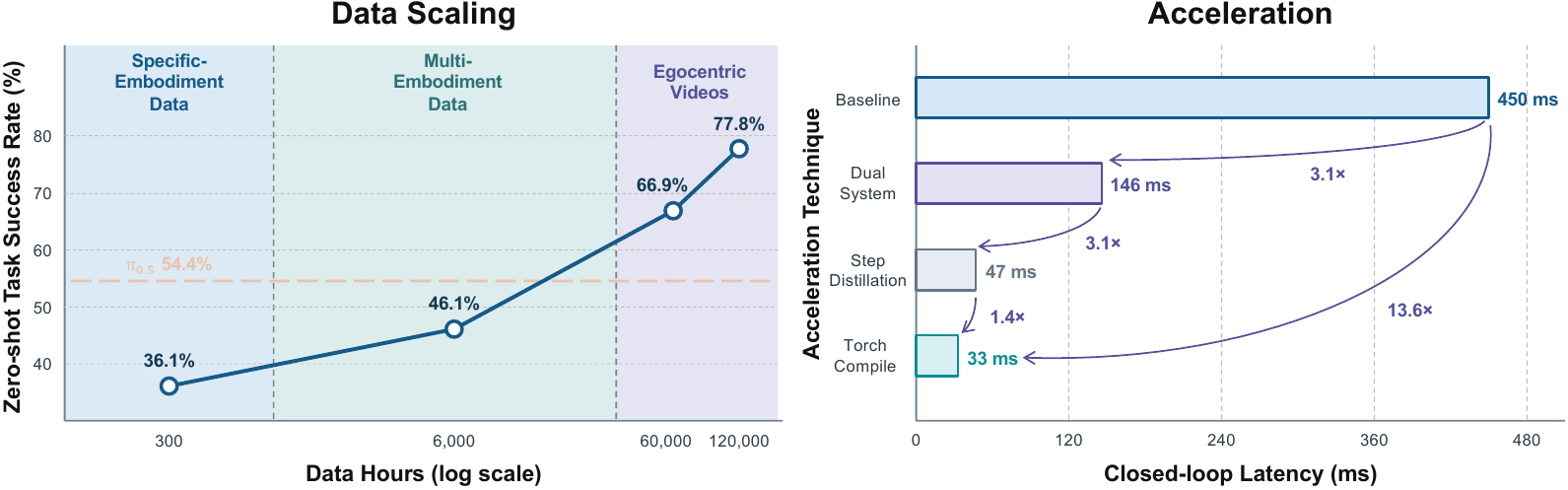}
  \caption{\textbf{Performance scaling with expanded data and acceleration.} \textbf{Left:} Zero-shot task success rate monotonically increases as the training data scales up from specific-embodiment data (36.1\% at 300 hours) to multi-embodiment datasets (46.1\% at 6,000 hours), and improves up to 77.8\% when further incorporating large-scale egocentric video data (120,000 hours). \textbf{Right:} Progressive acceleration techniques cut closed-loop latency from 450\,ms to 33\,ms on NVIDIA RTX 4090, achieving a 13.6$\times$ overall speedup.}
  \label{fig:teaser}
  \vspace{-0.05in}
\end{figure}

Our main evaluation asks whether this scaling path improves generalization, not merely video prediction quality or in-distribution imitation. We construct a suite of zero-shot manipulation tasks that are absent from the training data. As illustrated in Figure~\ref{fig:teaser}, the results show an appealing and consistent scaling trend. With only target-robot post-training, the model reaches 36.1\% success. Adding cross-embodiment video-action mid-training improves success to 46.1\%, showing that heterogeneous robot data provides transferable action grounding. Adding large-scale video pre-training further improves performance to 66.9\% with 60,000 hours of video and 77.8\% with 120,000 hours. These gains suggest that video pre-training plays an important role in improving generalization beyond the demonstrated tasks. In this sense, first-person video provides an effective and economical scaling axis for embodied intelligence.

We further evaluate \name{} on standard benchmarks. On LIBERO-Plus~\cite{fei2025libero}, we outperform all WAM-based methods. On RoboTwin 2.0~\cite{chen2025robotwin}, it achieves competitive results across diverse simulated manipulation tasks. On the challenging RoboCasa365~\cite{nasiriany2026robocasa365} benchmark, \name{} achieves the strongest overall performance among all compared methods except the VLA-based Xiaomi-Robotics-1~\cite{xiaomi_robotics_1}, which leverages 100,000 hours of real-robot data. The advantage of our model is especially pronounced on unseen tasks, where it substantially surpasses every other WAM-based baseline. 
This pattern is consistent with the central claim of the report: large-scale embodied video pre-training is most valuable when task and environment distributions shift.

A practical WAM must also be fast enough for closed-loop control. Compared with reactive VLAs, generative WAMs are typically slower because they predict future visual states and often rely on iterative denoising. This latency can create a reactivity gap on contact-rich tasks, where the robot must correct quickly after a failed grasp, or unexpected collision. \name{} addresses this with an asynchronous Slow-Fast control design. The Slow model is a larger WAM trained with more data and capacity, providing strong generalization and world-aware representations. A lightweight Fast module reuses shallow representations from the Slow model to infer actions at a higher control rate. This design preserves the generalization benefits of a large world model while reducing closed-loop latency. Together with diffusion-step distillation and CUDA graph optimization, the system reaches a 33\,ms control loop on NVIDIA RTX 4090 in Figure~\ref{fig:teaser}.

In summary, this work makes three contributions.
\begin{itemize}
    \item We present \name{}, framing video scaling as a practical route toward generalizable World Action Models, and provide empirical evidence that scaling video pre-training significantly improves zero-shot generalization and strengthens performance on challenging manipulation benchmarks.
    \item We introduce a three-stage training framework—comprising egocentric video pre-training, multi-embodiment video-action mid-training, and target-robot post-training—coupled with a unified state-action representation.
    \item We propose a Slow-Fast WAM architecture that enables 30\,Hz closed-loop physical robot control.
\end{itemize}

\section{Related Work}

\subsection{Vision Language Action Models}
Vision-Language-Action (VLA) models have become a central paradigm for generalist robot control. Early large-scale policies such as RT-1 and RT-2~\citep{rt1,rt2} showed that heterogeneous robot demonstrations can be absorbed into a unified language-conditioned policy. More recent systems, including OpenVLA, $\pi_0$, $\pi_{0.5}$, GR00T, and Gemini Robotics~\citep{openvla,pi0,pi05,gr00t,gemini_robotics}, extend this recipe with pretrained vision-language backbones, diffusion or flow-based action heads, and broader cross-embodiment robot data. By inheriting web-scale visual and semantic priors, these models achieve strong instruction following, object-level generalization, and transfer across diverse manipulation settings.

A complementary line of work uses foundation models as high-level planners rather than end-to-end visuomotor policies. These modular systems decompose robot behavior into semantic reasoning, affordance prediction, task planning, and low-level skill execution~\citep{saycan,palm_e,voxposer,progprompt}. Such decomposition improves interpretability and can reuse existing controllers, but it often depends on predefined skill libraries and carefully engineered interfaces between abstract reasoning and physical execution. End-to-end VLAs reduce this hand-engineered modularity by directly mapping observations and language instructions to actions, but they place a heavier burden on action-labeled robot demonstrations to teach spatial, temporal, and physical regularities.

Despite this progress, most VLA models remain primarily action-centric. Their pretrained backbones are typically optimized on static image-text or video-language objectives, while downstream control is learned through supervised imitation over comparatively limited robot trajectories. As a result, VLAs may inherit rich semantic knowledge about \emph{what} to do, but often lack an explicit model of \emph{how} the physical scene evolves under actions. This limitation becomes especially salient in long-horizon manipulation, novel physical skills, contact-rich interaction, and settings where historical context is required to disambiguate the current state. These observations motivate recent efforts to complement VLA-style semantic priors with predictive world-modeling objectives that directly capture temporal dynamics.

\subsection{World Action Models}
World models learn predictive representations of environment dynamics and have long been studied for model-based control and planning~\citep{planet,dreamer,dreamerv3}. In robotics, they differ mainly in the state representation used for prediction. Latent-space methods learn compact dynamics models for planning or reinforcement learning; 3D methods predict geometric evolution with point clouds or particles; and pixel- or video-space methods directly predict future visual observations~\citep{latent_world_model,pointworld,robocraft,unipi,robodreamer}. Video-based models are particularly attractive for generalist manipulation because they provide dense supervision from every frame transition and can exploit large-scale video pretraining to capture broad spatiotemporal priors.

Recent work has begun to integrate video prediction with robot policy learning. One family follows an \emph{imagine-then-act} paradigm: the model first generates future visual states and then recovers actions through inverse dynamics, planning, or a separate policy conditioned on the predicted future~\citep{unipi,robodreamer,video_prediction_policy}. While intuitive, this pipeline can suffer from open-loop drift, mismatch between generated and real observations, and additional latency from test-time video generation. A second family jointly models visual dynamics and actions within a single generative architecture, giving rise to World Action Models (WAMs)~\cite{kim2026cosmos,zhang2026native,team2026motubrain,ye2026gigaworld}. By coupling future-frame prediction with action generation, WAMs use visual dynamics as an auxiliary or implicit planning signal for visuomotor control.

Several recent WAMs instantiate this idea at scale. DreamZero~\citep{ye2026world} frames WAMs as zero-shot policies by adapting pretrained video diffusion models to jointly predict future videos and actions, showing that video-derived physical priors can improve generalization to unseen tasks and new embodiments. LingBot-VA~\citep{lingbotva2026} formulates robot control as autoregressive video-action diffusion, interleaving video and action tokens under causal attention and using KV-cache-based history to preserve long-horizon context. LingBot-VA~2.0~\citep{zhang2026native} further advocates native video-action pretraining for robot control, introducing an embodiment-oriented visual-action tokenizer, causal pretraining, and sparse expert capacity rather than simply adapting a generic video generator. Together, these works suggest that world modeling is not merely an auxiliary representation-learning objective, but can serve as an independent foundation for robot policy learning alongside vision-language pretraining.

However, existing WAMs still face important deployment challenges. First, high-quality video generation is computationally expensive, and naively placing video prediction on the closed-loop control path can reduce control frequency. Second, many joint world-action architectures bind video prediction and action execution to the same short temporal horizon, even though visual world prediction and low-level motor correction naturally operate at different timescales. These limitations motivate architectures that decouple long-horizon predictive reasoning from high-frequency reactive control while preserving information flow between the two.

\subsection{Asynchronous Dual-System Policies}
Dual-system architectures provide a natural way to reconcile deliberative reasoning with real-time control. In robotics, such systems typically combine a slow but expressive pathway for semantic understanding, planning, or predictive modeling with a fast pathway for reactive action execution. Prior work has explored this principle in several forms: hierarchical manipulation systems and skill-level diffusion planners decompose long-horizon tasks into executable subgoals or skills~\citep{yang2026hivla,liang2024skilldiffuser}; dual-policy systems combine a generalist model with a lightweight specialist controller~\citep{robodual,jiang2025galaxea,chen2025fast}; slow-fast sensorimotor policies use fast feedback pathways for contact-rich correction~\citep{reactive_diffusion_policy}; and asynchronous VLA or action-chunking systems move expensive model inference off the control-critical path~\citep{asyncvla,black2026real}. These approaches show that separating slow reasoning from fast execution can improve deployability, especially when the slow model is too expensive to run at the robot control rate.

Recent WAM-based policies extend this principle from semantic planning to predictive world modeling. AHA-WAM~\cite{ahawam2026} explicitly reorganizes WAM inference into a low-frequency video-DiT world planner and a high-frequency action-DiT executor. Its asynchronous horizon-adaptive design reuses long-horizon planner context across multiple action updates, while observation-guided context routing adapts stale planner context to the latest closed-loop observation. This formulation highlights a key insight: the world branch need not generate short-horizon plans at every control step; instead, it can act as a slower predictive substrate that provides reusable context for a faster action policy.

\section{Model Architecture}
\label{sec:arch}
\Needspace{12\baselineskip}
\subsection{Unified Representation}
\label{ssec:unified_representation}

Robot datasets expose incompatible control interfaces, ranging from Cartesian end-effector commands for a single arm to bi-manual systems with torso, base, and dexterous-hand controls. We map these native interfaces into a 100-dimensional semantic state-action space so that every coordinate retains consistent physical meaning across embodiments. State and action use the same slot layout: state describes the current robot configuration, whereas action represents a future control chunk.

\paragraph{Semantic slot layout.}
Each end-effector pose occupies 9 dimensions: 3D translation and a
continuous 6D rotation representation. The remaining slots cover grippers,
arm joints, torso, mobile base, and dexterous hands, as summarized in
Table~\ref{tab:unified_action_space}. Each embodiment activates only the slots
defined by its native interface. DROID~\cite{khazatsky2024droid}, for example, activates the left end-effector, left gripper, and seven left arm joint slots, yielding 17 valid coordinates. Bimanual embodiments activate the corresponding right-arm slots and, when available, torso, base, and hand slots.

\begin{table}[t]
  \centering
  \small
  \setlength{\tabcolsep}{8pt}
  \caption{\textbf{The semantic layout of the unified 100D state--action interface.}
  Individual embodiments activate only the slots that are physically defined
  by their native interface, while inactive coordinates are zero-filled and masked.}
  \begin{tabular}{lcl}
    \toprule
    \textbf{Slot} & \textbf{Dimension} & \textbf{Description} \\
    \midrule
    $[0,9)$   & 9  & Left end-effector position and rotation \\
    $[9,10)$  & 1  & Left gripper \\
    $[10,19)$ & 9  & Right end-effector position and rotation \\
    $[19,20)$ & 1  & Right gripper \\
    $[20,27)$ & 7  & Left arm joints \\
    $[27,34)$ & 7  & Right arm joints \\
    $[34,38)$ & 4  & Torso \\
    $[38,54)$ & 16 & Mobile base and auxiliary motion channels \\
    $[54,77)$ & 23 & Left hand joints \\
    $[77,100)$ & 23 & Right hand joints \\
    \bottomrule
  \end{tabular}
  \label{tab:unified_action_space}
\end{table}

\paragraph{Chunk-relative action geometry.}
Absolute coordinates are poorly aligned across robots and scenes. We therefore represent end-effector actions relative to the proprioceptive state at the first frame of each action
chunk, which serves as the reference anchor.
We denote the anchor pose as $(p_0,R_0)\in\mathbb{R}^3\times\mathrm{SO}(3)$, and the target end-effector pose as $(p_h,R_h)\in\mathbb{R}^3\times\mathrm{SO}(3)$ at horizon $h$. The relative pose is transformed as follows:
\begin{equation}
\Delta p_h = R_0^{-1}(p_h - p_0),
\qquad
\Delta R_h = R_0^{-1}R_h.
\label{eq:unified_relative_pose}
\end{equation}
Here, $\Delta p_h$ and $\Delta R_h$ denote the relative translation and rotation. We parameterize $\Delta R_h$ in a 6D rotation format (rot6d) using its first two columns. Following the same principle, joint targets are defined relative to the chunk anchor ($\Delta q_h = q_h - q_0$), where $q_0$ and $q_h$ correspond to the initial and target joint configurations.

Channel normalization is applied using dataset-wide statistics. Relative translation is normalized to $[-1,1]$ based on robust percentile bounds, while the 6D rotation representation remains unscaled. At deployment, an inverse transformation restores physical units before relative targets are composed with the current robot state.

\paragraph{Validity mask.}
Undefined slots are zero-filled and accompanied by state/action validity masks.
The masks exclude inactive coordinate supervision, as
detailed in the mid-training objective in
Section~\ref{sssec:training_video_action_midtrain_training_recipe}.
\subsection{Slow-Fast Dual-System}
\label{ssec:arch_dualsystem}

\begin{figure}[t]
  \centering
  \includegraphics[width=\linewidth]{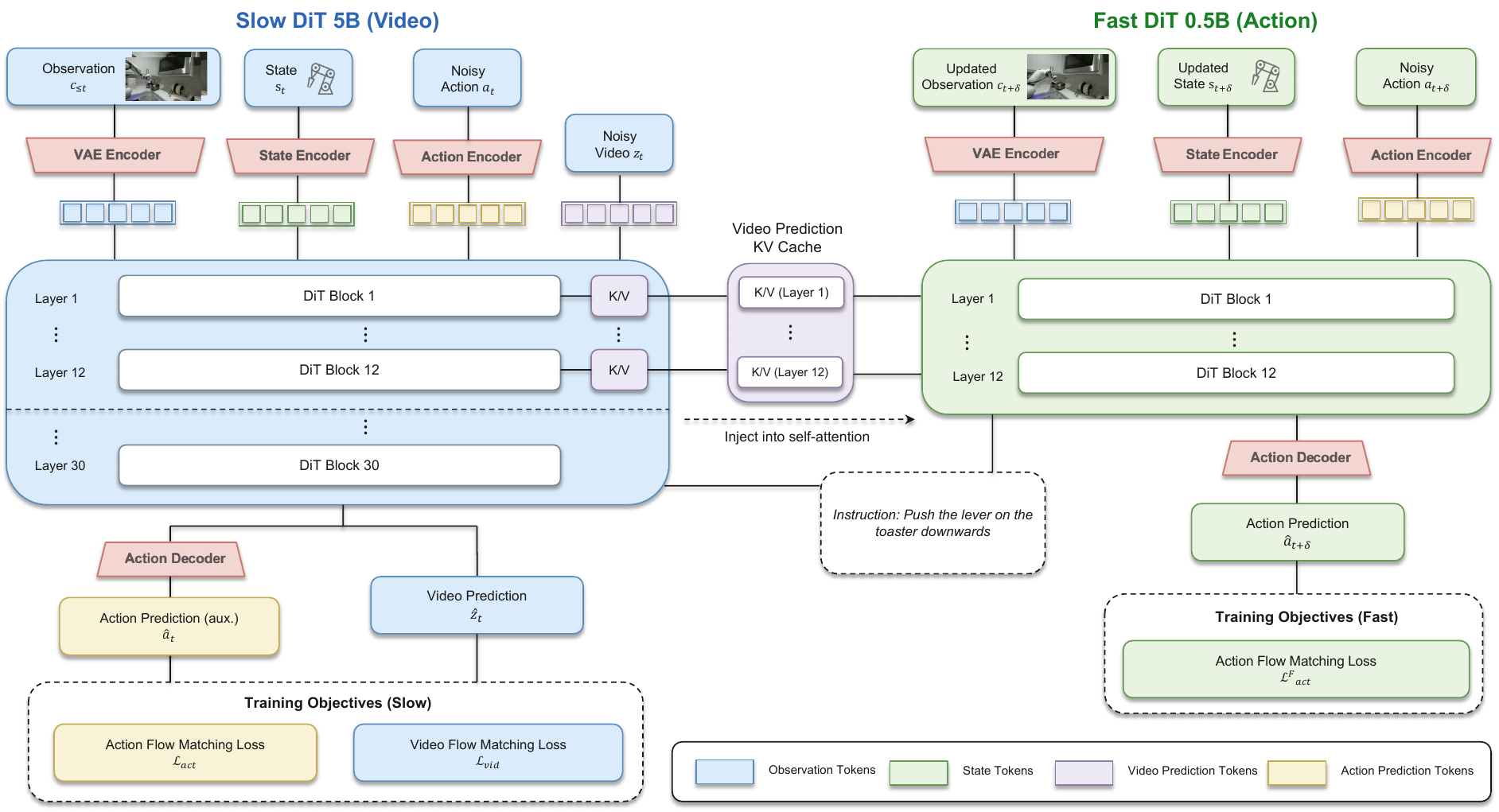}
  \caption{\textbf{Slow-Fast dual-system architecture.} The Slow DiT processes observations, robot state, language instruction, along with noisy video and action tokens, to jointly predict future video latents and their corresponding actions (serves only as auxiliary supervision for video-action alignment). Concurrently, the Fast DiT ingests the updated observation and state, conditioned on the Slow DiT's video K/V caches, to generate fine-grained final actions for high-frequency closed-loop control.}
  \label{fig:slow_fast_arch}
\end{figure}

Figure~\ref{fig:slow_fast_arch} shows the Slow-Fast architecture used for low-latency WAM control. The design separates high-capacity world modeling from high-frequency action generation. A Slow DiT (5B) maintains a strong video-centric world model and produces intermediate visual dynamics representations, while a lightweight Fast DiT (0.5B) consumes these representations together with the latest robot observation and state to predict the action used for control.

The Slow branch operates on the video-action interface introduced above. Given RGB observations, proprioceptive state $s_t$, and text instruction $\ell$, the system encodes them into visual latent tokens $z$, state tokens, and language embeddings, respectively. During training, the Slow branch additionally ingests noisy future video tokens alongside noisy action tokens. The primary objective of this branch is to model causal visual dynamics: it predicts future video latents while extracting multi-layer DiT key-value (K/V) features that summarize visual transitions. Notably, auxiliary action supervision is attached to the Slow branch not for runtime deployment, but rather to enforce strong semantic alignment between the predicted video dynamics and the underlying action sequences.

The Fast branch serves as the dedicated action generation module for real-time execution. Rather than executing a high-capacity world model at every control step, the Fast DiT reuses the layer-wise video K/V cache from the Slow branch as conditioning guidance. 
It leverages updated observation latents, current state tokens, language conditions, and noisy action tokens to predict the final action sequence. This design allows the controller to immediately react to the latest visual and proprioceptive feedback ahead of the Slow branch, while continuously benefiting from its rich spatial-temporal representation.

Mechanistically, the coupling between the two branches occurs directly within the self-attention layer. 
The Slow DiT computes self-attention over its unified video, action, and state tokens following the causal masking pattern detailed in Section~\ref{sssec:training_video_action_midtrain_training_recipe}, simultaneously caching the K/V features derived from its video stream at each layer. 
In the Fast DiT, action queries attend to their own observation, action, and state representations, while concurrently cross-attending to the cached Slow video tokens. As illustrated in Figure~\ref{fig:slow_fast_arch}, this K/V injection effectively bridges the two branches, transferring visual dynamics without the costly explicit generation of full future video sequences. The complete Slow and Fast attention masks are provided in Appendix~\ref{app:attention_details}.

By default, we instantiate a shallow Fast tower aligned with the early layers of the Slow DiT. Specifically, the Fast branch cross-attends to the video K/V caches from the first $12$ Slow layers, injecting local visual and motion priors while retaining the low latency required for high-frequency control.
During real-robot deployment (Section~\ref{ssec:real_robot_scaling}), the Fast branch directly generates the final control actions.

\section{Training Pipeline}
\subsection{Overview}
\label{ssec:training_overview}

\begin{figure}[t]
  \centering
  \includegraphics[width=0.95\linewidth]{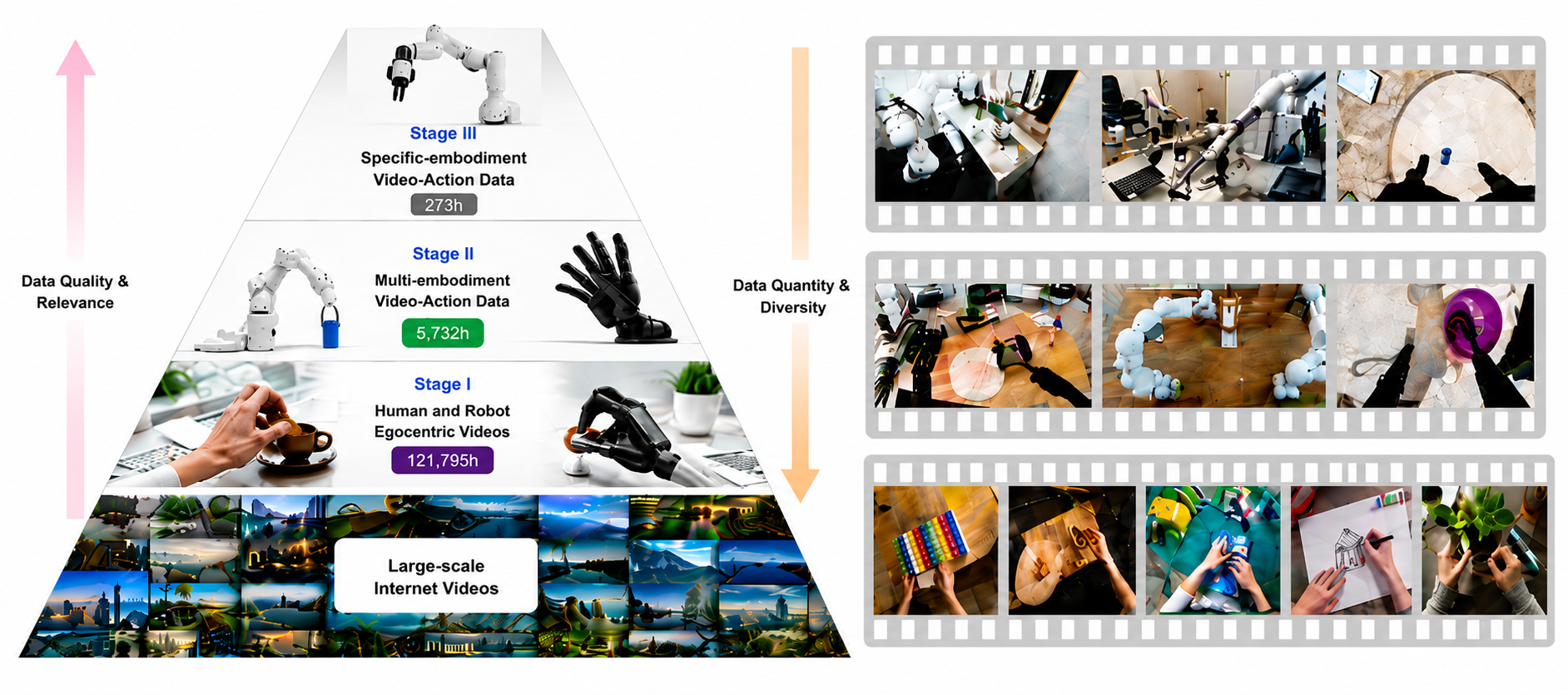}
  \caption{\textbf{Data pyramid.} Three-stage training progresses from broad visual diversity to deployment-specific embodiment. \textbf{Stage~\uppercase\expandafter{\romannumeral 1} (Pre-training)} learns general visual dynamics from diverse video sources without action supervision. \textbf{Stage~\uppercase\expandafter{\romannumeral 2} (Mid-training)} introduces robot actions from multiple embodiments, grounding visual dynamics in cross-embodiment control. \textbf{Stage~\uppercase\expandafter{\romannumeral 3} (Post-training)} specializes the model on target embodiments and benchmarks for deployment.}
  \label{fig:data_pyramid}
\end{figure}

\name{} is trained with a three-level data pyramid, as shown in Figure~\ref{fig:data_pyramid}, that progressively adapts a pre-trained text-to-video diffusion transformer to embodied intelligence. The curriculum separates two forms of supervision, \ie, video and action, which are typically entangled in robot learning. The proposed training paradigm has three stages. Specifically, Stage I, \emph{video pretraining}, uses large-scale heterogeneous videos to adapt the visual generative prior to causal embodied dynamics without requiring action annotations. Stage II, \emph{video-action mid-training}, introduces robot trajectories with synchronized observations, proprioceptive states, language instructions, and actions. Rather than treating action prediction as the final objective at this stage, action supervision grounds the Slow world model in control-relevant dynamics and encourages its visual representations to retain motor-relevant information through a unified cross-embodiment action interface. Stage III, \emph{post-training}, specializes the model to a target embodiment and introduces the lightweight Fast branch for responsive closed-loop control.

For training efficiency, we optimize only the Slow branch during video pretraining and video-action mid-training. Since these stages cover the largest and most heterogeneous data mixtures, we concentrate the training budget on learning a strong video-centric world model instead of jointly optimizing both towers throughout the full curriculum. By the start of post-training, the Slow representations have already been grounded by action supervision, substantially simplifying the mapping from visual dynamics to executable actions. We therefore introduce Fast only at this stage, where it can efficiently specialize to the target deployment domain on top of the aligned Slow representation.

Across all stages, we use a unified flow-matching objective. During video pretraining, it is applied only to visual latents, training the Slow branch to model causal forward dynamics. During video-action mid-training, the same objective is extended to action chunks, using action prediction as an auxiliary alignment signal that injects motor relevance into the shared visual representation. During post-training, Fast branch consumes these aligned Slow features together with the latest observation and proprioceptive state to produce low-latency actions for closed-loop control.

\subsection{Stage~\uppercase\expandafter{\romannumeral 1}: Video Pre-training}

Video pre-training repurposes generic video generation priors from Wan2.2-TI2V-5B~\cite{wan2025wan} into a causal visual dynamics model for embodied interaction. 
By learning from large-scale egocentric videos without action annotations, this stage acquires generalizable priors over objects, scenes, motion, and contact that provide a strong representational basis for subsequent action grounding. 
We elaborate on our recipe from two perspectives: data and training strategy.

\subsubsection{Data Recipe}
\label{sssec:training_video_pretrain_data_recipe}

The goal of video pre-training is to obtain an embodied video world model before introducing action supervision. 
We adopt a two-phase video-only curriculum. 
First, we train the model on the broadest mixture, covering egocentric human videos, simulated manipulation trajectories, and real robot demonstrations. 
The source corpus includes EPIC-KITCHENS~\citep{damen2020epic}, Egocentric-100K~\citep{buildaiegocentric100k2025}, EgoDex~\citep{hoque2026egodex}, HOT3D-Aria~\citep{banerjee2025hot3d}, DreamDojo~\citep{gao2026dreamdojo}, GenRobot~\citep{genrobot2025realomin}, RoboCOIN~\citep{wu2025robocoin}, DROID~\citep{khazatsky2024droid}, AgiBot~\citep{bu2025agibot}, Galaxea~\citep{jiang2025galaxea}, RoboMIND2~\citep{wu2024robomind}, InternData-A1~\citep{tian2026interndata}. 
Egocentric human sources contribute diverse first-person hand-object interaction patterns; simulated sources provide clean and controllable motion sequences; robot sources expose the model to manipulation scenes, camera viewpoints, and object dynamics closer to real-world deployment.

We subsequently continue video pre-training on a curated embodied subset. This dataset excludes noisy web-scale sources that often lack detailed task instructions or suffer from low visual resolution. 
Instead, we concentrate on manipulation-centric video sources, incorporating datasets such as DreamDojo, RoboCOIN, DROID, AgiBot, Galaxea, RoboMIND2, and InternData-A1, alongside a substantial volume of human egocentric proprietary video. This targeted domain adaptation shifts the video world model toward high-quality robotic viewpoints and refined instruction compliance, all while maintaining the core video-only objective.

The pre-processing retains the task-relevant RGB stream from raw sources, normalizes spatio-temporal scales, and segments long recordings into trainable episodes. Clips undergo random cropping, resizing, and photometric jitter prior to VAE encoding. Each sample comprises a fixed-length visual trajectory, language instructions when available, and null state/action placeholders. Specifically, the visual trajectory is formatted as $8K+1$ frames across $K$ temporal chunks (with $K=4$ yielding 33 frames in our default setup). The dummy action stream spans a 24-step horizon but carries no semantic supervision. This unified design establishes a standardized data interface across pre-training and downstream stages within the same causal prediction pipeline.

\subsubsection{Training Recipe}
\label{sssec:training_video_pretrain_training_recipe}

To support heterogeneous camera availability, video pre-training uses a unified three-view visual interface. Human egocentric clips typically provide only a primary view, whereas robot demonstrations often contain multiple synchronized views. We therefore represent every sample with a shared three-row vertical canvas: available views are placed in canonical rows, and absent views are padded. This preserves multi-view robot observations while allowing single-view human videos to share the same latent-video representation. A corresponding view-validity mask is propagated through the model. Invalid view regions are excluded from video-token attention and from the flow-matching loss, so padded cameras neither contribute supervision nor introduce spurious context.

For a clean future video latent block $z$ and Gaussian noise $\epsilon$, we form
\begin{equation}
z_t=(1-\sigma_t^{\mathrm{vid}})z+\sigma_t^{\mathrm{vid}}\epsilon,
\qquad
v^{\mathrm{vid}}=\epsilon-z.
\end{equation}
The video loss is computed over valid visual tokens,
\begin{equation}
\mathcal{L}_{\mathrm{vid}}
=
\mathbb{E}_{z,\epsilon,t}
\left[
\left\|m^{\mathrm{vid}} \odot \left(\hat{v}_{\theta}^{\mathrm{vid}}(z_t, c_{\le t}, \ell) - v^{\mathrm{vid}}\right)\right\|_2^2
\right],
\end{equation}
where $c_{\le t}$ denotes the clean causal video context, $\ell$ denotes the language condition when available, and $m^{\mathrm{vid}}$ combines temporal validity with the view-validity mask introduced by the unified three-view interface. Action and state streams are instantiated only as null placeholders: action supervision $\mathcal{L}_{\mathrm{act}}$ of Slow branch is disabled, and the Fast branch is not involved.
To support prediction from variable-length clean contexts, video latents are partitioned into temporal blocks, each with clean and noisy counterparts. A noise level is sampled independently for each block and shared across all tokens within the block to maintain temporal coherence. 
During training, clean video blocks provide causal context, while their noisy counterparts serve as denoising targets. A block-causal teacher-forcing attention scheme preserves the same temporal dependency structure as autoregressive rollout, while allowing multiple future video blocks to be denoised in parallel. This enables efficient multi-block training without sacrificing the causal information flow required for closed-loop prediction (see Appendix~\ref{app:slow_attention}).

\subsection{Stage~\uppercase\expandafter{\romannumeral 2}: Video-Action Mid-training}

Building upon the pre-trained backbone, mid-training incorporates real-robot trajectories using the action encoders and decoders detailed in Section~\ref{sec:arch}. Here, the Slow world model jointly predicts future video latents and action chunks.
This joint prediction forces action and video tokens to interact within the shared backbone, grounding pre-trained visual dynamics with cross-embodiment action semantics and producing representations rich in motor-relevant information. Note that during this stage, the Fast branch remains inactive.

\subsubsection{Data Recipe}
\label{sssec:training_video_action_midtrain_data_recipe}

Mid-training introduces a unified cross-embodiment trajectory supervision. Without loss of generality, we select four representative embodiment families—namely DROID~\cite{khazatsky2024droid}, AgiBot~\cite{bu2025agibot}, Galaxea~\cite{jiang2025galaxea}, and RoboMIND2-Franka~\cite{wu2024robomind}—covering single- and dual-arm manipulation. While each embodiment retains its heterogeneous sensorimotor modalities, all data are projected into a standardized training interface. This design preserves embodiment-specific dynamics while framing the learning process as a unified video-action prediction problem.

Each training sample is anchored at a key timestep and incorporates four synchronized modalities: multi-view video context, proprioceptive state, a future action chunk, and language instructions. To ensure precise cross-modal alignment, the visual context and the 24-step target action chunk cover the exact same temporal horizon starting from the anchor timestep, with video frames sampled at evenly spaced offsets. Language supervision is extracted directly from original annotations, including high-level task instructions and camera view descriptions. 
Crucially, temporal sampling windows are strictly confined within language-consistent segments to prevent merging disjointed subgoals into a single sample.

Each embodiment is mapped onto the shared three-row vertical canvas according to its sensor configuration: DROID uses two exterior cameras and a wrist camera; AgiBot uses head and left/right hand cameras; Galaxea uses head and left/right wrist cameras; RoboMIND2-Franka uses top and left/right wrist cameras. 
A shared visual normalization pipeline, including crop, resize, photometric augmentation, and canvas packing, standardizes the input geometry.

As described in Section~\ref{ssec:unified_representation}, native proprioceptive states and actions are unified into a 100-dimensional space. Specific embodiment families populate corresponding sub-vectors within this representation: DROID occupies a single-arm subset; AgiBot and Galaxea populate bimanual arm, hand/gripper, torso, and mobile-base slots; and RoboMIND2-Franka occupies dual-Franka end-effector, gripper, and joint slots. Each state and action tensor is accompanied by a binary validity mask to handle inactive dimensions.

\subsubsection{Training Recipe}
\label{sssec:training_video_action_midtrain_training_recipe}

Mid-training optimizes a unified video-action flow-matching objective,
\begin{equation}
\mathcal{L}^S_{sup}
=
\lambda_{\mathrm{vid}}\mathcal{L}_{\mathrm{vid}}
+
\lambda_{\mathrm{act}}\mathcal{L}_{\mathrm{act}},
\end{equation}
where $\mathcal{L}_{\mathrm{vid}}$ is the future video loss and $\mathcal{L}_{\mathrm{act}}$ is the corresponding action loss. For a clean action chunk $a$ and Gaussian noise $\eta$, the corrupted action is $a_t=(1-\sigma_t^{act})a+\sigma_t^{act}\eta$, and the target is $v^{\mathrm{act}} = \eta-a$. The video loss is reframed as
\begin{equation}
\mathcal{L}_{\mathrm{vid}}
=
\mathbb{E}_{z,\epsilon,t}
\left[
\left\|m^{\mathrm{vid}} \odot \left(\hat{v}_{\theta}^{\mathrm{vid}}(z_t, a_t, c_{\le t}, s_t, \ell) - v^{\mathrm{vid}}\right)\right\|_2^2
\right],
\end{equation}
and the action loss is calculated as
\begin{equation}
\mathcal{L}_{\mathrm{act}}
=
\mathbb{E}_{a,\eta,t}
\left[
\left\|m^{act} \odot \left(\hat{v}_{\theta}^{\mathrm{act}}(a_t, z_t, c_{\leq t}, s_t, \ell)- v^{\mathrm{act}}\right)\right\|_2^2
\right].
\end{equation}
Here $m^{act}$ is the action validity mask, and $s_t$ is the proprioceptive state. The mask restricts supervision to physically meaningful dimensions for active coordinates. Invalid slots are zeroed before encoding and are excluded from the loss.
We couple the video and action noise levels within a denoising step so that uncertainty in the visual rollout is aligned with uncertainty in the motor command that realizes it.

The block-causal attention pattern coordinates video, action, and state streams, with state tokens serving purely as conditioning inputs. Video tokens comprise clean history context, noisy future tokens, and clean future targets, whereas action tokens consist of noisy future tokens and clean future targets. Crucially, noisy future video and action tokens mutually attend to each other while jointly conditioning on the instruction, state tokens, and clean video history.

This dependency is pivotal for world-action modeling: the video stream generates plausible future observations conditioned on predicted actions, while the action stream infers the control sequences required to drive these visual transitions. During policy inference, the clean target video and action tokens are omitted, enabling the model to progressively denoise their noisy counterparts.

\subsection{Stage~\uppercase\expandafter{\romannumeral 3}: Video-Action Post-training}
\label{ssec:training_posttraining}

Post-training specializes the model to a target embodiment by adapting it to the deployment environment, control interface, and task distribution, while preserving the visual dynamics priors acquired during pretraining and the cross-embodiment grounding learned during mid-training. It proceeds in two phases: first, specializing the Slow branch to the target domain, and then training the Fast action branch with Slow frozen.

\subsubsection{Target-Domain Slow Branch Specialization}

We evaluate our model across two deployment settings: real-robot manipulation on DROID~\citep{khazatsky2024droid} and three simulation benchmarks—LIBERO-Plus~\citep{fei2025libero}, RoboTwin-2.0~\citep{chen2025robotwin}, and RoboCasa365~\citep{nasiriany2026robocasa365}. All DiT backbone parameters are initialized from the mid-trained checkpoint to retain pre-trained visual dynamics and cross-embodiment grounding. 
As DROID is present in previous stages, we directly preserve its unified $100$-dimensional state and action interface. 
For the unseen simulation benchmarks, we attach lightweight, embodiment-specific encoders and decoders (see Section~\ref{sec:arch}) to match their native control spaces. 
Empirically, such an interface-level adaptation converges rapidly without compromising backbone vision-action transfer. 

\subsubsection{Fast Branch Training}

After specializing the Slow branch, we freeze its parameters and train Fast using an action-only flow-matching objective. 
Fast is initialized from the first 12 layers of the Slow DiT, with all Fast parameters subsequently optimized.
Because Fast uses a narrower hidden dimension, we interpolate the transferred weights along the channel dimensions to match its model width.
To reproduce the information pattern of asynchronous closed-loop execution, for an action chunk of horizon $H$, we uniformly sample a time offset $\delta\in\{0,\ldots,H-1\}$. Fast receives the updated observation and state $(c_\delta,s_\delta)$ and predicts the shifted action chunk $(a_\delta,\ldots,a_{H-1})$; positions beyond the trajectory are padded and excluded from the loss. Thus, a single Slow prediction can be reused across observations arriving throughout its execution, allowing Fast to continuously close the control loop without waiting for the next Slow rollout.

We denote this shifted action chunk as $\tilde a=\operatorname{Pad}(a_{\delta:H})$. To simulate asynchronous replanning while preserving action continuity, we sample a prefix length $p$ and teacher-force the first $p$ actions, treating them as the action prefix already committed by the previous Fast prediction. 
For action index $i$, the input to Fast is
\begin{equation}
a_{t,i}^F
=
\begin{cases}
\tilde a_i, & i<p,\\
(1-\sigma_t^{act})\tilde a_i+\sigma_t^{act}\eta_i, & i\ge p.
\end{cases}
\end{equation}
The corresponding target velocity is $v^{act,F}=\eta-\tilde a$. Thus, the prefix remains clean while only the suffix is noised and predicted. Let $m^{act,F}$ combine the action validity mask $m^{act}$ with additional offset-padding and prefix masks, the action loss is defined as
\begin{equation}
\mathcal{L}^F_{sup}
=
\mathbb{E}_{a,\eta,t,\delta,p}
\left[
\left\|m^{act,F} \odot
\left(
\hat{v}_{\theta}^{\mathrm{act},F}
\left(a_t^F,c_\delta,s_\delta,\mathcal{K}_{\mathrm{slow}},\ell,\delta\right)
-
v^{act,F}
\right)\right\|_2^2
\right].
\end{equation}
Here $\mathcal{K}_{\mathrm{slow}}$ is the frozen Slow visual guidance and $\ell$ is the language condition. The corresponding Fast attention pattern is illustrated in Appendix~\ref{app:fast_attention}. The offset exposes Fast to fresh observations after partial execution, while the clean, loss-masked prefix preserves continuity with committed actions. Together they realize training-time Real-Time Chunking (RTC)~\cite{black2026real} for fast asynchronous control.

\section{Acceleration Schemes}
\label{sec:acc}
\subsection{Overview}
\name{} accelerates closed-loop control through three complementary mechanisms: asynchronous Slow-Fast inference, diffusion step distillation, and Torch compile. At deployment, the Slow and Fast branches operate at different frequencies: Slow updates long-horizon world-model guidance asynchronously, while Fast uses the latest observation and available Slow guidance to generate high-frequency actions without waiting for each Slow rollout. Following post-training, Distribution Matching Distillation (DMD)~\cite{Yin_2024_CVPR} is applied sequentially to the Slow branch's joint video-action generation and the Fast branch's action generation, reducing each branch from eight to two DiT evaluations.
\subsection{Asynchronous Inference in Dual-System}

\begin{figure}[t]
  \centering
  \includegraphics[width=0.95\linewidth]{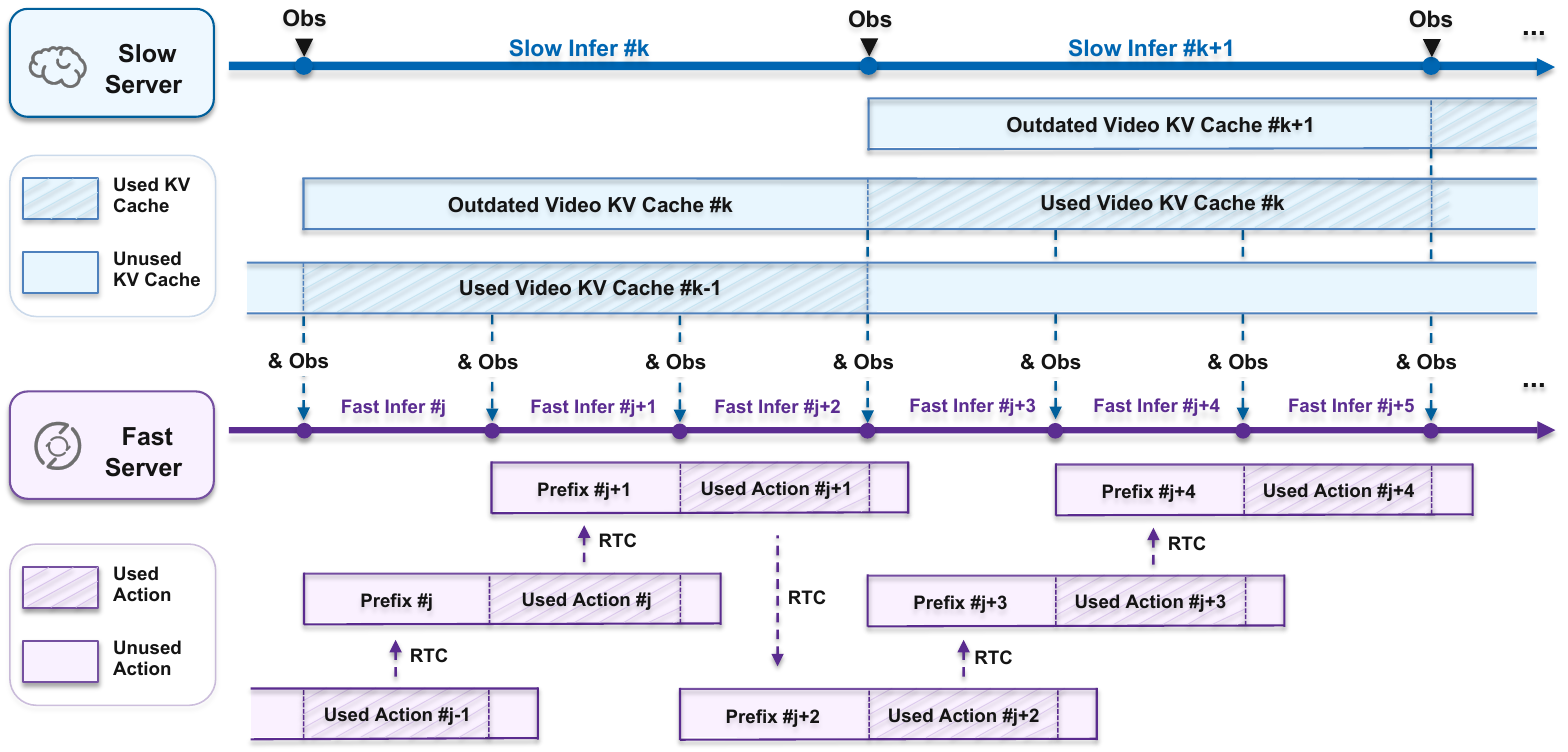}
  \caption{\textbf{Asynchronous inference in the Slow-Fast dual system.} The Slow stream operates at a lower frequency and produces future video K/V caches. The Fast stream runs at a higher frequency, using updated observations, states, and Slow K/V guidance to generate action chunks. The generated actions are continuously executed and provide new observations for subsequent updates. ``Outdated'' marks a Slow video K/V cache based on past observations that is stale at the current timestep, while ``prefix'' denotes actions already committed for execution from the previous chunk and retained for continuity. The figure is purely schematic.}
  \label{fig:async_infer}
\end{figure}

Figure~\ref{fig:async_infer} depicts the asynchronous inference schedule of the Slow-Fast dual system. The design decouples two processes with different temporal requirements: Slow updates long-horizon guidance, while Fast performs high-frequency closed-loop action generation. 
Instead of waiting for Slow inference to complete, the two streams run concurrently, allowing Fast to continuously refine actions using the latest observations while asynchronously receiving updated Slow guidance.

\paragraph{Slow guidance generation.}
The Slow stream serves as a low-frequency world-model predictor. Given the observation latent, proprioceptive state, and language condition, it performs a long-horizon rollout and exports the layer-wise video K/V cache. This cache encodes the predicted visual evolution and is reused by multiple Fast requests as temporal guidance. Once a new Slow prediction becomes available, Fast accesses the updated cache without interrupting the ongoing control loop.

\paragraph{Fast closed-loop refinement.}
The Fast stream performs high-frequency action generation using the latest observation, state, Slow video K/V cache, and noisy action chunk. It predicts an action chunk of horizon $H$, which is decoded and sent to the robot execution interface. 
During asynchronous updates, only future unexecuted actions are replaced by newly predicted actions, while the already committed actions remain unchanged.

\paragraph{Temporal consistency via RTC.}
To improve continuity across consecutive Fast predictions, we adopt an RTC inference mechanism. Each Fast request receives a short prefix from the previous action sequence and predicts the remaining suffix conditioned on the latest observation, state, and Slow guidance. The prefix anchors the transition between consecutive action chunks, while the suffix enables online correction under changing environments.

\subsection{Step Distillation}
\label{sec:step_distillation}

Following target-domain post-training, we apply DMD~\cite{Yin_2024_CVPR} as a sequential two-stage distillation procedure. 
We first distill the task-specialized Slow branch from eight to two DiT evaluations per Slow rollout. 
The resulting distilled Slow branch retains the joint prediction of future video latents and action tokens. 
We then freeze the distilled Slow branch and distill the Fast action branch from eight to two DiT evaluations per Fast request, conditioned on the layer-wise video K/V guidance from the distilled Slow branch. 
The resulting distilled Slow--Fast dual system uses two DiT evaluations per invocation of each branch while preserving the original model architecture, action horizon, and asynchronous closed-loop execution protocol.

For each distillation stage, indexed by $b \in \{S,F\}$ for the Slow and Fast branches, respectively, DMD maintains three networks: a deployable student generator $G_{\theta}^{b}$, a frozen real-score model $D_{r}^{b}$ initialized from the corresponding teacher, and an online fake-score model $D_{f}^{b}$ that tracks the evolving student distribution. The student produces a clean sample $x_g^b$ from a truncated two-step rollout. For the Slow branch, $x_g^S=(z_g,a_g)$ contains both future-video latents $z_g$ and an action chunk $a_g$; for the action-only Fast branch, $x_g^F=a_g$.

To estimate the distribution mismatch, we sample a noise level $\sigma \sim \mathcal{U}(0.02,0.98)$ and re-noise $x_g^b$. The real- and fake-score models then produce the corresponding clean-sample estimates $\hat{x}_{0,r}^{b}$ and $\hat{x}_{0,f}^{b}$. Here, $\hat{x}_{0,r}^{b}$ represents the estimate induced by the frozen teacher distribution, whereas $\hat{x}_{0,f}^{b}$ represents the estimate induced by the current student distribution. Their normalized difference defines the distribution-matching direction
\begin{equation}
    \label{eq:dmd_direction}
    g_{\mathrm{DMD}}^{b}
    =
    \frac{\hat{x}_{0,f}^{b}-\hat{x}_{0,r}^{b}}
    {\operatorname{mean}_{\mathrm{valid}}
    (\left|x_g^{b}-\hat{x}_{0,r}^{b}\right|)+\epsilon_{\mathrm{num}}}.
\end{equation}
The operator $\operatorname{mean}_{\mathrm{valid}}(\cdot)$ averages only over valid, non-masked video or action dimensions, and $\epsilon_{\mathrm{num}}$ is a small constant for numerical stability. The DMD direction is treated as a stop-gradient target: gradients propagate through the generated sample $x_g^b$, but not through the two score-model predictions or the normalization term.

We use a truncated two-step UniPC rollout~\cite{NEURIPS2023_9c2aa1e4} and randomly sample one of the two denoising steps for gradient computation. All preceding steps are evaluated without gradient, so each update backpropagates through only the selected student step. The corresponding shifted noise level is obtained from the scheduler, avoiding reconstruction from discretized timesteps. In both stages, we adopt the two-time-scale optimization scheme of improved DMD~\cite{NEURIPS2024_54dcf253}, updating the fake-score model five times for every generator update to closely track the evolving distribution.

\paragraph{Slow-branch distillation.}
The Slow student and its real- and fake-score models are initialized from the task-specialized eight-evaluation Slow branch. 
Because Slow jointly predicts future-video latents and actions, its distillation objective preserves both output spaces. 
The generator retains the original supervised flow-matching objective $\mathcal{L}_{\mathrm{sup}}^{S}$ as an optimization anchor and is trained with
\begin{equation}
    \mathcal{L}_{g}^{S}
    =
    \mathcal{L}_{\mathrm{sup}}^{S}
    + \mathcal{L}_{\mathrm{DMD},v}^{S}
    + \mathcal{L}_{\mathrm{DMD},a}^{S},
\end{equation}
while the online fake-score model is optimized using
\begin{equation}
    \mathcal{L}_{f}^{S}
    =
    \mathcal{L}_{\mathrm{FM},v}^{S}
    + \mathcal{L}_{\mathrm{FM},a}^{S}.
\end{equation}
The subscripts $g$ and $f$ denote the generator and fake-score objectives, respectively, while the superscripts $S$ and $F$ denote the Slow and Fast branches. The subscript $\mathrm{FM}$ denotes flow matching. Here, $\mathcal{L}_{\mathrm{DMD},v}^{S}$ and $\mathcal{L}_{\mathrm{DMD},a}^{S}$ apply the direction in Eq.~(\ref{eq:dmd_direction}) to the generated video and action components, respectively. The terms $\mathcal{L}_{\mathrm{FM},v}^{S}$ and $\mathcal{L}_{\mathrm{FM},a}^{S}$ train the fake-score model with standard flow matching on detached Slow-student samples. All loss terms use unit weights. This stage compresses the Slow rollout while preserving the coupling between predicted visual dynamics and their corresponding action trajectories.

\paragraph{Fast-branch distillation.}
After the distilled Slow branch converges, we freeze it in evaluation mode and detach its layer-wise video K/V features from the Fast optimization graph. For each training sample, the distilled Slow branch performs two DiT evaluations and exports a layer-wise video K/V representation at each denoising step. The Fast branch also performs two DiT evaluations, with its $j$-th denoising step consuming the Slow guidance exported at the corresponding step $j$, for $j \in \{0,1\}$. Unlike the Slow branch, the Fast branch predicts actions only; therefore, $x_g^F$, $\hat{x}_{0,r}^{F}$, and $\hat{x}_{0,f}^{F}$ contain only action tokens. We initialize the Fast student and its real- and fake-score models from the task-specialized eight-evaluation Fast branch and optimize
\begin{equation}
    \mathcal{L}_{g}^{F}
    =
    \mathcal{L}_{\mathrm{sup}}^{F}
    + \mathcal{L}_{\mathrm{DMD},a}^{F},
\end{equation}
and
\begin{equation}
    \mathcal{L}_{f}^{F}
    =
    \mathcal{L}_{\mathrm{FM},a}^{F}.
\end{equation}
The supervised term $\mathcal{L}_{\mathrm{sup}}^{F}$ anchors the Fast student to the ground-truth action flow, $\mathcal{L}_{\mathrm{DMD},a}^{F}$ matches the student action distribution to the frozen Fast teacher, and $\mathcal{L}_{\mathrm{FM},a}^{F}$ updates the online fake-score model using detached Fast-student actions. Only the Fast student and its online fake-score model are updated during this stage; the distilled Slow conditioner and the Fast real-score model remain frozen. At deployment, all real- and fake-score models are discarded, leaving only the two-step Slow and Fast students for inference.

\section{Experiments}
\subsection{Zero-Shot Real-Robot Evaluation}
\label{ssec:real_robot_scaling}

To validate the individual contributions of egocentric video pre-training and multi-embodiment video-action mid-training, we conduct a series of real-robot evaluations. The evaluation suite consists of 12 zero-shot tasks categorized into two distinct protocols: Standard tasks and Perturbed tasks.

\subsubsection{Evaluation Protocol}
\label{sssec:real_robot_evaluation_setup}

\paragraph{Evaluation setting.}
All experiments are conducted on a 7-DoF Franka arm platform. Each \name{} variant is post-trained on the DROID dataset and evaluated on held-out task--scene configurations without task-specific demonstrations. At deployment, the policy receives two external RGB views, one wrist-camera view, the current proprioceptive state, and a language instruction. 

\begin{figure*}[p]
  \centering
  \small
  % 压缩行高与表格边距
  \renewcommand{\arraystretch}{0.95}
  \setlength{\tabcolsep}{1.5pt}

  % ===== Subfigure (a): Standard Suite =====
  \begin{subfigure}{\linewidth}
    \centering
    \begin{tabular}{@{}c@{\hspace{2pt}}cccc@{}}
      & {\small\textbf{1. Bowl Stacking}}
      & {\small\textbf{2. Cup Selection}}
      & {\small\textbf{3. Bread Transfer}}
      & {\small\textbf{4. Block Stacking}} \\[1pt]
      \raisebox{-.5\height}{\rotatebox[origin=c]{90}{\small Initial}} &
      \raisebox{-.5\height}{\includegraphics[width=0.205\linewidth]{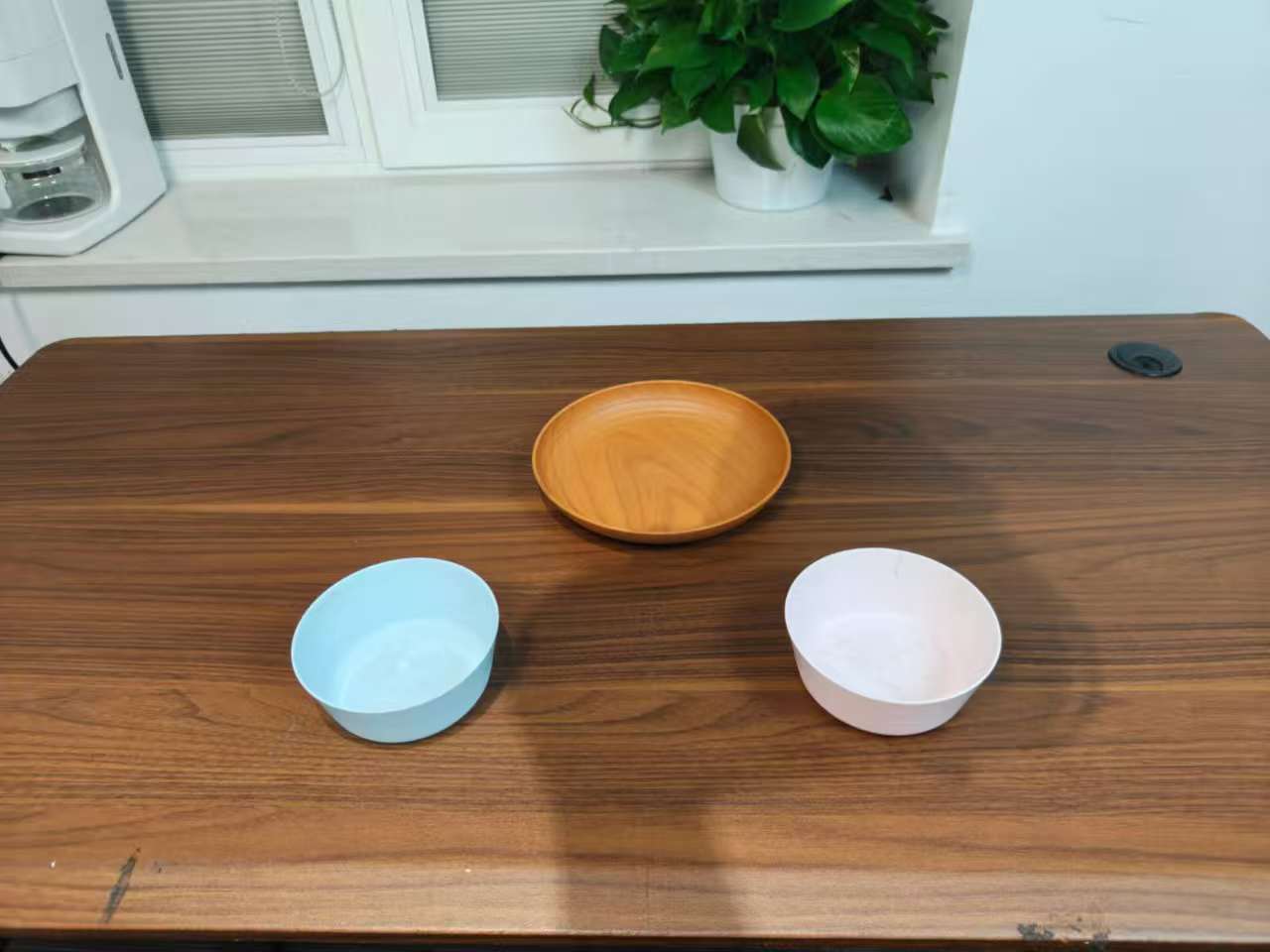}} &
      \raisebox{-.5\height}{\includegraphics[width=0.205\linewidth]{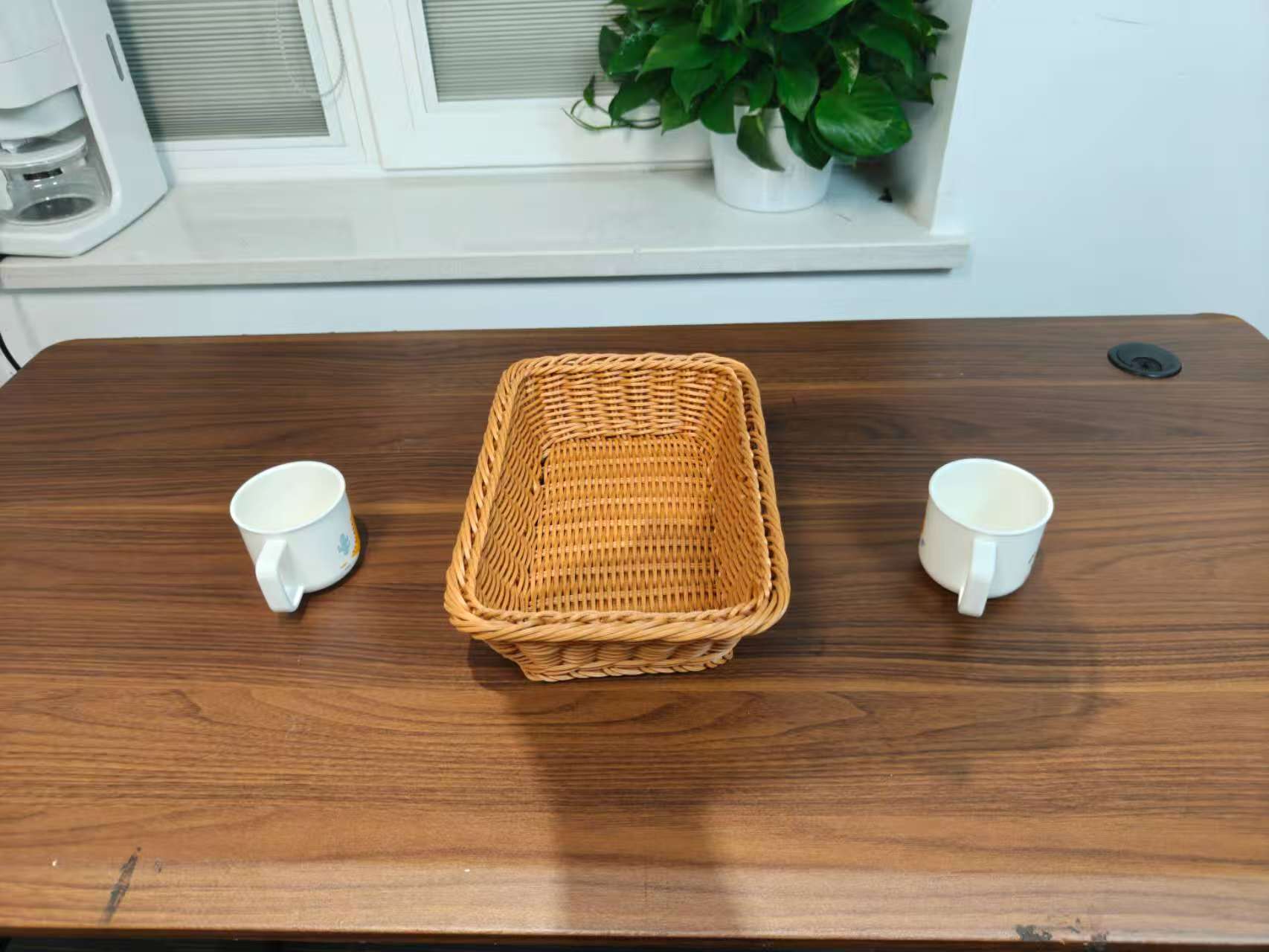}} &
      \raisebox{-.5\height}{\includegraphics[width=0.205\linewidth]{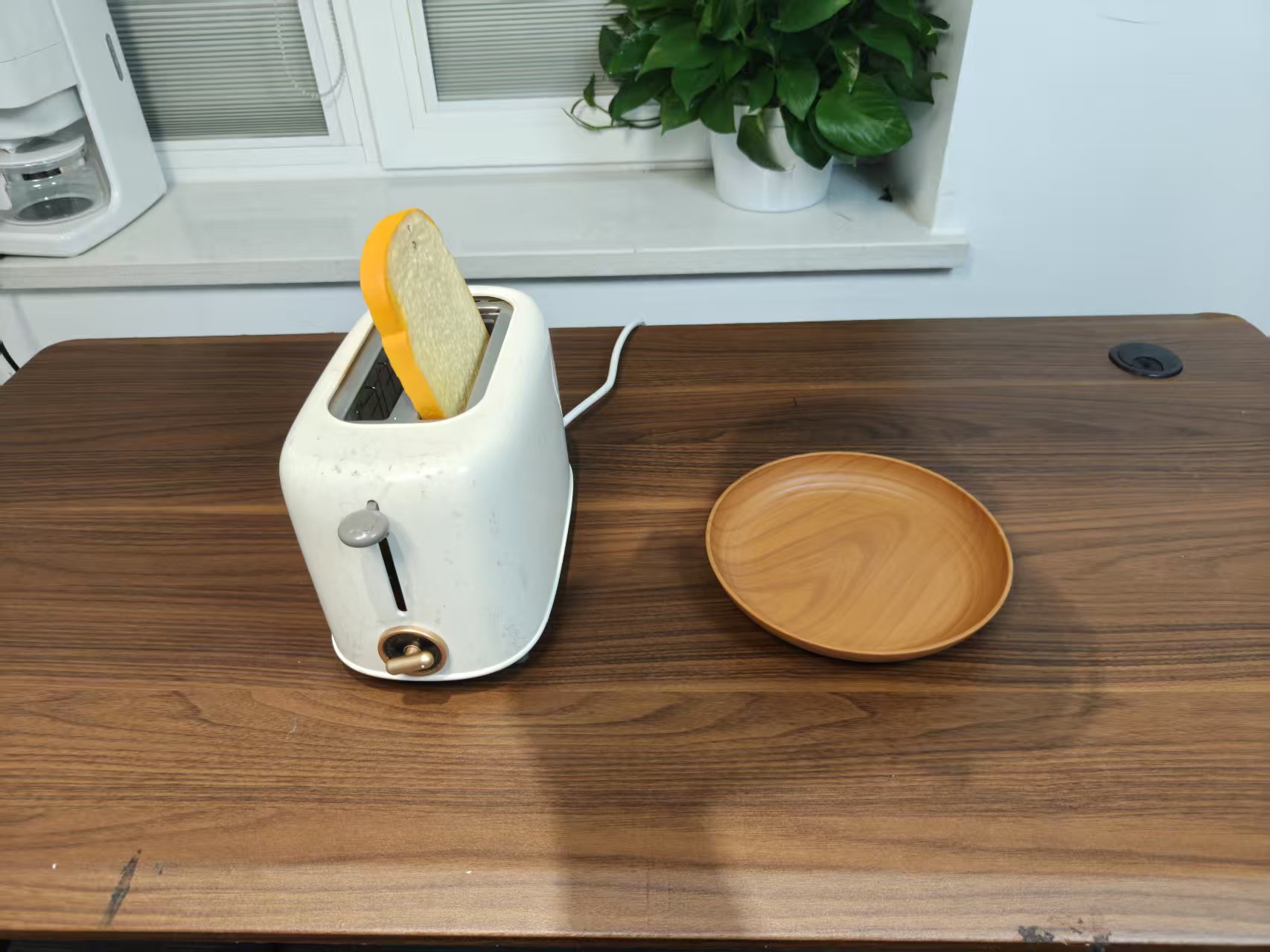}} &
      \raisebox{-.5\height}{\includegraphics[width=0.205\linewidth]{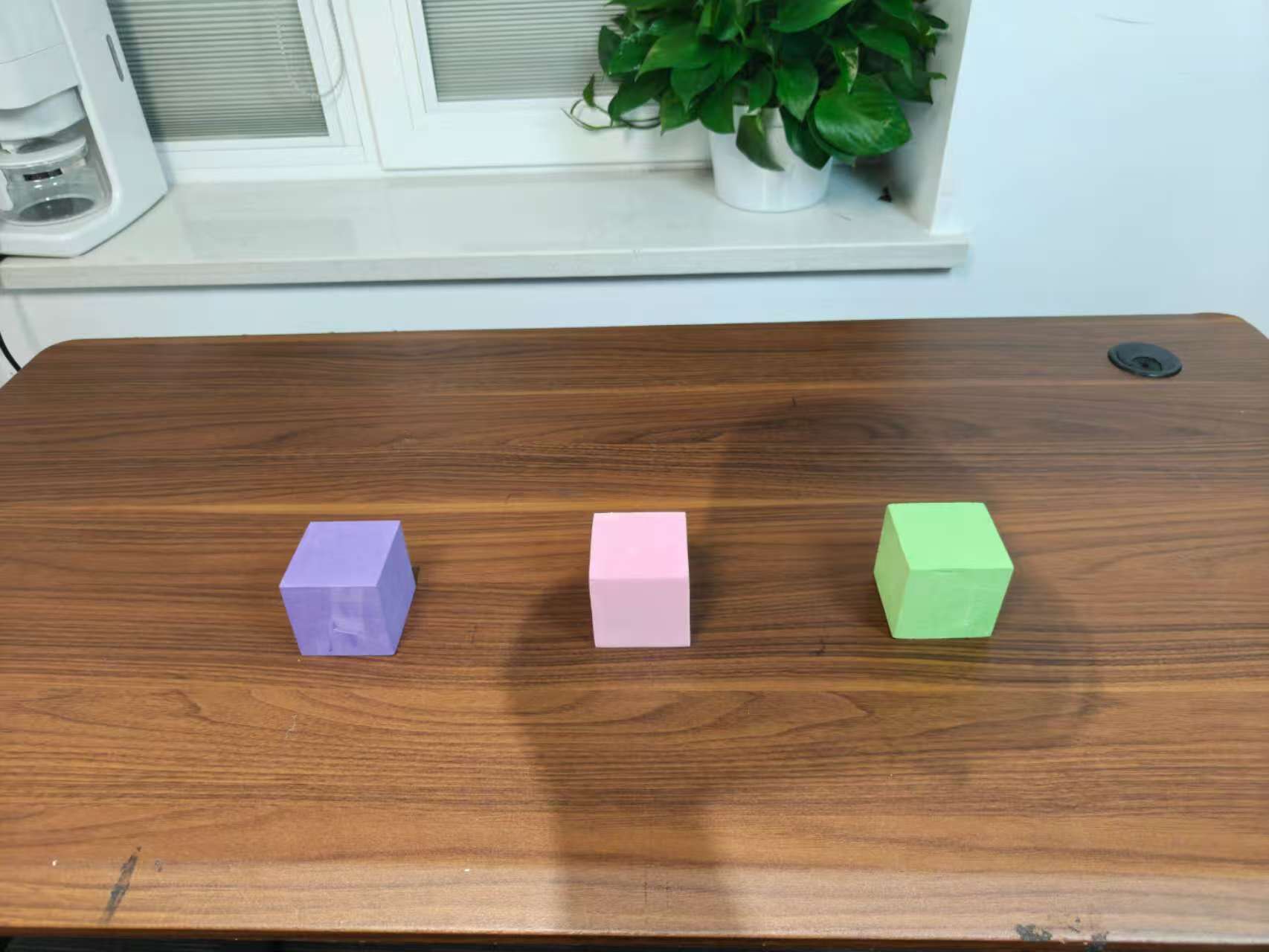}} \\
      \raisebox{-.5\height}{\rotatebox[origin=c]{90}{\small Final}} &
      \raisebox{-.5\height}{\includegraphics[width=0.205\linewidth]{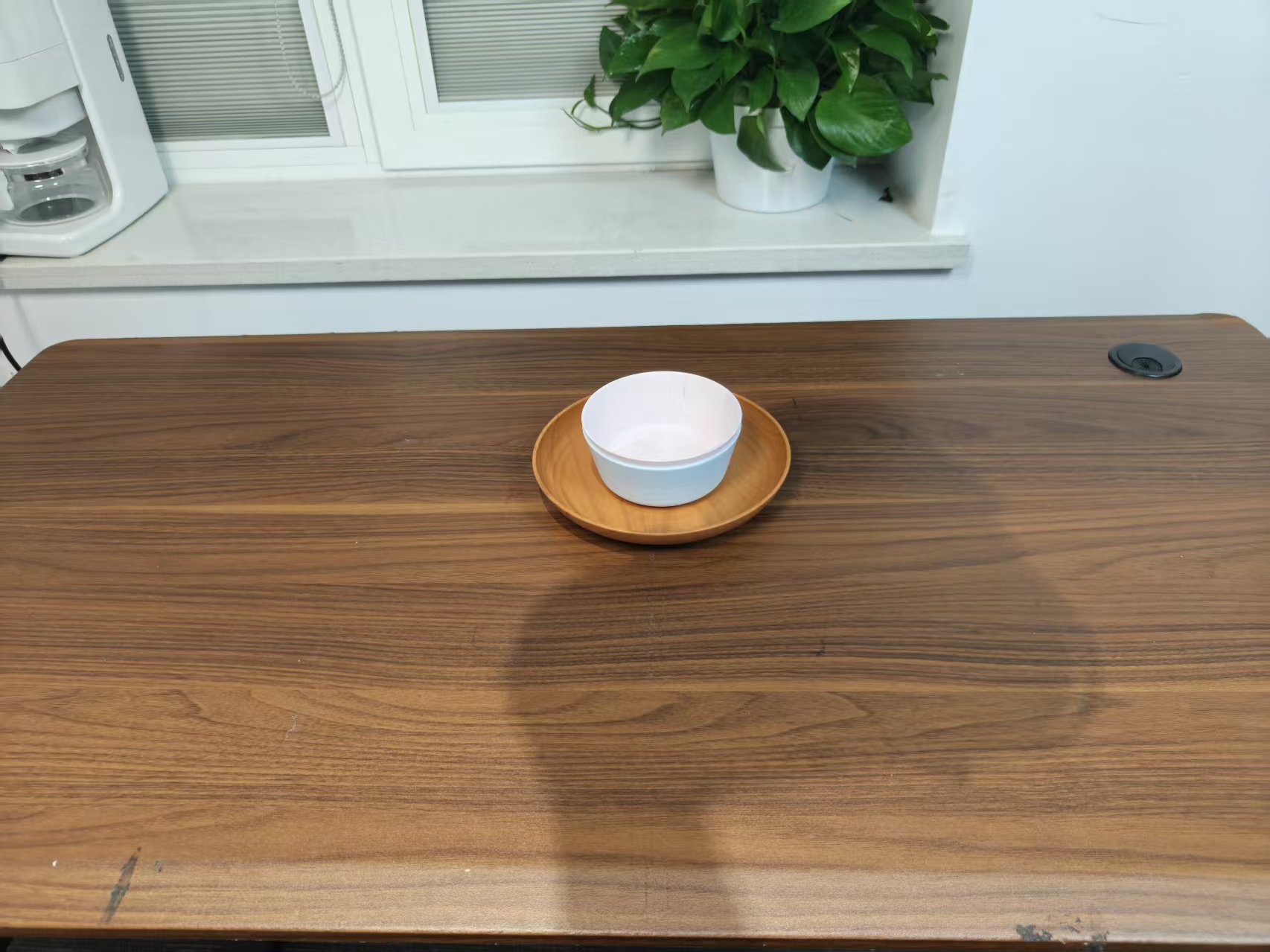}} &
      \raisebox{-.5\height}{\includegraphics[width=0.205\linewidth]{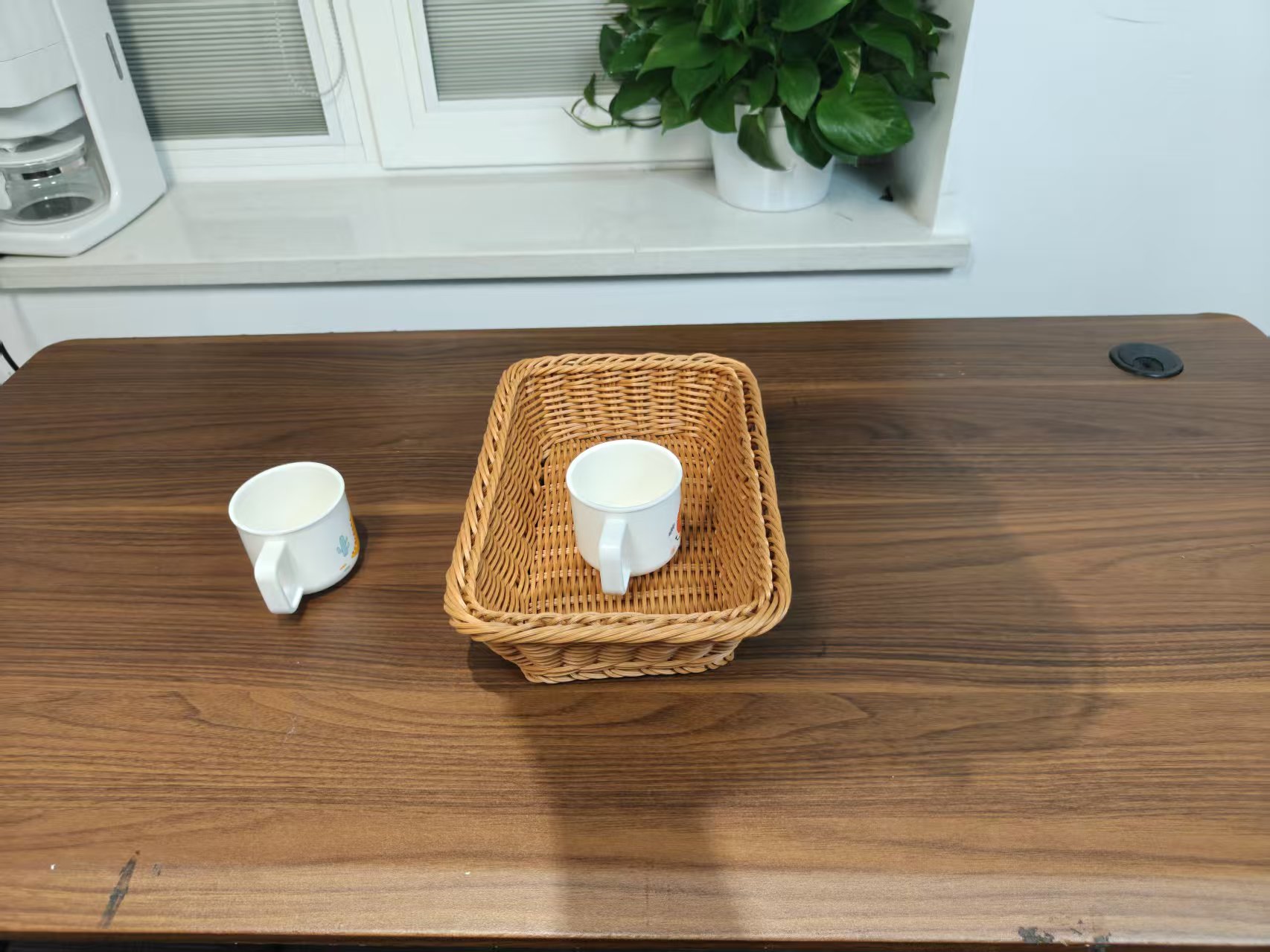}} &
      \raisebox{-.5\height}{\includegraphics[width=0.205\linewidth]{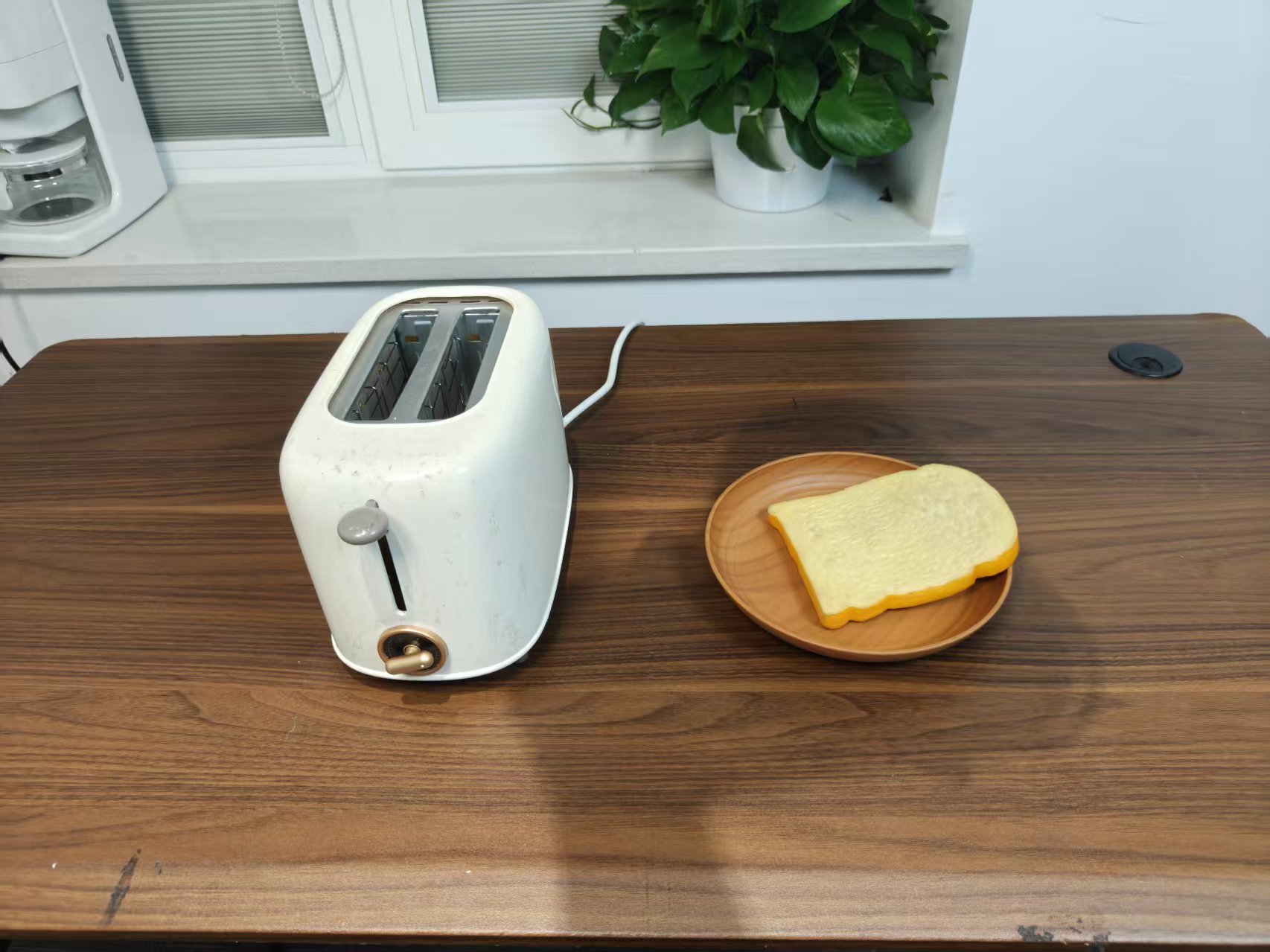}} &
      \raisebox{-.5\height}{\includegraphics[width=0.205\linewidth]{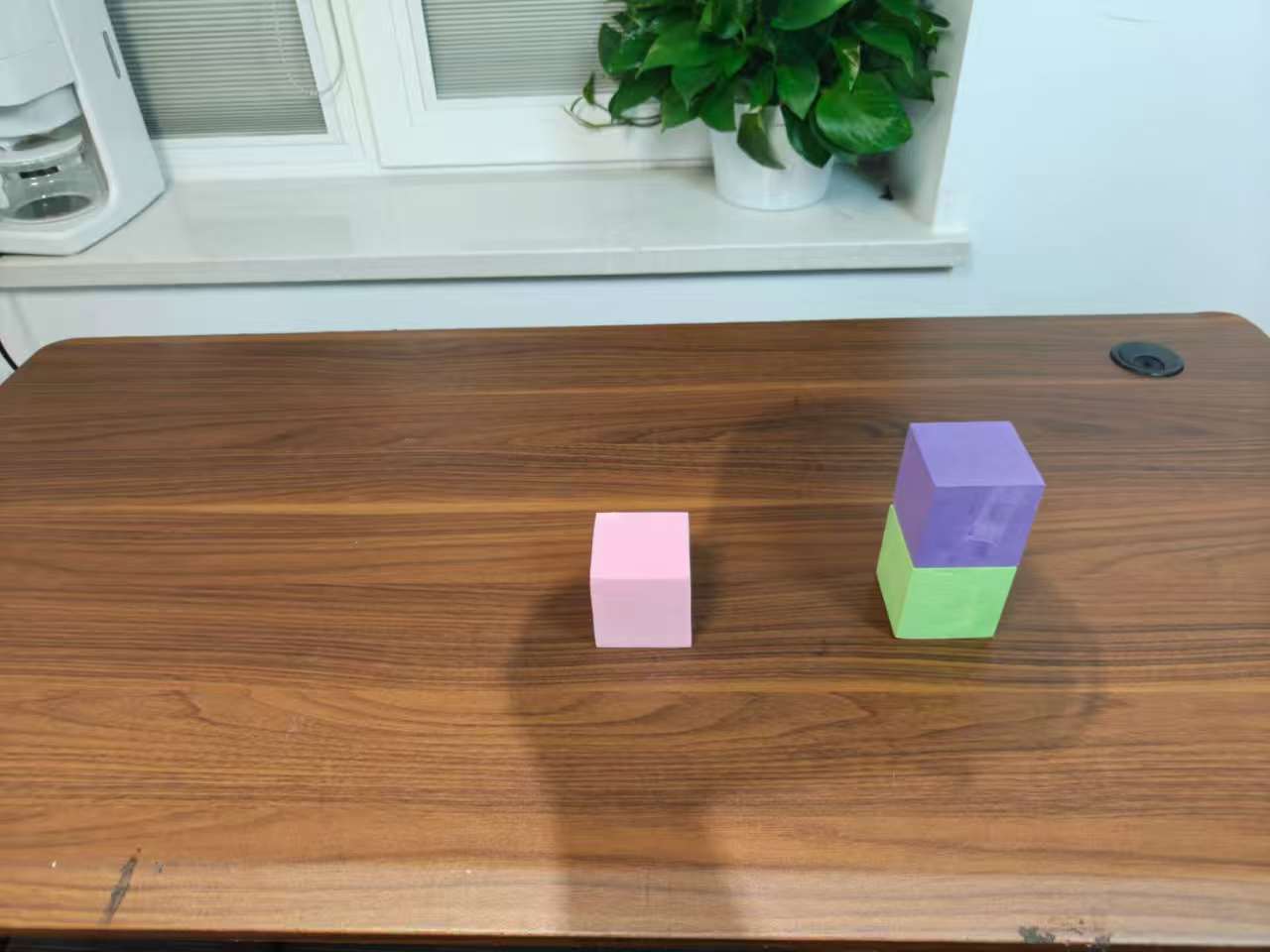}} \\[3pt]
      % \rowcolor{vscaleaccent!6}
      \hspace{1pt}\raisebox{0pt}{\rotatebox[origin=c]{90}{\small Prompt}}\hspace{-1pt} &
      \parbox[c][2.2em][c]{0.205\linewidth}{\centering\small\itshape\shortstack{Stack bowls and place\\them on a plate.}}
      & \parbox[c][2.2em][c]{0.205\linewidth}{\centering\small\itshape\shortstack{Pick up the instructed\\left or right cup.}}
      & \parbox[c][2.2em][c]{0.205\linewidth}{\centering\small\itshape\shortstack{Move bread from the\\toaster to a plate.}}
      & \parbox[c][2.2em][c]{0.205\linewidth}{\centering\small\itshape\shortstack{Stack the purple block\\on instructed block.}} \\[12pt]
      & {\small\textbf{5. Air-fryer Opening}}
      & {\small\textbf{6. Toy Collection}}
      & {\small\textbf{7. Microwave Closing}}
      & {\small\textbf{8. Toaster Activation}} \\[1pt]
      \raisebox{-.5\height}{\rotatebox[origin=c]{90}{\small Initial}} &
      \raisebox{-.5\height}{\includegraphics[width=0.205\linewidth]{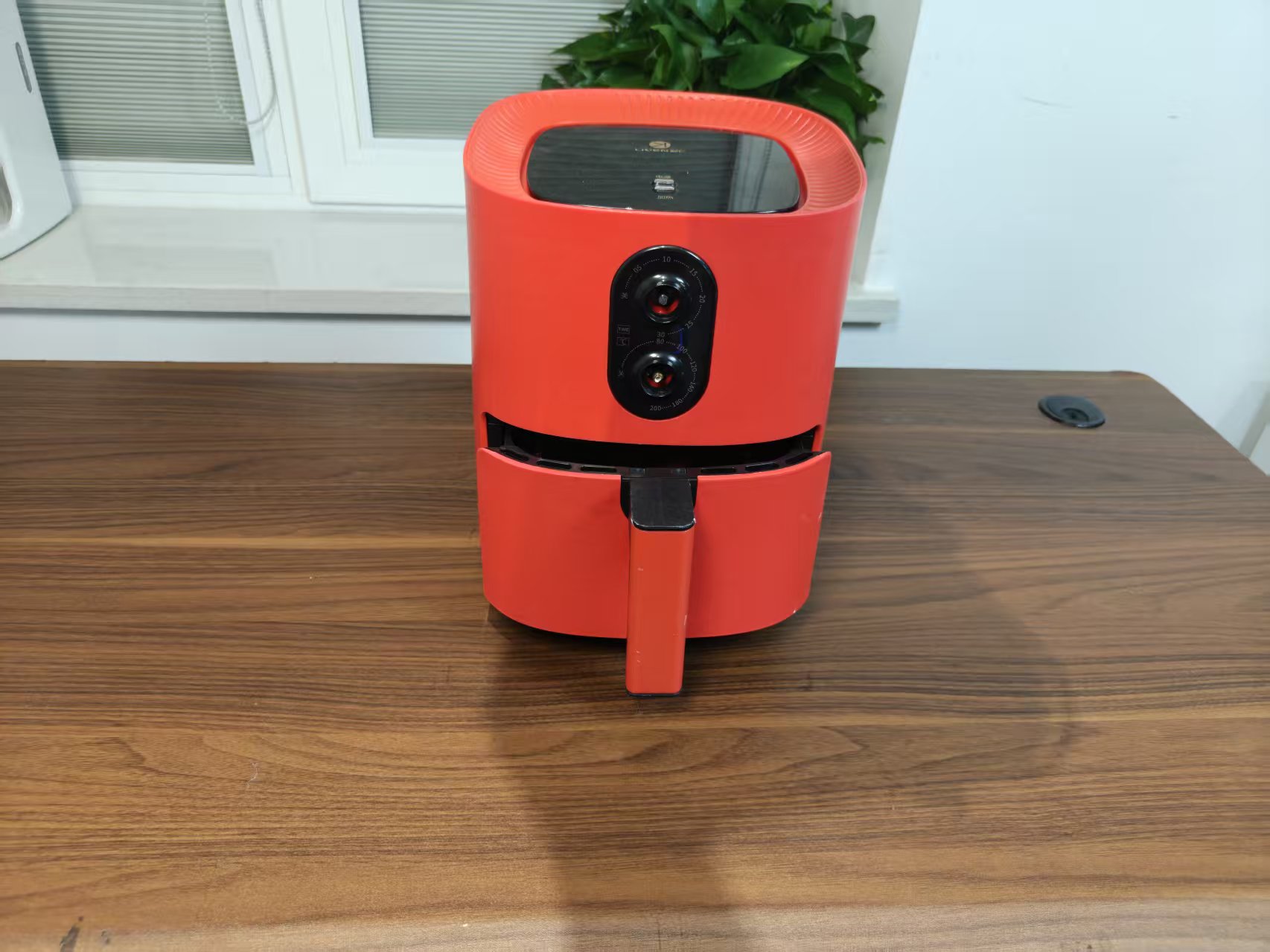}} &
      \raisebox{-.5\height}{\includegraphics[width=0.205\linewidth]{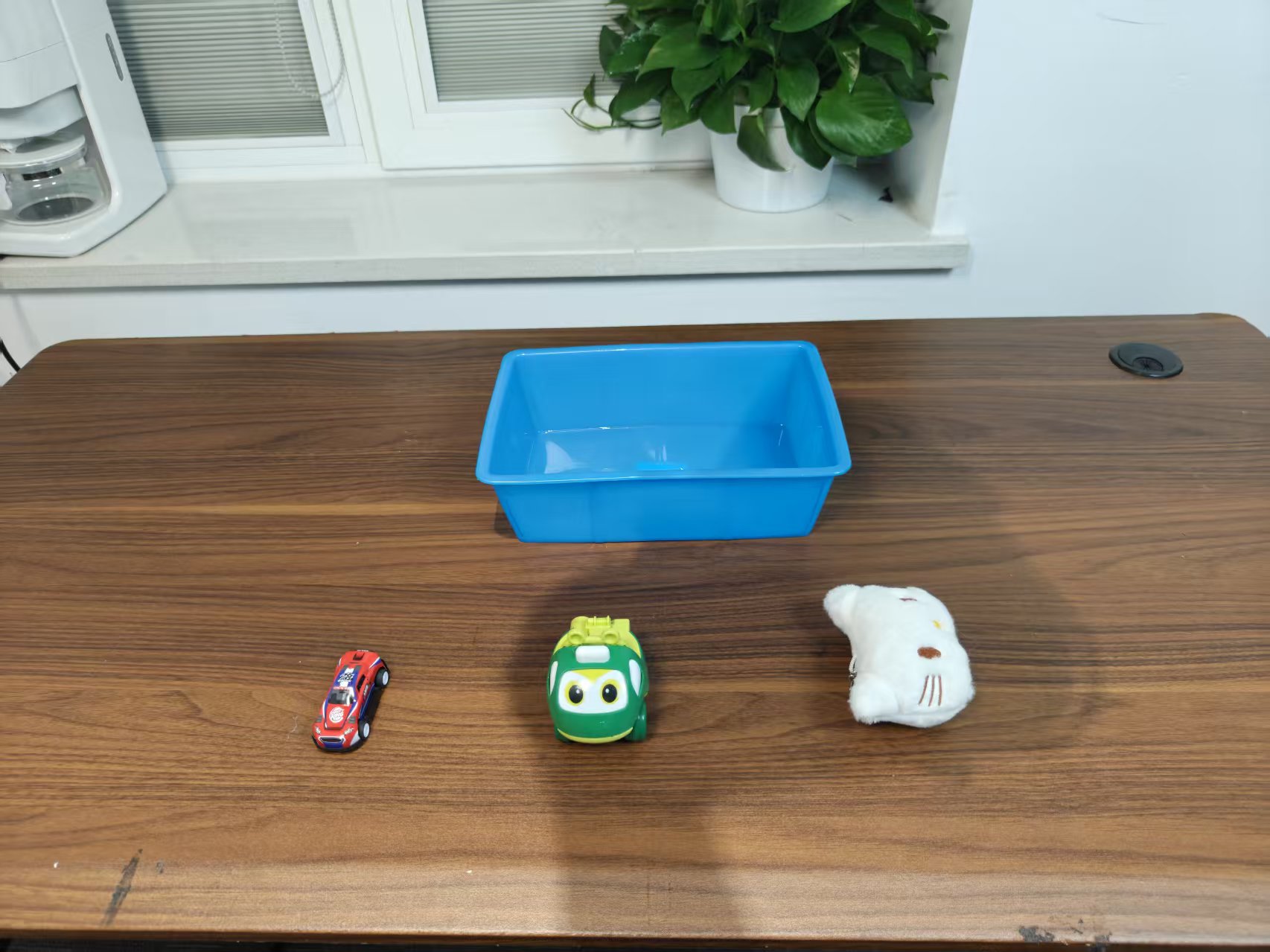}} &
      \raisebox{-.5\height}{\includegraphics[width=0.205\linewidth]{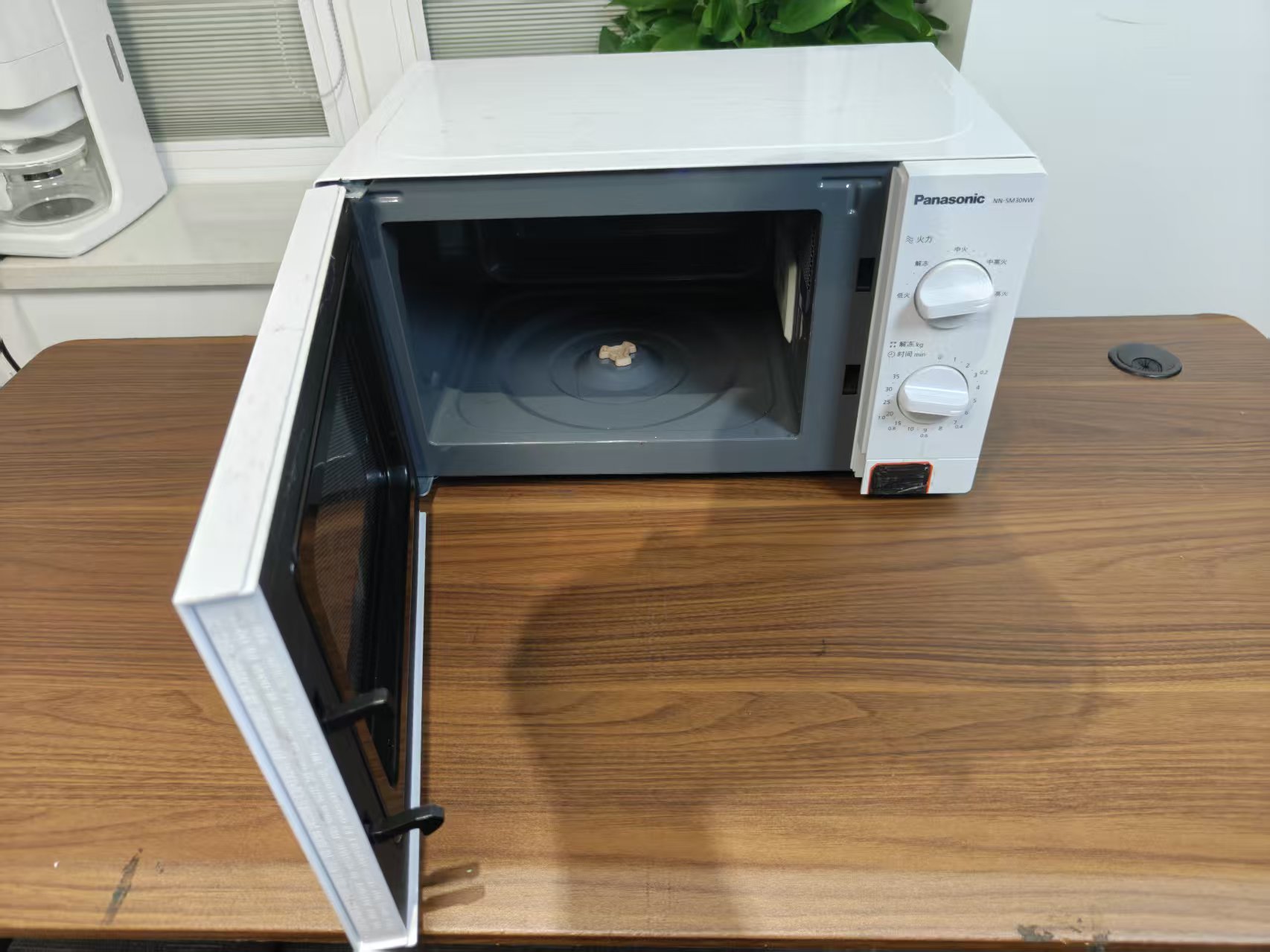}} &
      \raisebox{-.5\height}{\includegraphics[width=0.205\linewidth]{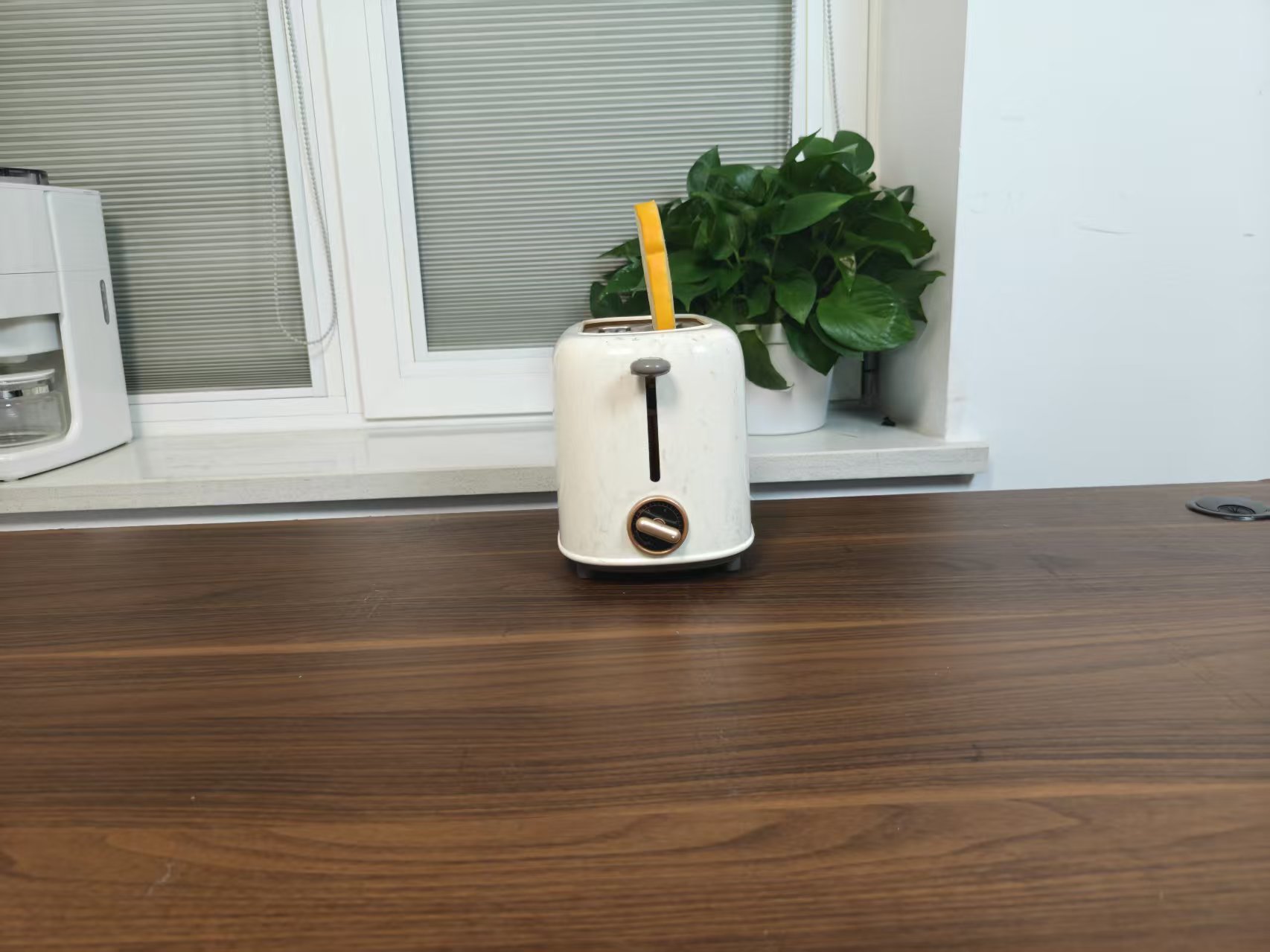}} \\
      \raisebox{-.5\height}{\rotatebox[origin=c]{90}{\small Final}} &
      \raisebox{-.5\height}{\includegraphics[width=0.205\linewidth]{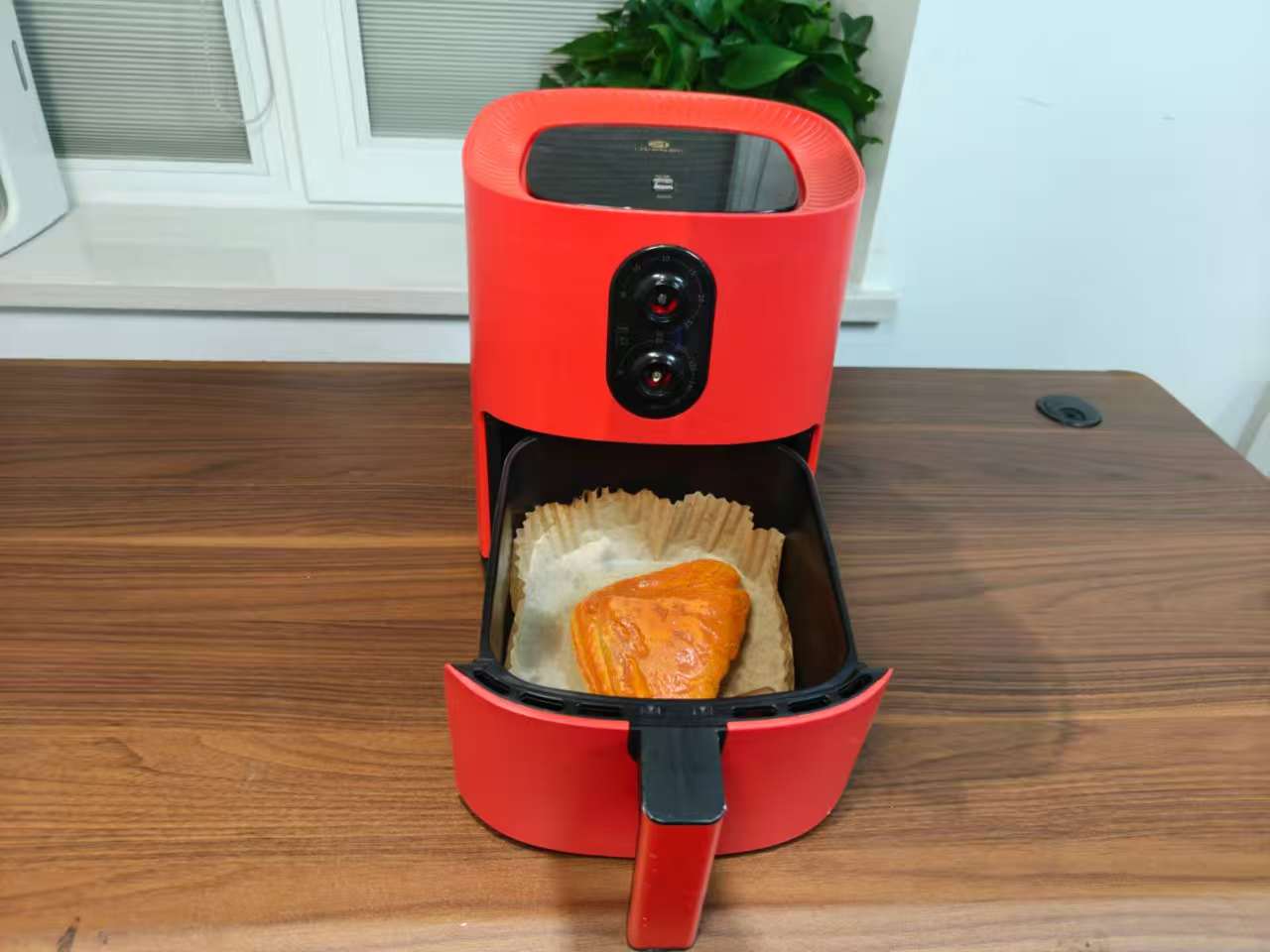}} &
      \raisebox{-.5\height}{\includegraphics[width=0.205\linewidth]{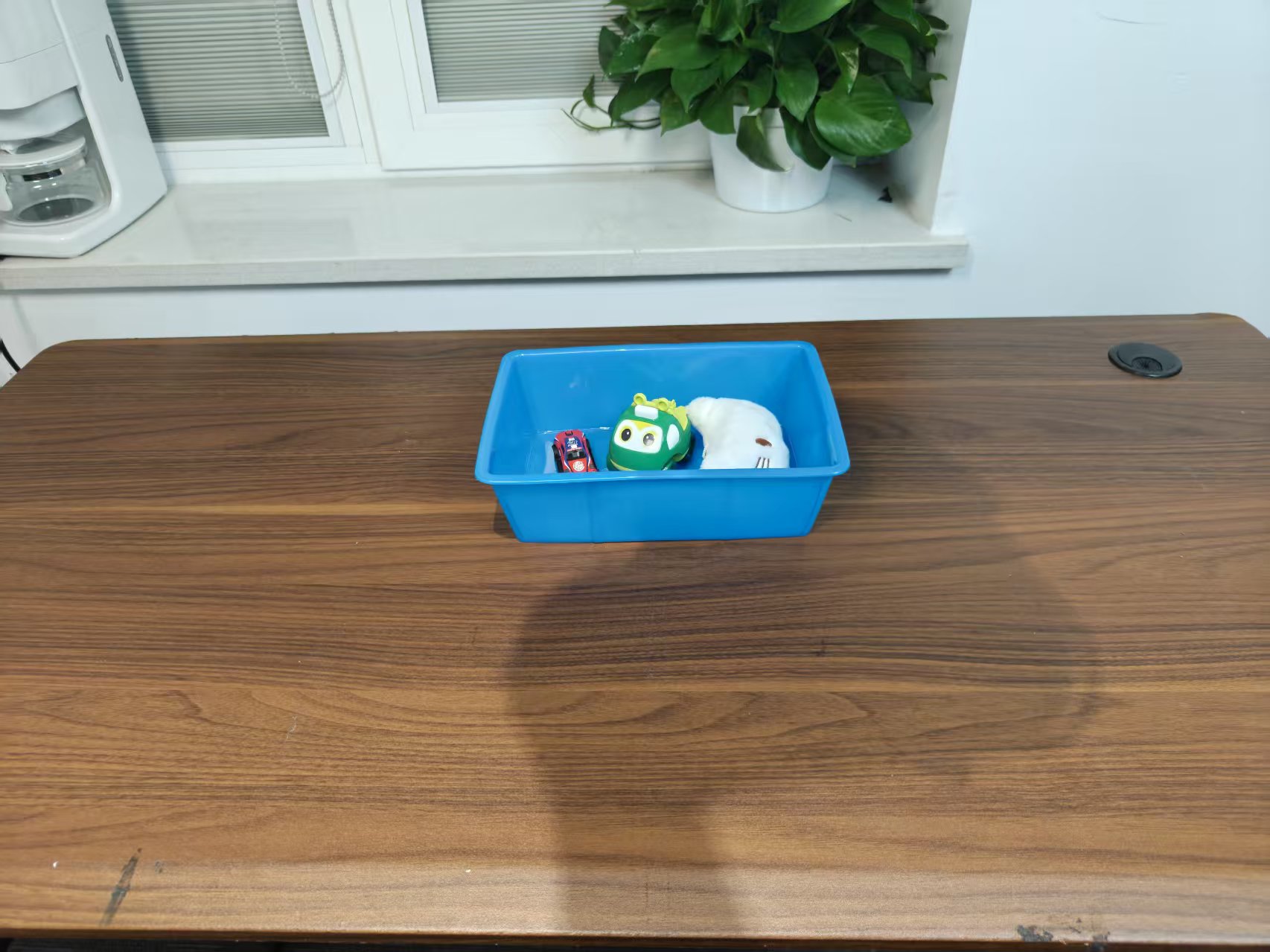}} &
      \raisebox{-.5\height}{\includegraphics[width=0.205\linewidth]{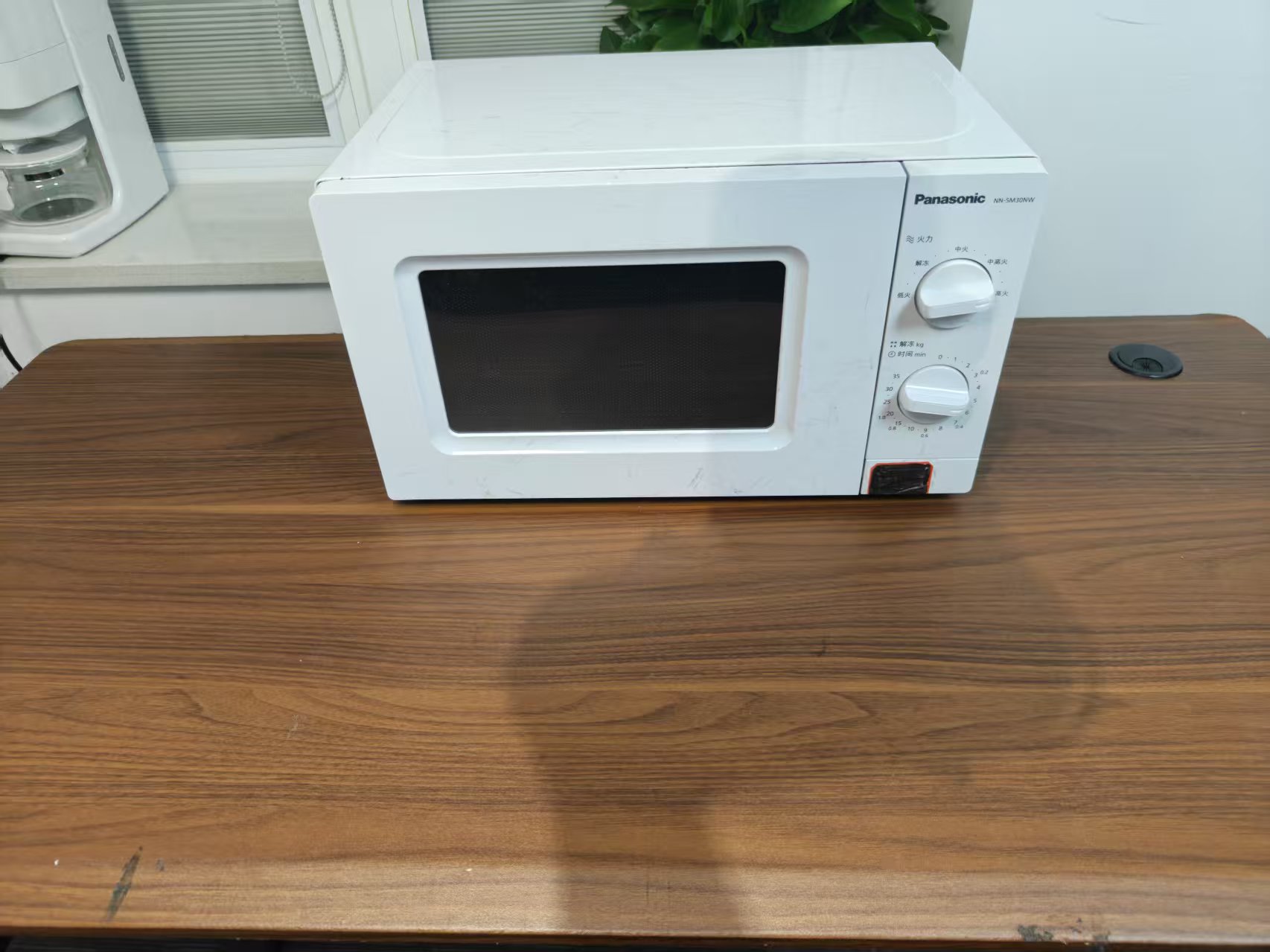}} &
      \raisebox{-.5\height}{\includegraphics[width=0.205\linewidth]{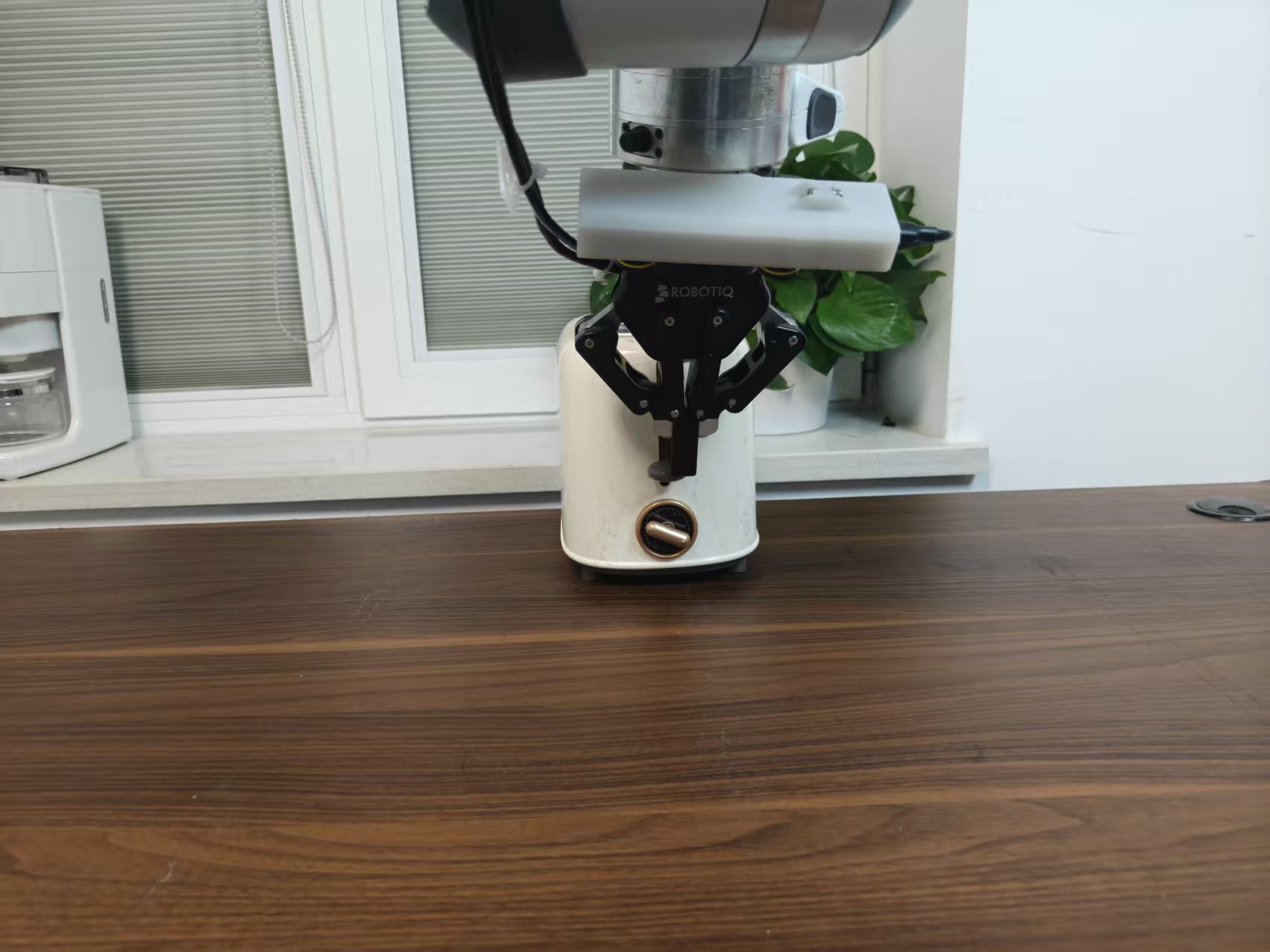}} \\[3pt]
      % \rowcolor{vscaleaccent!6}
      \hspace{1pt}\raisebox{0pt}{\rotatebox[origin=c]{90}{\small Prompt}}\hspace{-1pt} &
      \parbox[c][2.2em][c]{0.205\linewidth}{\centering\small\itshape\shortstack{Open the air-fryer\\drawer.}}
      & \parbox[c][2.2em][c]{0.205\linewidth}{\centering\small\itshape\shortstack{Put all toys into\\the blue box.}}
      & \parbox[c][2.2em][c]{0.205\linewidth}{\centering\small\itshape\shortstack{Close the microwave\\door.}}
      & \parbox[c][2.2em][c]{0.205\linewidth}{\centering\small\itshape\shortstack{Push down the\\toaster switch.}}
    \end{tabular}
    \caption{\textbf{Standard real-robot suite.}}
    \label{fig:real_robot_standard_tasks}
  \end{subfigure}

  \vspace{4pt}

  % ===== Subfigure (b): Perturbed Suite =====
  \begin{subfigure}{\linewidth}
    \centering
    \begin{tabular}{@{}c@{\hspace{2pt}}cccc@{}}
      & {\small\textbf{9. Air-fryer Opening$^*$}}
      & {\small\textbf{10. Toaster Activation$^*$}}
      & {\small\textbf{11. Bread Placement}}
      & {\small\textbf{12. Bowl Stacking$^*$}} \\[1pt]
      \raisebox{-.5\height}{\rotatebox[origin=c]{90}{\small Initial}} &
      \raisebox{-.5\height}{\includegraphics[width=0.205\linewidth]{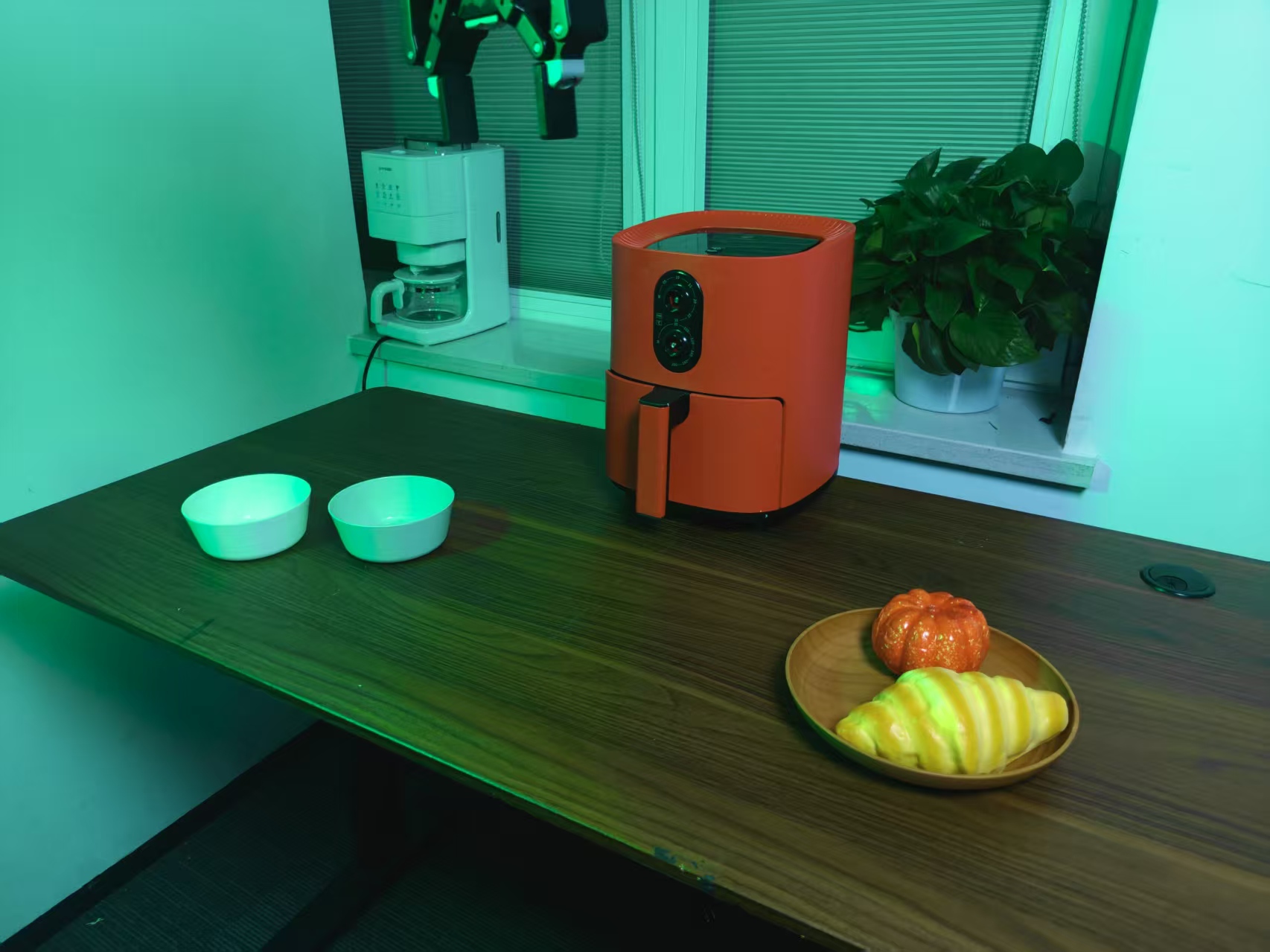}} &
      \raisebox{-.5\height}{\includegraphics[width=0.205\linewidth]{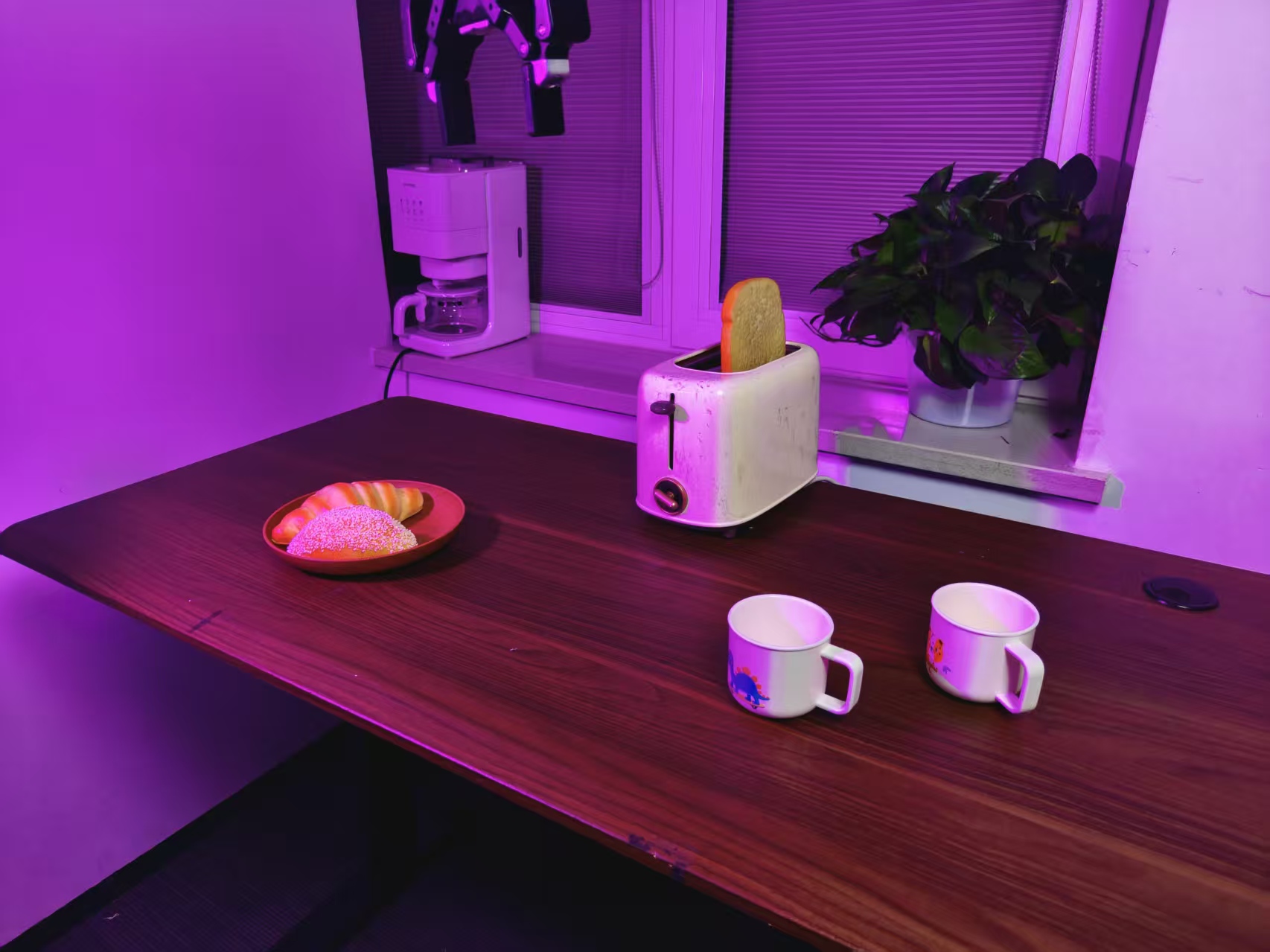}} &
      \raisebox{-.5\height}{\includegraphics[width=0.205\linewidth]{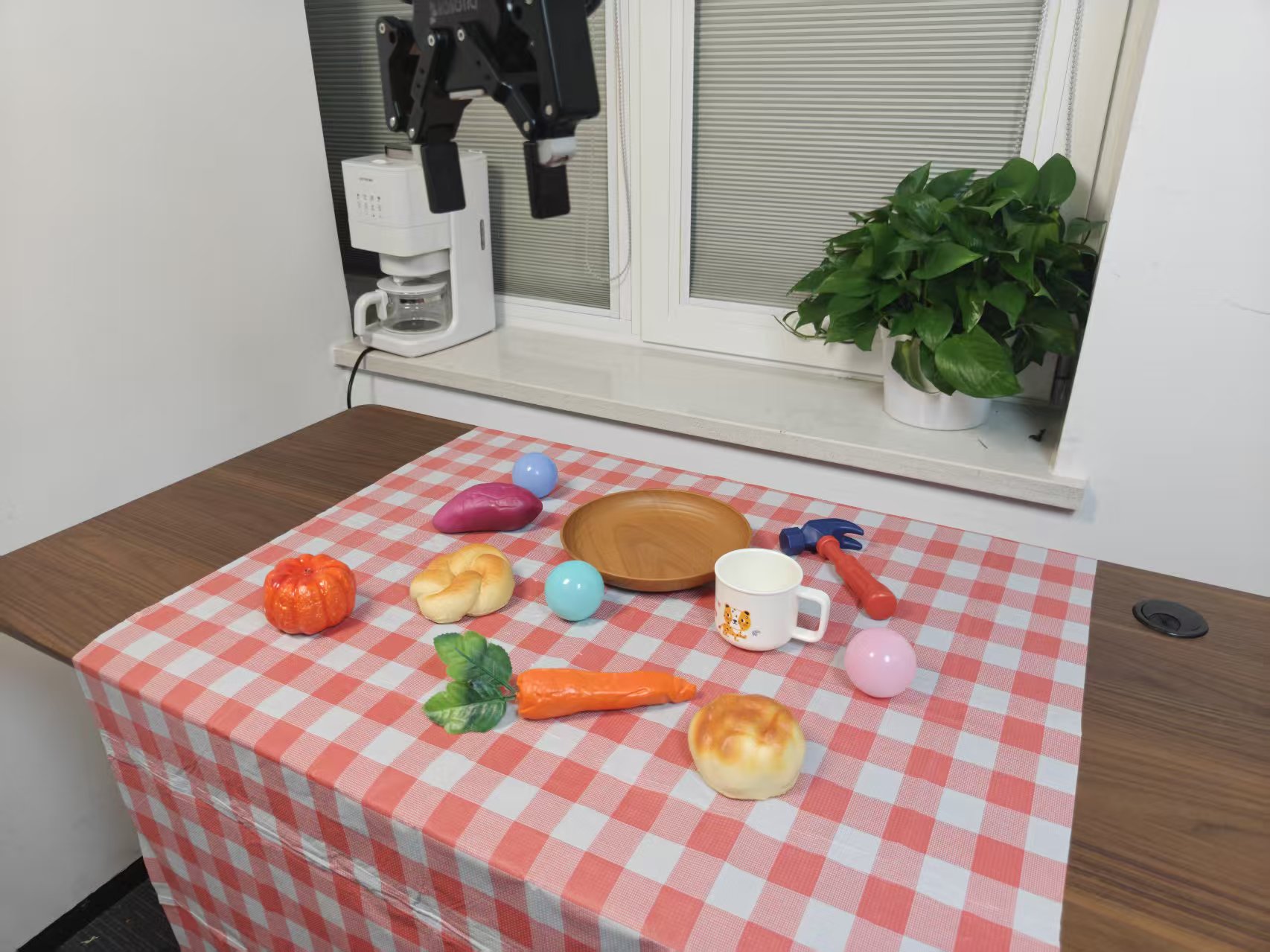}} &
      \raisebox{-.5\height}{\includegraphics[width=0.205\linewidth]{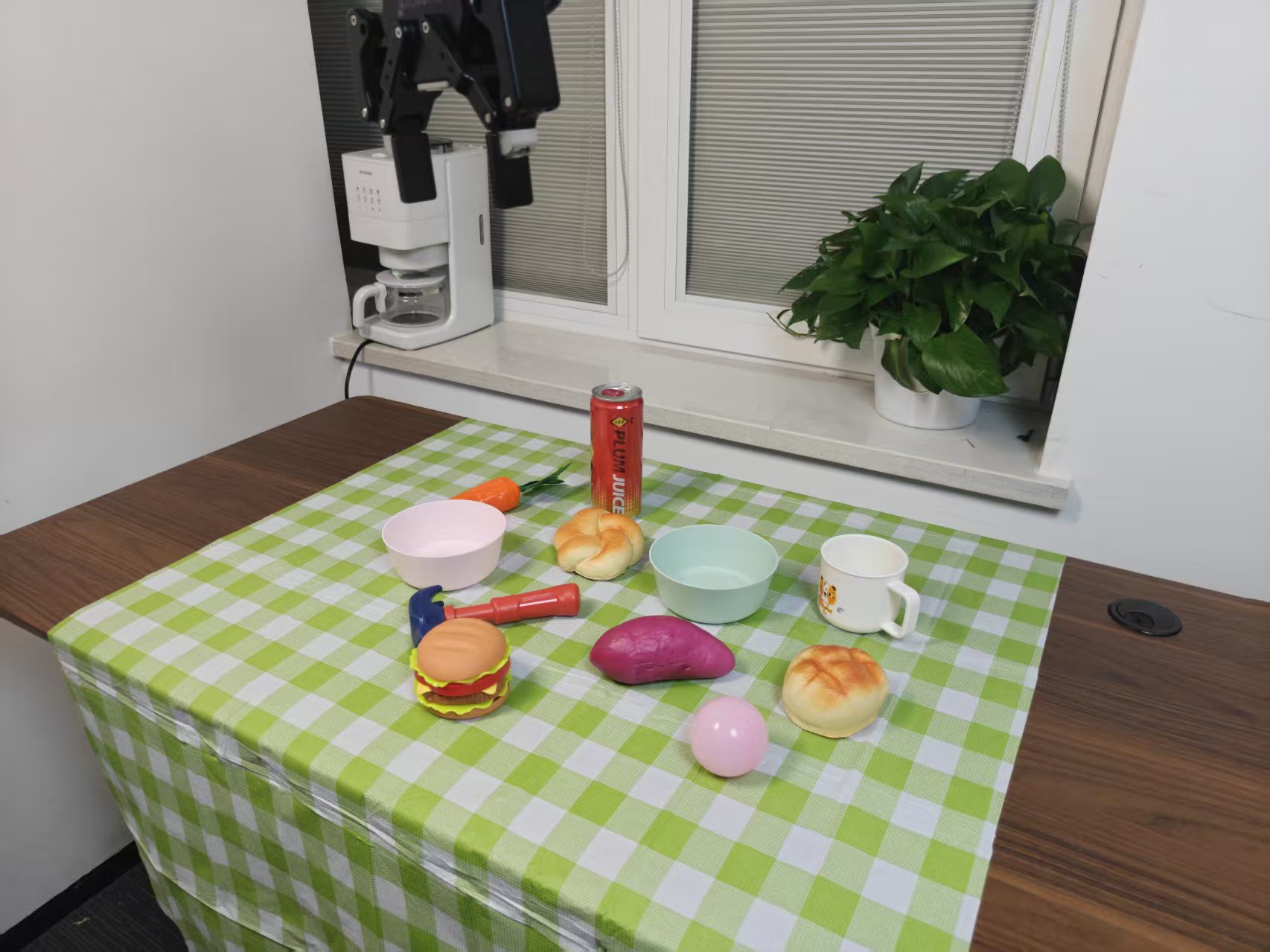}} \\
      \raisebox{-.5\height}{\rotatebox[origin=c]{90}{\small Final}} &
      \raisebox{-.5\height}{\includegraphics[width=0.205\linewidth]{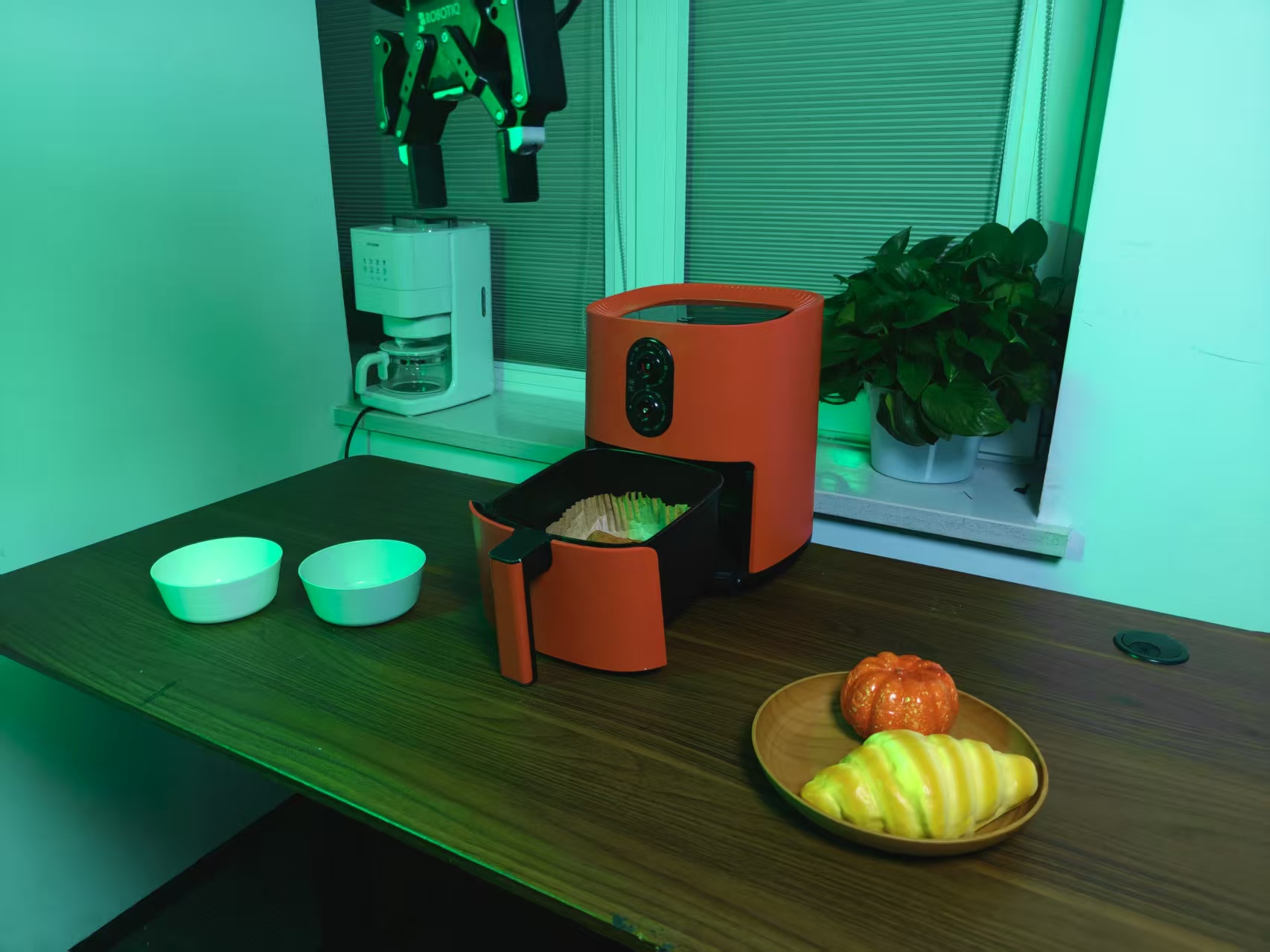}} &
      \raisebox{-.5\height}{\includegraphics[width=0.205\linewidth]{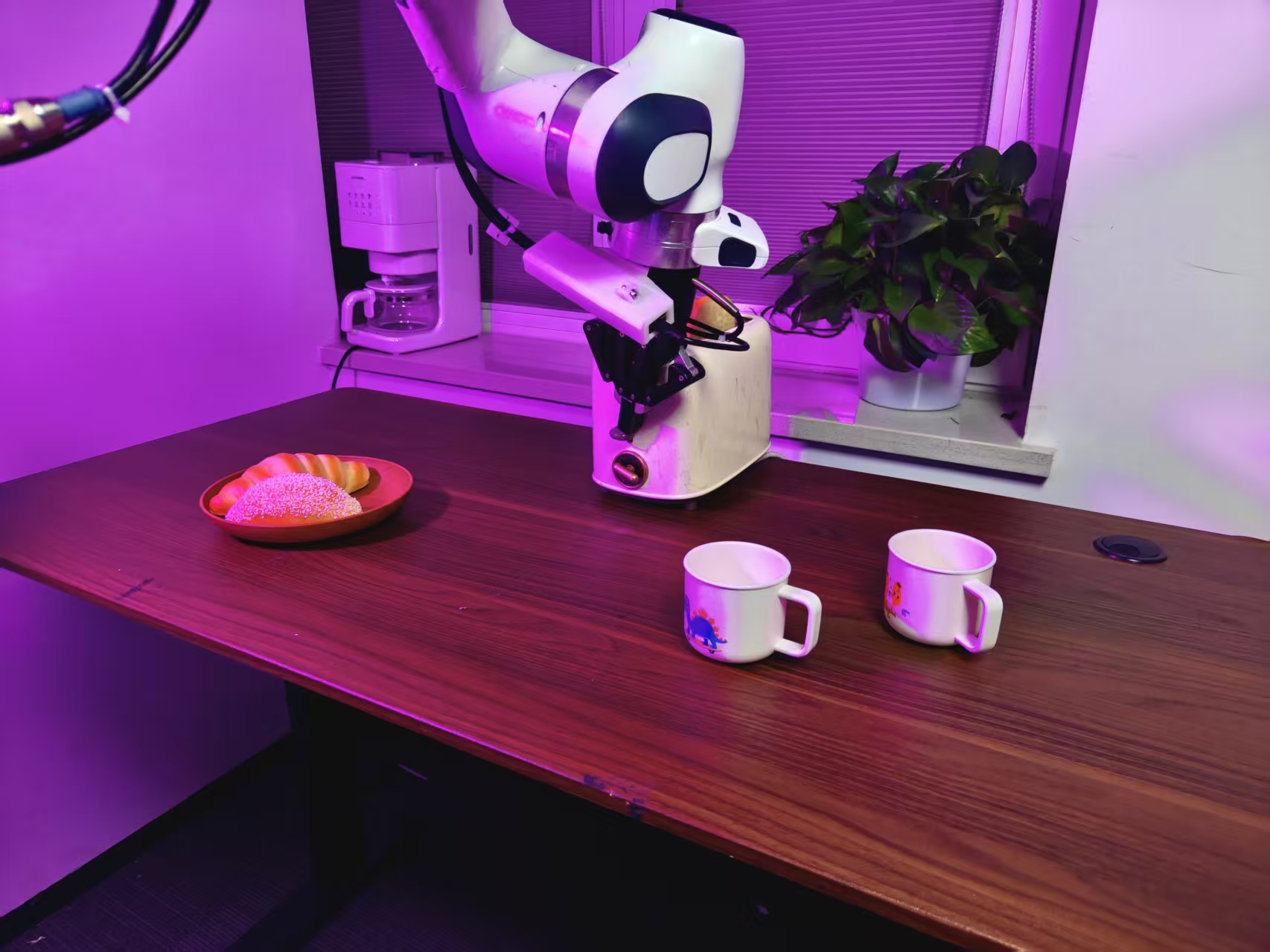}} &
      \raisebox{-.5\height}{\includegraphics[width=0.205\linewidth]{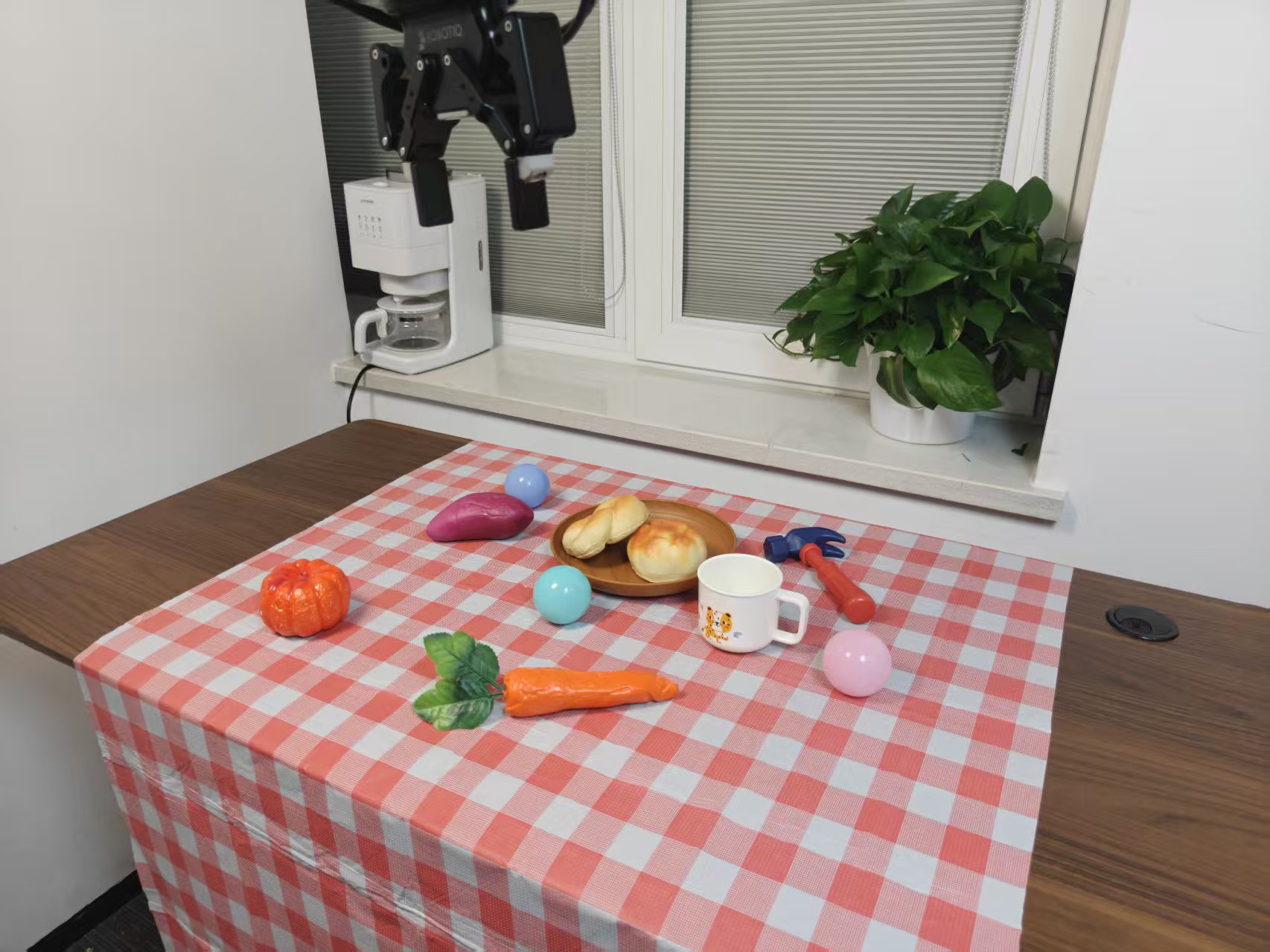}} &
      \raisebox{-.5\height}{\includegraphics[width=0.205\linewidth]{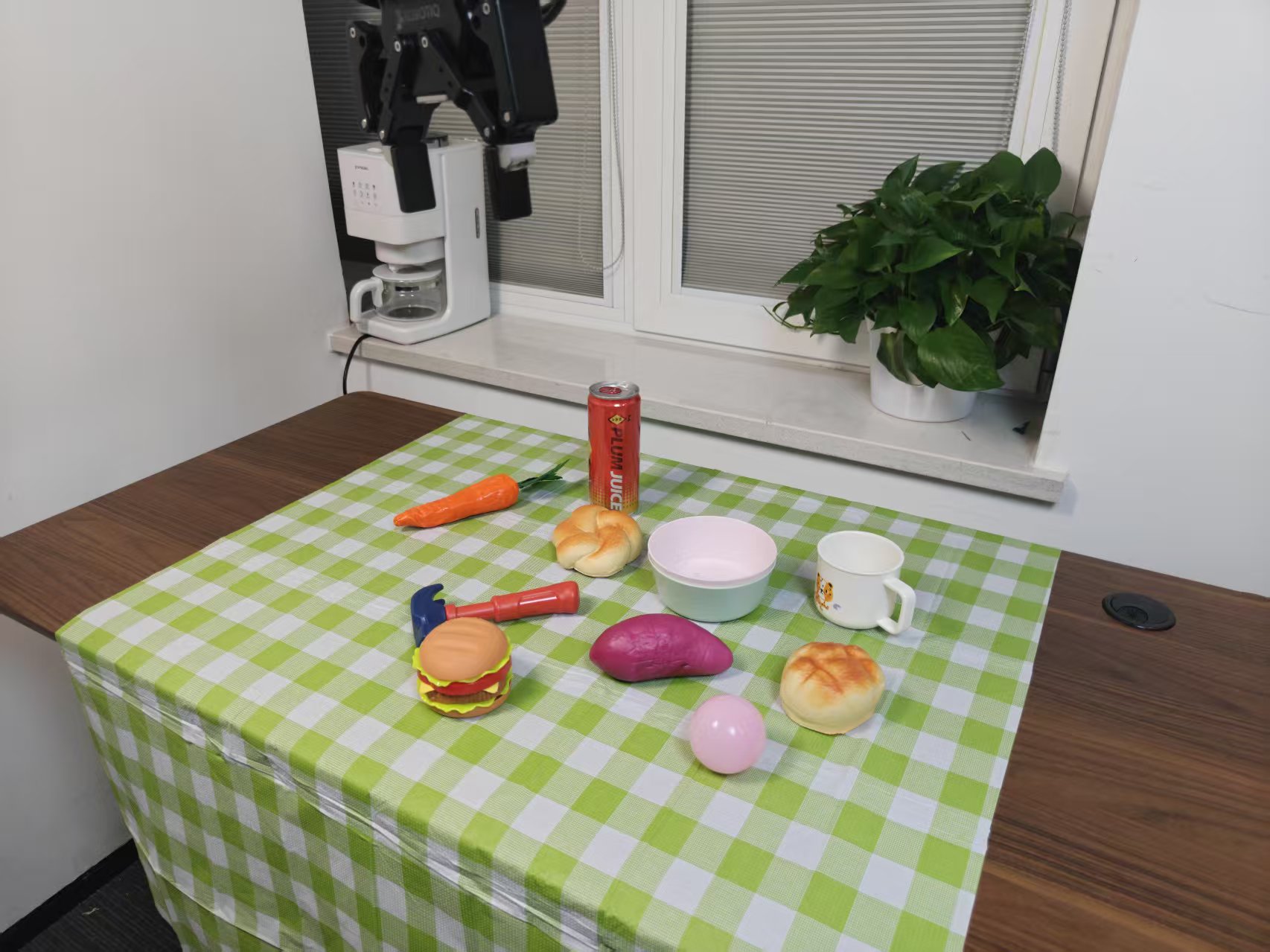}} \\[3pt]
      % \rowcolor{vscaleaccent!6}
      \hspace{1pt}\raisebox{0pt}{\rotatebox[origin=c]{90}{\small Prompt}}\hspace{-1pt} &
      \parbox[c][2.2em][c]{0.205\linewidth}{\centering\small\itshape\shortstack{Open the air-fryer\\drawer.}}
      & \parbox[c][2.2em][c]{0.205\linewidth}{\centering\small\itshape\shortstack{Push down the\\toaster switch.}}
      & \parbox[c][2.2em][c]{0.205\linewidth}{\centering\small\itshape\shortstack{Place two pieces of\\bread on a plate.}}
      & \parbox[c][2.2em][c]{0.205\linewidth}{\centering\small\itshape\shortstack{Stack two bowls.}}
    \end{tabular}
    \caption{\textbf{Perturbed real-robot suite} ($^*$denotes perturbed versions of tasks in the standard suite).}
    \label{fig:real_robot_perturbed_tasks}
  \end{subfigure}

  \vspace{2pt}

  \caption{\textbf{Overview of zero-shot real-robot evaluation suites}, showing initial/target states and instructions. Top: \textbf{Standard suite} (8 tasks) covering spatial grounding, stacking, object transport, articulated objects, and contact-rich control. Bottom: \textbf{Perturbed suite} (4 tasks) incorporating environmental shifts—appliance tasks feature dynamic glare, flashing light, and distractors, while bread/bowl tasks introduce unseen tablecloths and novel distractor objects.}
  \label{fig:real_robot_all_tasks}
\end{figure*}

\paragraph{Task suites.}
The benchmark comprises two suites. The \emph{Standard} suite encompasses eight tasks under controlled laboratory conditions, covering language-conditioned target selection, object transport and stacking, articulated-object interaction, and contact-rich appliance manipulation (Figure~\ref{fig:real_robot_standard_tasks}). The \emph{Perturbed} suite introduces dynamic lighting, background distractors, and novel tabletop appearances across four tasks (Figure~\ref{fig:real_robot_perturbed_tasks}), with specific perturbation details provided in Table~\ref{tab:real_robot_perturbed_conditions}. Detailed task instructions and success criteria are reported in Appendix~\ref{app:real_robot_tasks}.

\paragraph{Rollout and metrics.}
Camera placement, language instructions, scene resets, action decoding, and safety limits are fixed across methods. Each task is evaluated over 10 trials. All tasks use binary success except Toys, which scores three object placements per rollout and therefore contains 30 placement outcomes. Suite results are task-macro averages, so each task contributes equally. Additional rollout details are provided in Appendix~\ref{app:real_robot_protocol}.

\paragraph{Comparison methods.}
As illustrated in Table~\ref{tab:real_robot_results}, we evaluate four pre-training configurations to investigate the impact of progressive data scaling: (1) baseline, (2) multi-embodiment video-action data, (3) multi-embodiment + 60K hours of egocentric video, and (4) multi-embodiment + 120K hours of egocentric video. Across all variants, the Slow System is post-trained on DROID for 60K steps and subsequently frozen, while the Fast System is trained for 160K steps. All variants use eight denoising steps at inference. To isolate the effects of pre-training data, all models share identical DROID data and optimization schedules during post-training (see Appendix~\ref{app:real_robot_training} for implementation details). For external comparison, we evaluate official checkpoints of $\pi_{0.5}$~\cite{pi05} and DreamZero~\cite{ye2026world} post-trained on DROID.

\begin{table}[t]
  \centering
  \small
  \setlength{\tabcolsep}{3.5pt}
  \renewcommand{\arraystretch}{1.18}
  \renewcommand{\tabularxcolumn}[1]{m{#1}}
    \caption{\textbf{Perturbed-suite conditions.} Tasks are grouped by unseen perturbation type, corresponding to Task 9\textasciitilde12 in Figure~\ref{fig:real_robot_perturbed_tasks}.}
  \begin{tabularx}{\linewidth}{@{}
    >{\raggedright\arraybackslash}m{0.38\linewidth}
    >{\raggedright\arraybackslash}m{0.21\linewidth}
    >{\raggedright\arraybackslash}X@{}}
    \toprule
    \textbf{Task Description} & \textbf{Perturbation Type} & \textbf{Evaluation Condition} \\
    \midrule
    \textit{Open the air-fryer drawer;\newline
    Push down the toaster switch} &
    Dynamic Illumination &
    Glare and flashing lights with surrounding distractor objects \\ \midrule
    \textit{ Place two pieces of bread on a plate;\newline
    Stack two bowls} &
    Scene Appearance &
    Tablecloth background with unseen distractor objects \\
    \bottomrule
  \end{tabularx}

  \label{tab:real_robot_perturbed_conditions}
\end{table}

\begin{table*}[t]
  \centering
\caption{\textbf{Zero-shot real-robot success rates (\%).} Results are
  task-level macro averages over the Standard, Perturbed, and combined suites.
  All numeric quantities in the Configuration column are measured in hours.
  The four \name{} configurations are cumulative and exclude their common
  DROID post-training. The Baseline is initialized from Wan2.2-TI2V-5B~\cite{wan2025wan}. We use 8-step inference. Both $\pi_{0.5}$ and DreamZero use their
  official DROID-post-trained checkpoints, matching the target-domain
  post-training data used by our models. $\pi_{0.5}$ additionally uses web-scale multimodal data. Per-task success counts are provided in
  Appendices~\ref{app:standard_suite_results} and~\ref{app:perturbed_suite_results}.}
  \label{tab:real_robot_results}

  \small
  \setlength{\tabcolsep}{4pt}
  \renewcommand{\arraystretch}{1.18}
  \begin{tabular*}{0.92\textwidth}{@{\extracolsep{\fill}}llccc@{}}
    \toprule
    \multirow{2}{*}{\textbf{Policy}} & \multirow{2}{*}{\textbf{Pre-training Data Configuration}} &
    \textbf{Standard} &
    \textbf{Perturbed} &
    \textbf{Average} \\
    & & (8 tasks) & (4 tasks) & (12 tasks) \\
    \midrule
    $\pi_{0.5}$~\cite{pi05} &
    $>$10K Robot Data &
    65.4 & 32.5 & 54.4 \\
    DreamZero~\cite{ye2026world} &
    N/A &
    61.7 & 37.5 & 53.6 \\
    \cmidrule{1-5}
    \multirow{4}{*}{\name{}} &
    Baseline &
    46.7 & 15.0 & 36.1 \\
    & $+$ 6K Multi-Embodiment &
    57.9 & 22.5 & 46.1 \\
    & $+$ 6K Multi-Embodiment $+$ 60K Egocentric Videos &
    82.9 & 35.0 & 66.9 \\
    & $+$ 6K Multi-Embodiment $+$ 120K Egocentric Videos &
    \textbf{87.9} & \textbf{57.5} & \textbf{77.8} \\
    \bottomrule
  \end{tabular*}
  
\end{table*}

\subsubsection{Real-Robot Performance Analysis}
\label{sssec:real_robot_aggregate_results}

\paragraph{Performance overview.}

Table~\ref{tab:real_robot_results} reports suite-level success rates for the four \name{} variants alongside $\pi_{0.5}$~\cite{pi05} and DreamZero~\cite{ye2026world}. Under identical DROID post-training, overall success increases monotonically from 36.1\% for our Baseline configuration to 77.8\% for the full model. Specifically, the full model achieves 87.9\% on the Standard suite and 57.5\% on the Perturbed suite, outperforming $\pi_{0.5}$ and the 14B-parameter DreamZero by a large margin. Detailed per-task breakdowns are provided in Appendices~\ref{app:standard_suite_results} and~\ref{app:perturbed_suite_results}.

\paragraph{Scaling trends.}
Action-labeled multi-embodiment trajectories primarily strengthen executable control. Adding 6K hours of such data raises Standard success from 46.7\% to 57.9\%, Perturbed success from 15.0\% to 22.5\%, and Overall success from 36.1\% to 46.1\%. Because each task is evaluated on a limited number of rollouts, per-task results inherently exhibit sampling noise and are not uniformly positive; we therefore place greater weight on consistent suite-level improvements than on individual task fluctuations. The most pronounced gains occur on contact-rich articulated-object manipulation: microwave closing improves from 0/10 to 9/10, and air-fryer opening from 6/10 to 9/10, indicating that cross-embodiment supervision transfers most clearly to demanding execution behaviors.

Egocentric video primarily improves visual generalization and robustness to observation shifts. Adding 60K hours of such data to the multi-embodiment configuration raises Standard success from 57.9\% to 82.9\%, Perturbed success from 22.5\% to 35.0\%, and Average success from 46.1\% to 66.9\%. The corresponding gains include multi-stage tasks that require tracking visual progress, such as bowl stacking (2/10 to 7/10), bread transfer (4/10 to 10/10), and Toys (16/30 to 28/30). Scaling the video corpus from 60K to 120K hours yields a modest 5.0-point gain on Standard, but a striking 22.5-point jump on Perturbed---with all four Perturbed tasks showing marked improvements under unseen backgrounds, distractors, and dynamic illumination. These gains are consistent with broader priors over visual state evolution and task progress driving robustness under visual shifts.

\begin{figure}[t]
  \centering
  \begin{minipage}[t]{0.485\linewidth}
    \centering
    \IfFileExists{figs/realrobot_case_block_grounding.png}{%
      \includegraphics[width=\linewidth,interpolate=true]{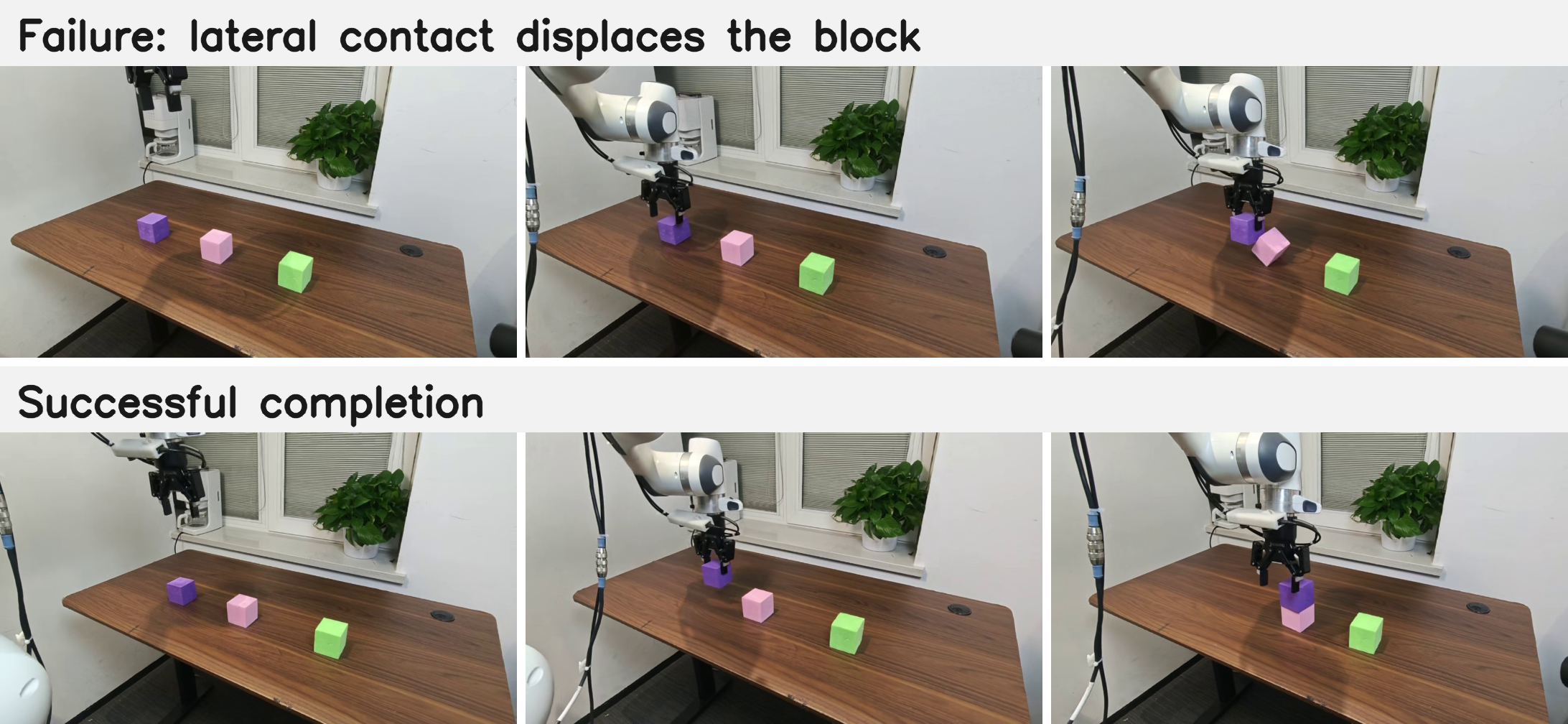}%
    }{%
      \fbox{\parbox[c][0.44\linewidth][c]{0.94\linewidth}{\centering
      \scriptsize Block-stacking comparison\\
      failure/success $\times$ three frames}}%
    }\\[-1pt]
    {\small\textbf{(a) Block-stacking Geometry}}
  \end{minipage}\hfill
  \begin{minipage}[t]{0.485\linewidth}
    \centering
    \IfFileExists{figs/realrobot_case_bowl_sequence.png}{%
      \includegraphics[width=\linewidth,interpolate=true]{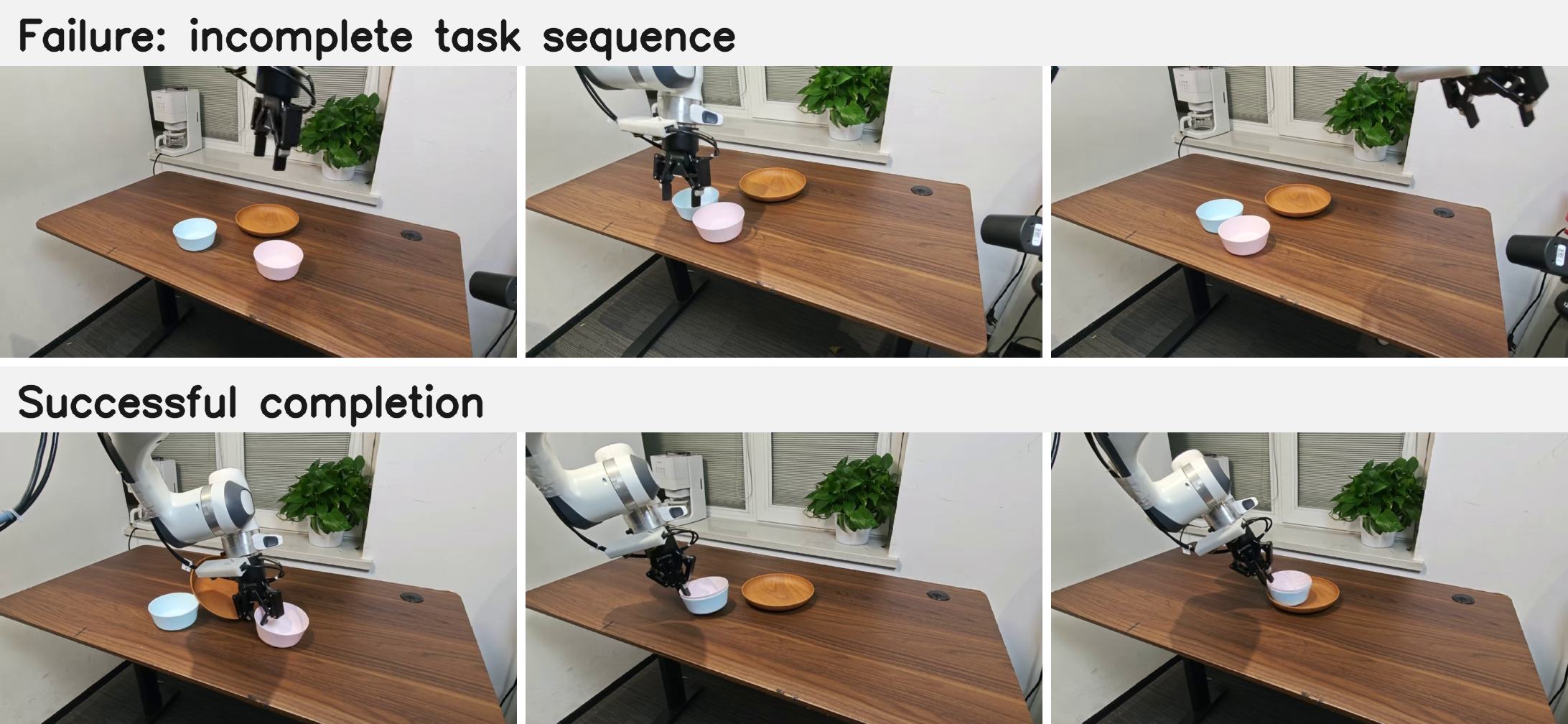}%
    }{%
      \fbox{\parbox[c][0.44\linewidth][c]{0.94\linewidth}{\centering
      \scriptsize Bowl-stacking comparison\\
      two rollouts $\times$ three frames}}%
    }\\[-1pt]
    {\small\textbf{(b) Multi-stage Bowl Stacking}}
  \end{minipage}

  \vspace{0.6em}
  \begin{minipage}[t]{0.485\linewidth}
    \centering
    \IfFileExists{figs/realrobot_case_microwave_contact.png}{%
      \includegraphics[width=\linewidth,interpolate=true]{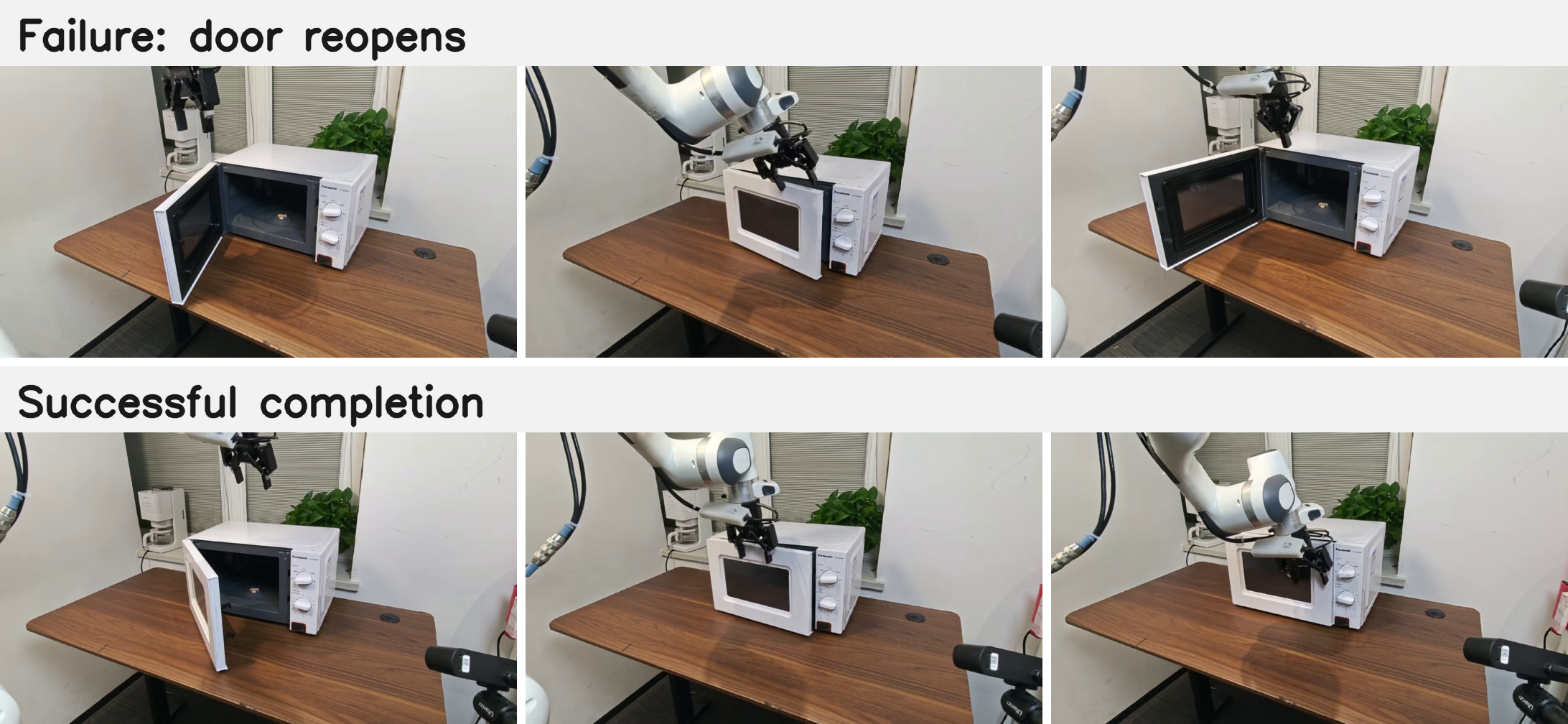}%
    }{%
      \fbox{\parbox[c][0.44\linewidth][c]{0.94\linewidth}{\centering
      \scriptsize Microwave-closing comparison\\
      two rollouts $\times$ three frames}}%
    }\\[-1pt]
    {\small\textbf{(c) Contact-rich Door Closing}}
  \end{minipage}\hfill
  \begin{minipage}[t]{0.485\linewidth}
    \centering
    \IfFileExists{figs/realrobot_case_visual_shift.png}{%
      \includegraphics[width=\linewidth,interpolate=true]{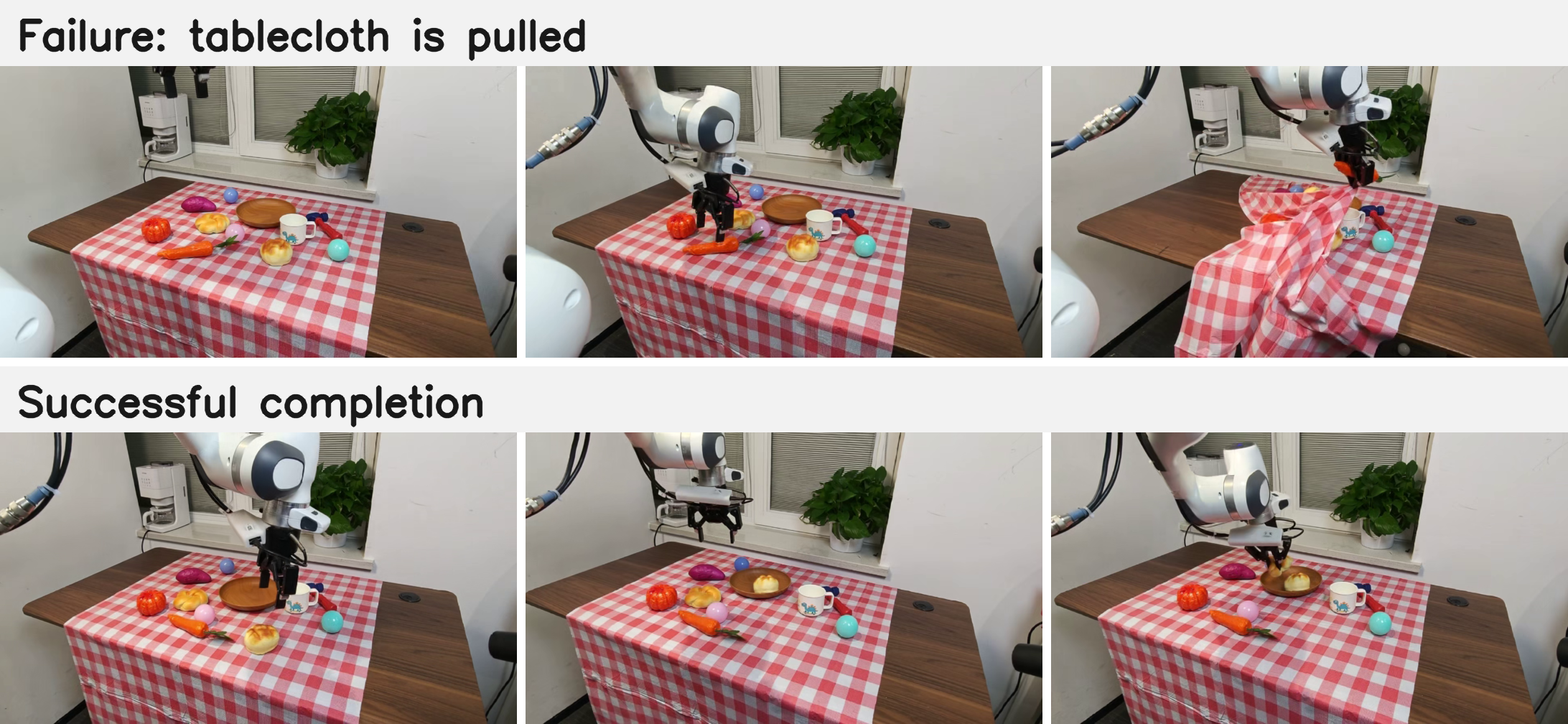}%
    }{%
      \fbox{\parbox[c][0.44\linewidth][c]{0.94\linewidth}{\centering
      \scriptsize Perturbed-scene comparison\\
      two rollouts $\times$ three frames}}%
    }\\[-1pt]
    {\small\textbf{(d) Robustness under Visual Shift}}
  \end{minipage}
  \caption{Representative real-robot rollouts, ordered from left to right.
  Panel (a) contrasts a lateral placement failure with successful top-down
  placement under the same pink-target instruction. Panel (b) contrasts an
  incomplete bowl sequence with placement of the completed stack on the plate.
  Panel (c) contrasts a door that reopens after contact with a complete
  microwave closure. Panel (d) contrasts pulling the tablecloth with completed
  bread placement in the cluttered Perturbed scene.}
  \label{fig:real_robot_rollout_analysis}
\end{figure}

\paragraph{Qualitative analysis.}
Figure~\ref{fig:real_robot_rollout_analysis} illustrates differences in
approach geometry, multi-stage task completion, sustained contact, and
robustness in a cluttered Perturbed scene. The stronger configurations more
often maintain task progress; however, the remaining failures fall into two groups:
failure to advance from a correct intermediate state and local interaction
errors such as target displacement, lost contact, or engagement with the
tablecloth. Indeed, despite reaching 87.9\% on Standard, the full configuration
achieves only 5/10--7/10 on each Perturbed task. These cases indicate that
broader visual priors should be complemented by progress-aware replanning and
responsive corrective control.

\begin{table}[t]
  \centering
  \small
  \setlength{\tabcolsep}{5pt}
  \renewcommand{\arraystretch}{1.12}
  \caption{\textbf{Ablation of inference latency and task performance.}
  Per-task zero-shot success counts and task-macro average success rates are
  reported under the Standard and Perturbed real-robot evaluation settings.
  End-to-end inference latency is reported for the 8-step Slow and Dual-System, as well as 2-step DMD-distilled variants. The DMD-distilled Dual-System includes compilation.}
  \label{tab:dual_system_per_task}

  \begin{tabular}{lllccc}
    \toprule
    \textbf{Setting} & \textbf{Task} & \textbf{Category}
    & \textbf{Slow}
    & \textbf{Dual-System}
    & \textbf{DMD-Distilled} \\
    \midrule

    \multirow{9}{*}{Standard}
      & Bowls     & Stacking               & \textbf{10/10} & 8/10 & 8/10 \\
      & Basket    & Spatial Grounding      & 6/10 & \textbf{10/10} & 9/10 \\
      & Bread     & Object Transfer        & \textbf{10/10} & 9/10 & 9/10 \\
      & Blocks    & Spatial Stacking       & \textbf{9/10} & 7/10 & 8/10 \\
      & Air Fryer & Articulated Object     & 8/10 & \textbf{10/10} & 9/10 \\
      & Toys      & Multi-Object Transport & 23/30 & \textbf{28/30} & 27/30 \\
      & Microwave & Articulated Object     & 6/10 & \textbf{9/10} & \textbf{9/10} \\
      & Toaster   & Contact-Rich Control   & \textbf{8/10} & \textbf{8/10} & 7/10 \\
      \cmidrule(lr){2-6}
      & \textbf{Average (\%)} $\uparrow$
      & 8 tasks
      & 80.8 & \textbf{87.9} & 85.0 \\

    \midrule

    \multirow{5}{*}{Perturbed}
      & Bowls     & Stacking               & \textbf{6/10} & 5/10 & \textbf{6/10} \\
      & Bread     & Object Transfer        & 2/10 & \textbf{7/10} & 5/10 \\
      & Air Fryer & Articulated Object     & 2/10 & 5/10 & \textbf{7/10} \\
      & Toaster   & Contact-Rich Control   & 2/10 & \textbf{6/10} & 4/10 \\
      \cmidrule(lr){2-6}
      & \textbf{Average (\%)} $\uparrow$
      & 4 tasks
      & 30.0 & \textbf{57.5} & 55.0 \\

      \midrule
      \multicolumn{2}{l}{\textbf{Overall Average (\%)} $\uparrow$}
        & 12 tasks
        & 63.9 & \textbf{77.8} & 75.0 \\
    
    \midrule
    \multicolumn{3}{l}{\textbf{Latency (ms)} $\downarrow$}
      & 449.6 & 145.6 & \textbf{33.0} \\
    \bottomrule
  \end{tabular}
\end{table}

\subsection{Zero-shot Task Improvement by Dual-System}
\label{ssec:dual_system_real_robot}

\paragraph{Protocol.}
To isolate the performance gains brought specifically by the Slow--Fast Dual-System architecture, we keep all configurations unchanged and vary only the deployment model architecture, yielding two variants: Slow and Slow--Fast. Both variants are evaluated on the eight Standard and four Perturbed real-robot tasks defined in Section~\ref{ssec:real_robot_scaling}. All policies are post-trained on the same DROID split for 60k optimization steps. The Slow policy operates at a lower closed-loop frequency, generating a 24-step action chunk. In contrast, the dual-system Fast branch runs in real time, conditioning on the latest observation while leveraging the predictive context from the Slow branch. This enables dynamic error correction and real-time contact adjustments without invoking a full world-model rollout.

\paragraph{Results and analysis.}
We report per-task success rates in Table~\ref{tab:dual_system_per_task}, which provide a more diagnostic view of Dual-System behavior. The breakdown separates short-horizon spatial grounding from contact-rich, articulated-object, and multi-object manipulation tasks, where failures often stem from delayed visual correction, inaccurate contact timing, or stale scene-state estimates. This per-task analysis reveals whether Dual-System improves robustness across the held-out suite or mainly raises the average by solving a small subset of easier tasks.

The task-level results indicate that the Dual-System is most effective on tasks that demand frequent visual feedback and recovery from changing scene states. On the Standard suite, it improves Basket, Air Fryer, Toys, and Microwave, spanning spatial grounding, articulated-object interaction, and long-horizon multi-object execution. These gains increase the task-macro average from 80.8\% to 87.9\%. The improvement is more pronounced under the more challenging Perturbed setting, where the task-macro average increases from 30.0\% to 57.5\%, compared with a 7.1 percentage-point gain on the Standard suite. Such settings are particularly sensitive to stale observations and delayed corrections, making them well suited to the Fast branch's online RGB-conditioned updates. Overall, these results highlight the value of the Dual-System for improving closed-loop execution on challenging tasks that require timely visual correction.

\paragraph{Acceleration Ablation.}
We evaluate deployment efficiency by comparing the Slow policy, the asynchronous Dual-System, and the DMD-distilled Dual-System with compilation. As reported in Table~\ref{tab:dual_system_per_task}, asynchronous execution reduces end-to-end inference
latency from 449.6\,ms to 145.6\,ms, corresponding to a $3.1\times$ speedup over the Slow policy. Distillation and compilation further reduce latency to 33.0\,ms, yielding a $4.4\times$ speedup over the undistilled Dual-System and a $13.6\times$ speedup overall. The accelerated model retains a 75.0\% overall task-macro success rate, 2.8 percentage points below the undistilled Dual-System.

\subsection{Benchmark Results}
\label{sec:benchmark_results}
We evaluate ZimaBlue across three representative simulation benchmarks—LIBERO-Plus~\citep{fei2025libero}, RoboTwin-2.0~\citep{chen2025robotwin}, and RoboCasa365~\citep{nasiriany2026robocasa365}—to comprehensively assess key dimensions of generalist manipulation. Specifically, LIBERO-Plus assesses robustness against observation, instruction, and environmental perturbations; RoboTwin 2.0 focuses on bimanual manipulation under domain randomization; and RoboCasa365 evaluates large-scale performance on both seen and unseen household tasks. For simplicity, we conduct all evaluations exclusively on the Slow System.
\subsubsection{LIBERO-Plus}
\label{sec:libero-plus}

LIBERO-Plus~\cite{fei2025libero} assesses policy robustness across seven controlled perturbation dimensions: camera viewpoints, robot initial states, language instructions, lighting conditions, background textures, sensor noise, and object layouts (visualized in Appendix~\ref{app:liberoplus_details}). We evaluate models under two settings: zero-shot transfer—where policies are fine-tuned strictly on standard LIBERO~\cite{liu2023libero}—and supervised fine-tuning (SFT) on the LIBERO-Plus training set. For a fair comparison, baselines are restricted to representative methods with publicly available code/checkpoints or official leaderboard records.

\begin{table}[t]
  \centering
  \small
  \setlength{\tabcolsep}{4.5pt}
  \caption{\textbf{Evaluation results on LIBERO-Plus.} ``Average'' represents the unweighted arithmetic mean of the success rates across all seven perturbation types.Within each evaluation protocol, the best and second-best performances are highlighted in \textbf{bold} and \underline{underlined}, respectively.}
  \label{tab:libero-plus-results}
  \begin{tabular}{l|ccccccc|c}
    \toprule
    \textbf{Model} & \textbf{Camera} & \textbf{Robot} & \textbf{Language} & \textbf{Light} & \textbf{Background} & \textbf{Noise} & \textbf{Layout} & \textbf{Average} \\
    \midrule
    \multicolumn{9}{c}{\textit{\textbf{Zero-shot Transfer}}} \\
    \midrule
    Fast-WAM~\cite{fastwam2026}          & 16.4 & 44.5 & 68.9 & 78.2 & 53.7 & 37.7 & 60.7 & 51.5 \\
    LingBot-VA~\cite{lingbotva2026}      & 40.9 & 83.0 & 86.4 & 82.3 & 53.1 & 64.4 & 76.2 & 69.5 \\
    $\pi_{0.5}$~\cite{pi05}              & 78.4 & 73.6 & 80.8 & 96.2 & 94.1 & 89.0 & 84.5 & 85.2 \\
    ImageWAM-9B~\cite{imagewam2026}      & \underline{79.8} & 58.7 & \textbf{95.2} & 96.1 & 91.2 & \underline{93.3} & 83.1 & 85.3 \\
    ABot-M0.5~\cite{chen2026abot}        & 70.5 & \underline{87.4} & 88.6 & 94.0 & 89.7 & 75.5 & \underline{85.2} & 84.4 \\
    InternVLA-A1.5~\cite{ma2026internvla} & \textbf{83.1} & 55.1 & 86.9 & \underline{96.4} & \textbf{98.2} & \textbf{95.6} & \underline{85.2} & \underline{85.8} \\
    
    \rowcolor{rowblue}
    \textbf{\name{} (Ours)}              & 58.1 & \textbf{88.9} & \underline{91.5} & \textbf{98.0} & 91.5 & 93.1 & \textbf{86.1} & \textbf{86.7} \\
    
    \midrule
    \multicolumn{9}{c}{\textit{\textbf{Supervised Fine-tuning}}} \\
    \midrule
    $\pi_0$~\cite{pi0}                   & 79.6 & 21.1 & 72.5 & 84.7 & 86.2 & 68.3 & 69.4 & 68.8 \\
    GR00T-N1.6~\cite{gr00t}              & 92.6 & 33.5 & 80.1 & 93.6 & 95.4 & 93.6 & 75.0 & 80.5 \\
    OpenVLA-OFT+PT~\cite{openvlaoft}     & 92.8 & 30.3 & \underline{85.8} & 94.9 & 93.9 & 89.3 & 77.6 & 80.7 \\
    ACoT-VLA~\cite{acotvla}              & \textbf{96.6} & 70.4 & 79.7 & 95.1 & \underline{97.1} & \underline{95.9} & \underline{85.0} & 88.5 \\
    CAC-VLA~\cite{cacvla2026}            & 91.2 & \underline{78.4} & 83.3 & \underline{97.5} & \underline{97.1} & 95.4 & \textbf{87.8} & \underline{90.1} \\
    
    \rowcolor{rowblue}
    \textbf{\name{} (Ours)}              & \underline{95.4} & \textbf{81.1} & \textbf{88.8} & \textbf{98.8} & \textbf{99.2} & \textbf{96.5} & 84.3 & \textbf{92.0} \\
    
    \bottomrule
  \end{tabular}
\end{table}

As illustrated in Table~\ref{tab:libero-plus-results}, \name{} achieves an overall success rate of 86.7\% in the zero-shot setting, outperforming InternVLA-A1.5~\cite{ma2026internvla} by 0.9 percentage points. It demonstrates particularly high robustness to initial robot state and lighting variations, though camera viewpoint changes represent a key bottleneck. With task-specific SFT, \name{} improves to 92.0\%, surpassing the second-best method, CAC-VLA~\cite{cacvla2026}, by 1.9 points under identical training protocols. Specifically, it achieves top-ranking performance across five perturbation types (initial states, language instructions, lighting, background textures, and sensor noise) while maintaining competitive accuracy on camera viewpoints and object layouts. Notably, the 5.3-point overall gain from SFT is primarily driven by a dramatic surge in camera viewpoint robustness (from 58.1\% to 95.4\%). This gap indicates that zero-shot viewpoint generalization is constrained by pre-training data diversity, marking a key target for future scaling.

\subsubsection{RoboTwin 2.0}
\label{ssec:benchmark_robotwin}

RoboTwin~2.0~\citep{chen2025robotwin} comprises 50 bimanual manipulation tasks evaluated under two conditions: Clean (standard environments) and Randomized (featuring substantial variations in background, lighting, object placement, and table appearance). Following the multi-task training protocol of recent video-action models~\citep{lingbotva2026,fastwam2026,chen2026abot}, we post-train a single policy model across all 50 tasks using a mixture of clean and randomized demonstrations. A single policy checkpoint is evaluated across all tasks, with success rates computed over 100 trials per task under each condition.

\begin{table}[t]
  \centering
  \small
  \setlength{\tabcolsep}{16pt} % 增大列间距（原为 8pt）
  \renewcommand{\arraystretch}{1.06}
  \caption{\textbf{Evaluation results on the RoboTwin~2.0 benchmark.} Success rates (\%) are averaged over 50 tasks for both ``Clean'' and ``Randomized'' settings, with ``Average'' denoting their mean. The best and second-best results are highlighted in \textbf{bold} and \underline{underlined}.}
  \label{tab:robotwin}
  \begin{tabular}{l|cc|c}
    \toprule
    \textbf{Model} & \textbf{Clean} & \textbf{Randomized} & \textbf{Average} \\
    \midrule
    \multicolumn{4}{c}{\textit{\textbf{Vision-Language-Action Models}}} \\
    \midrule
    $\pi_{0.5}$~\citep{pi05}                        & 82.7 & 76.8 & 79.8 \\
    Qwen-VLA~\citep{wang2026qwen}                  & 86.1 & 87.2 & 86.7 \\
    ABot-M0~\citep{yang2026abot}                   & 86.1 & 85.1 & 85.6 \\
    InternVLA-A1~\citep{internvla_a1}              & 89.4 & 89.6 & 89.5 \\
    StarVLA-$\alpha$~\citep{ye2026starvla}         & 88.7 & 87.8 & 88.3 \\
    % LingBot-VLA$^\dagger$~\citep{xu2026continue}   & 89.9 & 88.8 & 89.3 \\
    % LingBot-VLA+BCP$^\dagger$~\citep{xu2026continue}& \textbf{93.9} & 92.8 & 93.4 \\
    Qwen-RobotManip~\citep{yuan2026qwen}           & \textbf{93.7} & \textbf{94.0} & \textbf{93.9} \\
    InternVLA-A1.5~\citep{ma2026internvla}         & \underline{93.3} & \underline{93.0} & \underline{93.2} \\
    \midrule
    \multicolumn{4}{c}{\textit{\textbf{World Action Models}}} \\
    \midrule
    Motus~\citep{bi2026motus}                       & 88.7 & 87.0 & 87.8 \\
    LingBot-VA~\citep{lingbotva2026}                & 92.9 & 91.6 & 92.2 \\
    Fast-WAM~\citep{fastwam2026}                    & 91.9 & 91.8 & 91.8 \\
    FlowWAM~\citep{chen2026flowwam}                 & 92.9 & 92.1 & 92.5 \\
    LingBot-VA~2.0~\citep{zhang2026native}          & 93.8 & 93.4 & 93.6 \\
    ABot-M0.5~\citep{chen2026abot}                  & \underline{94.0} & \underline{94.2} & \underline{94.1} \\
    
    % ---- 你的模型行 (使用淡蓝底色) ----
    \rowcolor{rowblue}
    \textbf{\name{} (Ours)}                         & \textbf{94.7} & \textbf{94.3} & \textbf{94.5} \\
    \bottomrule
  \end{tabular}
\end{table}

As presented in Table~\ref{tab:robotwin}, \name{} achieves success rates of 94.7\% in the Clean setting and 94.3\% under Randomized evaluation, yielding an Average success rate of 94.5\%. \name{} consistently outperforms all compared VLA and WAM methods across Clean, Randomized, and Average metrics. Notably, compared to the strongest prior WAM baseline, ABot-M0.5~\citep{chen2026abot}, \name{} delivers improvements of 0.7 percentage points (pp) on Clean, 0.1~pp on Randomized, and 0.4~pp on Average. Furthermore, the minimal gap of only 0.4 pp between Clean and Randomized performance demonstrates that \name{} effectively preserves its manipulation capability under substantial visual domain randomization. Detailed per-task success rates are reported in Appendix~\ref{app:robotwin_task_results}.

\subsubsection{RoboCasa365}
\label{ssec:benchmark_robocasa365}

RoboCasa365~\citep{nasiriany2026robocasa365} evaluates generalist robot policies across 365 everyday manipulation tasks and 2,500 diverse kitchen environments. We train our policy on the pre-training Human300 split, and evaluate it under the standard 50-task multi-task protocol. The 50 target tasks comprise 18 Atomic-Seen, 16 Composite-Seen, and 16 Composite-Unseen tasks. Atomic-Seen measures the execution of individual skills; Composite-Seen evaluates long-horizon compositions observed during training, and Composite-Unseen tests zero-shot generalization to held-out composite task templates. Each task is evaluated over 50 rollouts. We report the average success rate for each split along with the trial-weighted Overall score across all 50 tasks.

\begin{table}[t]
  \centering
  \small
  \setlength{\tabcolsep}{8pt}
  \renewcommand{\arraystretch}{1.06}
  \caption{\textbf{Evaluation on RoboCasa365~\citep{nasiriany2026robocasa365} benchmark.} ``Average'' denotes success rates (\%) across three task. The best and second-best results are highlighted in \textbf{bold} and \underline{underlined}. $^*$ denotes results reported in the original paper; all other baseline results are taken from the official RoboCasa365 leaderboard~\citep{robocasa365_leaderboard}.}
  \label{tab:robocasa365}
  \begin{tabular}{l|ccc|c}
    \toprule
    \textbf{Model} & \textbf{Atomic-Seen} & \textbf{Composite-Seen} & \textbf{Composite-Unseen} & \textbf{Average} \\
    \midrule
    \multicolumn{5}{c}{\textit{\textbf{Vision-Language-Action Models}}} \\
    \midrule
    Diffusion Policy~\citep{chi2025diffusion}   & 15.7 & 0.2  & 1.3  & 6.1  \\
    $\pi_0$~\citep{pi0}                         & 34.6 & 6.1  & 1.1  & 14.8 \\
    $\pi_{0.5}$~\citep{pi05}                    & 39.6 & 7.1  & 1.2  & 16.9 \\
    GR00T-N1.5~\citep{gr00t}                  & 50.7 & 14.8 & 2.7  & 23.9 \\
    GR00T-N1.6~\citep{gr00t}                  & 51.1 & 9.4  & 1.7  & 21.9 \\
    Qwen-RobotManip$^*$~\citep{yuan2026qwen}        & 68.6 & 20.1 & \underline{14.9} & 35.9 \\
    RLDX-1~\citep{rldx1_2026}                   & 67.6 & \underline{27.9} & 8.5  & \underline{36.0} \\
    Xiaomi-Robotics-1~\citep{xiaomi_robotics_1} & \textbf{80.2} & \textbf{57.1} & \textbf{32.1} & \textbf{57.4} \\
    \midrule
    \multicolumn{5}{c}{\textit{\textbf{World Action Models}}} \\
    \midrule
    GigaWorld-Policy~0.1~\citep{ye2026gigaworld} & 44.4 & 11.8 & 2.9  & 20.7 \\
    WorldDreamer~\citep{worlddreamer2024}       & 66.3 & 26.7 & 9.0  & 35.3 \\
    ABot-M0.5~\citep{chen2026abot}              & 75.6 & 37.7 & 3.3  & 40.3 \\
    ABot-M0.6~\citep{robocasa365_leaderboard}              & \textbf{79.4} & \underline{48.3} & \underline{7.9}  & \underline{46.6} \\
    
    % ---- 你的模型行 (使用淡蓝底色) ----
    \rowcolor{rowblue}
    \textbf{\name{} (Ours)}                     & \underline{78.1} & \textbf{50.4} & \textbf{16.5} & \textbf{49.5} \\
    \bottomrule
  \end{tabular}
\end{table}

As shown in Table~\ref{tab:robocasa365}, \name{} achieves the highest Average, Composite-Seen, and Composite-Unseen success rates among all WAM methods, ranking second only to the VLA-based Xiaomi-Robotics-1~\citep{xiaomi_robotics_1}. Notably, while Xiaomi-Robotics-1 relies on 100,000 hours of real-world robot data, \name{} attains highly competitive performance using only 6,000 hours action-labeled robot trajectories. Compared to ABot-M0.6~\citep{robocasa365_leaderboard}, \name{} boosts Average success by 2.9 pp and Composite-Seen by 2.1 pp. Crucially, on Composite-Unseen tasks, \name{} more than doubles the success rate of ABot-M0.6 from 7.9\% to 16.5\%. This substantial advantage on unseen long-horizon tasks validates the effectiveness and generalization capabilities of our pre-training scheme. Detailed per-task success rates for all 50 evaluation tasks are reported in
Appendix~\ref{app:robocasa_task_results}.

\begin{figure*}[t]
\centering
\includegraphics[width=1.0\textwidth]{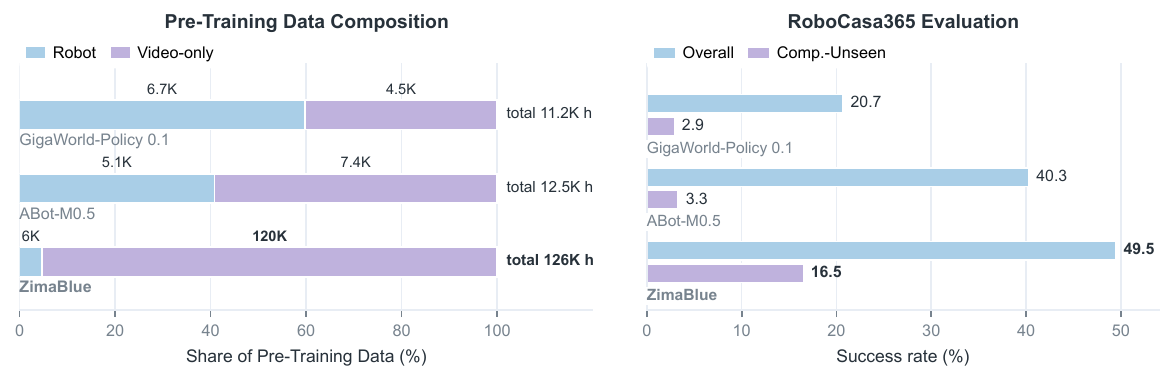}
\caption{\textbf{Pre-training Data Composition and Robocasa365 Performance.} \textbf{Left:} Comparison of pre-training data volume and ratio between robot trajectories and video-only data across different WAMs. \textbf{Right:} The results on RoboCasa365 demonstrate a direct positive correlation between pre-training video scale and success rates, particularly on Composite-Unseen tasks.}
\label{fig:robocasa_video_pretraining}
\end{figure*}

To evaluate the efficacy of video pre-training, we compare the pre-training data compositions and evaluation performance across different WAMs in Figure~\ref{fig:robocasa_video_pretraining}. All three methods utilize a comparable scale of real-robot demonstration data, ranging from 5.1K to 6.7K hours. However, \name{} incorporates 120K hours of video-only data—over 16 times that of ABot-M0.5~\citep{chen2026abot} and 26 times that of GigaWorld-Policy 0.1~\citep{ye2026gigaworld}. This massive scale-up in action-free video pre-training translates directly to substantial performance gains, boosting the Composite-Unseen success rate to 16.5\% (versus 3.3\% and 2.9\%). These results highlight the pivotal role of large-scale video pre-training in enabling strong generalizability to unseen scenes.

\begin{figure}[!t]
  \centering
  \includegraphics[width=0.95\textwidth]
  {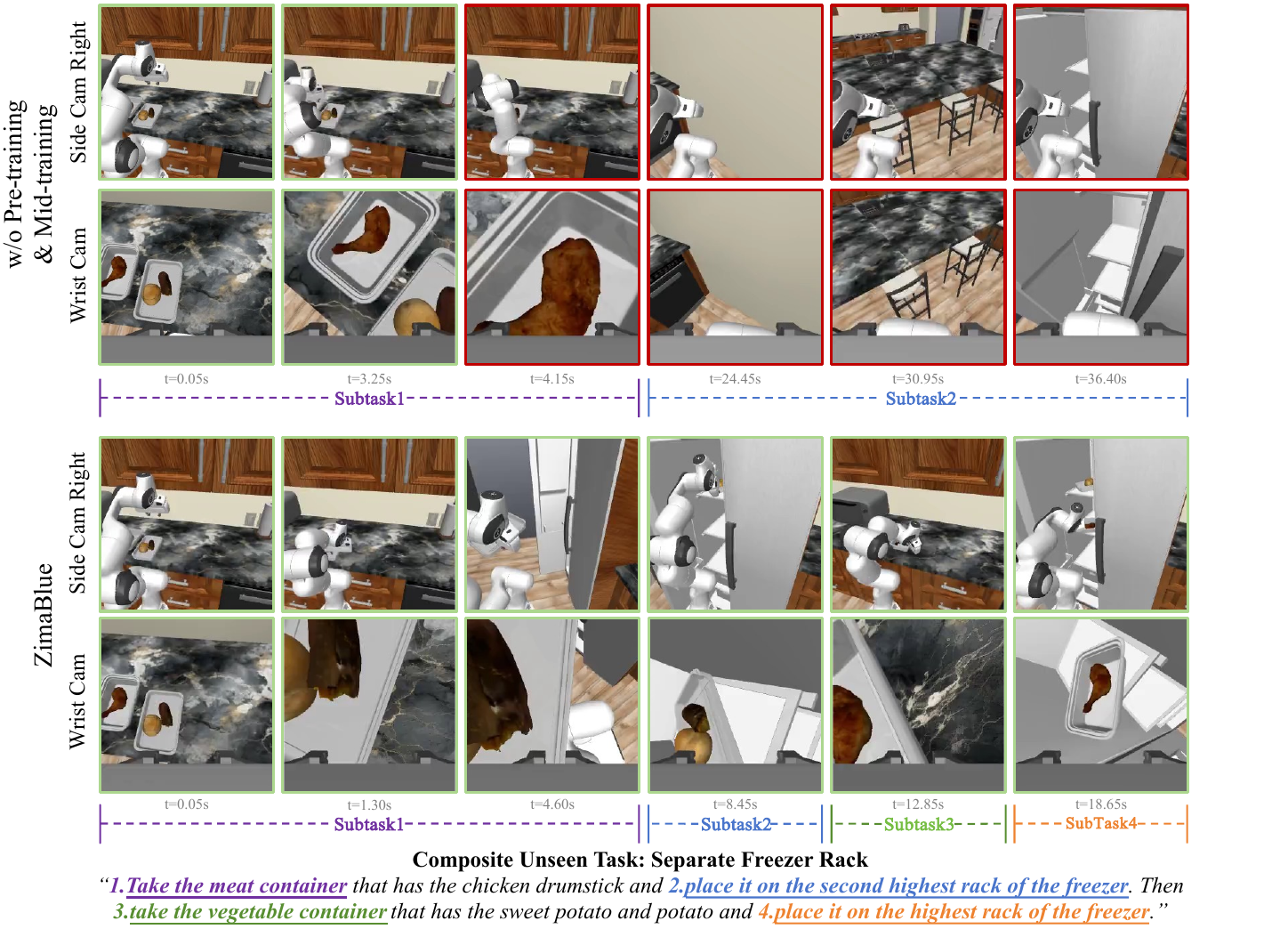}
  \vspace{0.1in}
  \caption{\textbf{Qualitative comparison on a Robocasa365~\citep{nasiriany2026robocasa365} Composite-Unseen task.} Green boxes highlight instruction-consistent actions, whereas red boxes denote execution errors. \name{} successfully places both containers onto the designated freezer racks. In contrast, the ablated policy targets an incorrect object and prematurely proceeds to the next stage despite a failed grasp.}
  \label{fig:ablation_unseen_freezer}
  \vspace{0.1in}
\end{figure}

To qualitatively illustrate the impact of video pre-training, Figure~\ref{fig:ablation_unseen_freezer} compares execution trajectories on a complex, long-horizon task. The task requires grounding two relational object descriptions alongside two ordinal spatial targets. \name{} accurately identifies the container holding the chicken drumstick, places it onto the second-highest freezer rack, and subsequently places the vegetable container onto the top rack. In contrast, the ablated policy without video pre-training mistakenly targets the chicken drumstick itself rather than its container. Even after failing the grasp, it prematurely advances to the placement stage, demonstrating a compound failure of referential grounding and subtask execution verification.

\section{Conclusion and Future Work}
\label{sec:conclusion}

We present \name{}, a scalable World Action Model that positions embodied video as a primary axis for scaling generalizable robot learning. Through a structured pipeline—large-scale egocentric pre-training, cross-embodiment video-action alignment, and target-robot post-training—\name{} acquires robust physical and dynamic priors that boost task and environmental generalization. To make these large generative priors practical, our Slow-Fast dual system decouples high-level world reasoning from real-time closed-loop control. Zero-shot real-robot evaluations and simulation benchmark results demonstrate that \name{} offers a viable paradigm for scaling embodied intelligence: \textit{learning broadly from videos, and acting efficiently via decoupled reasoning and control.}

The 120,000 hours of video in this work offer merely a glimpse into the full potential of scaling for robotics. We aim to push this frontier forward along four key directions. First and most importantly, expanding evaluation is key. Developing diverse, multi-embodiment benchmarks across rich task suites will allow us to thoroughly stress-test and quantify the model's true capabilities. Second, we will scale up pre-training across both data and model dimensions. Beyond incorporating vastly larger and more diverse egocentric human video, expanding model capacity will be essential to building a truly generalist, physics-aware foundation model. Third, endowing the system with stronger reasoning capabilities—such as through higher-level reasoning modules—can unlock complex, long-horizon tasks by enabling hierarchical planning and dynamic self-correction. Finally, bridging the gap toward rapid skill acquisition in novel environments calls for in-context learning. By treating physical demonstrations directly as prompt inputs, the robot can internalize and execute new behaviors at inference time—entirely without gradient updates.

\newpage
\section{Authors}
\label{sec:authors}

\noindent\textbf{Core Contributors}

\noindent Xionghao Wu$^{1,3,4,7}$, Yijun Yang$^{1,2,3,4}$, Shiyang Zhou$^{1,3,4}$, Haoze Sun$^{1,4,5,7}$, Jianhui Liu$^{4,5,7}$, Songsong Yu$^{4,6,7}$, Jiyao Zhang$^{8}$, Wenbo Li$^{9}$
\renewcommand{\thefootnote}{\corrauthormark}
\footnote{Corresponding author: fenglinglwb@gmail.com}
\renewcommand{\thefootnote}{\arabic{footnote}}

\vspace{0.05in}

\noindent{\small\itshape Listed by task with no indication of priority: $^{1}$Data, $^{2}$Pre-training, $^{3}$
Mid-training, $^{4}$Post-training, $^{5}$Dual-System, $^{6}$Acceleration, $^{7}$Deployment \& Demo, $^{8}$Guidance, $^{9}$Project Lead.}
\vspace{0.15in}

\noindent\textbf{Contributors}

\noindent Bo Wang, Guoqing Ma, Lin Song, Renjie Liao, Shenghe Zheng, Wei Tang, Xiaojuan Qi, Yanwei Li, Yuan Zhang, Zhuotao Tian

\vspace{0.05in}

\noindent{\small\itshape Listed in alphabetical order.}
\vspace{0.15in}

\noindent\textbf{Supervision}

\noindent Haoyang Huang, Nan Duan

\clearpage
\bibliographystyle{unsrtnat}
\bibliography{main}

\clearpage
\appendix
\section{Real-Robot Evaluation Details}
\label{app:real_robot_evaluation}

This section supplements the aggregate real-robot results in
Section~\ref{ssec:real_robot_scaling} and
Section~\ref{ssec:dual_system_real_robot} with task-level definitions and
evaluation conventions.

\subsection{Task Definitions and Success Criteria}
\label{app:real_robot_tasks}

All 12 tasks are held out from DROID post-training. Except for the per-object
scoring used in the Toys task, a rollout is successful only when the commanded
outcome is achieved at the end of the episode.

\medskip
\noindent\textbf{Standard suite.}\par
\begin{list}{\labelitemi}{%
  \setlength{\labelwidth}{0.5em}%
  \setlength{\labelsep}{0.45em}%
  \setlength{\leftmargin}{1.70em}%
  \setlength{\itemsep}{0pt}
  \setlength{\parsep}{0pt}
  \setlength{\topsep}{0pt}
  \setlength{\partopsep}{0pt}
  \setlength{\rightmargin}{0pt}}
  \item \textbf{\textit{Stack bowls and place them on a plate.}}
  Both bowls must remain stably stacked on the plate.
  \item \textbf{\textit{Pick up the instructed left or right cup.}}
  The cup on the instructed side must be lifted without being knocked over;
  picking up the cup on the other side is a failure.
  \item \textbf{\textit{Move bread from a toaster to a plate.}}
  The bread must be removed from the toaster and left resting on the plate.
  \item \textbf{\textit{Stack blocks.}}
  The purple block must remain stably on the instructed green or pink block.
  \item \textbf{\textit{Open an air-fryer drawer.}}
  The drawer must be moved from its initial closed state to the open state.
  \item \textbf{\textit{Put toys into a box.}}
  Each of the three target toys is scored separately and counts as successful
  when it is placed inside the blue box.
  \item \textbf{\textit{Close a microwave.}}
  The microwave door must be moved from its initial open state to the closed
  state.
  \item \textbf{\textit{Push down a toaster switch.}}
  The toaster switch must reach its activated position.
\end{list}

\medskip
\noindent\textbf{Visually Perturbed suite.}\par
\begin{list}{\labelitemi}{%
  \setlength{\labelwidth}{0.5em}%
  \setlength{\labelsep}{0.45em}%
  \setlength{\leftmargin}{1.70em}%
  \setlength{\itemsep}{0pt}
  \setlength{\parsep}{0pt}
  \setlength{\topsep}{0pt}
  \setlength{\partopsep}{0pt}
  \setlength{\rightmargin}{0pt}}
  \item \textbf{\textit{Open an air-fryer drawer.}}
  The drawer must be opened under glare and flashing illumination with
  surrounding distractor objects.
  \item \textbf{\textit{Push down a toaster switch.}}
  The switch must reach its activated position under glare and flashing
  illumination with surrounding distractor objects.
  \item \textbf{\textit{Place two pieces of bread on a plate.}}
  Both pieces of bread must rest on the plate under a tablecloth background and
  unseen-object clutter.
  \item \textbf{\textit{Stack two bowls.}}
  The two bowls must remain stably stacked under a tablecloth background and
  unseen-object clutter.
\end{list}

\subsection{Rollout Protocol and Metrics}
\label{app:real_robot_protocol}

All methods use the same robot, camera placement, language instruction, scene
reset, action decoder, and safety limits. Each policy receives two external RGB
views, one wrist view, and the current proprioceptive state. It predicts a
24-step action chunk and obtains fresh observations before the next closed-loop
update. Each task uses 10 rollouts. For seven Standard tasks and all four
Perturbed tasks, every rollout has a binary success outcome. The Toys task
contains three object placements per rollout; each placement is scored
separately, giving 30 binary placement outcomes. We first compute one success
rate per task, using 10 rollout outcomes for the other tasks and 30 placement
outcomes for Toys. The Standard, Perturbed, and Overall scores are then macro
averages over 8, 4, and 12 task-level rates, respectively. Thus, each task has
equal weight in its suite despite the finer-grained scoring used for Toys.

\subsection{DROID Adaptation Details}
\label{app:real_robot_training}

Each \name{} variant uses the same two-stage target-robot adaptation. The Slow
System is first post-trained on DROID for 60k steps. It is then frozen while the
Fast System is trained for 160k steps. Table~\ref{tab:droid_adaptation_recipe}
summarizes the optimization settings shared by the four pre-training
initializations.

\begin{table}[!htbp]
  \centering
  \small
  \setlength{\tabcolsep}{5pt}
  \renewcommand{\arraystretch}{1.12}
  \caption{\textbf{DROID adaptation settings.} The Slow System is frozen
  throughout Fast System training.}
  \label{tab:droid_adaptation_recipe}
  \begin{tabular}{lccccc}
    \toprule
    \textbf{Training stage} & \textbf{Steps} & \textbf{Global batch} &
    \textbf{Learning rate} & \textbf{Warmup} & \textbf{Action horizon} \\
    \midrule
    Slow System post-training & 60k & 256 & $1\times10^{-4}$ & 2\% & 24 \\
    Fast System training & 160k & 64 & $1\times10^{-4}$ & 1\% & 24 \\
    \bottomrule
  \end{tabular}
\end{table}

\subsection{Standard Suite Results}
\label{app:standard_suite_results}

The four \name{} variants use, respectively, no cross-embodiment pre-training,
multi-embodiment video-action pre-training, video pre-training followed by
video-action pre-training, and scaled video pre-training followed by
video-action pre-training. All four use the same DROID adaptation described in
Appendix~\ref{app:real_robot_training}. Table~\ref{tab:standard_suite_breakdown}
reports successful rollouts for every Standard task and evaluation variant.

\begin{table}[!htbp]
  \centering
  \small
  \setlength{\tabcolsep}{2.5pt}
  \renewcommand{\arraystretch}{1.12}
  \renewcommand{\tabularxcolumn}[1]{m{#1}}
  \caption{\textbf{Standard-suite evaluation details.} Results for the two
  baselines and four \name{} variants. Each entry reports successful evaluation
  units over attempts: rollouts for seven tasks and individual object placements
  for Toys. Here, VA means multi-embodiment video-action pre-training; V$+$VA
  means video pre-training followed by VA; and SV$+$VA means scaled video
  pre-training followed by VA. The macro averages correspond to the Standard
  scores reported in Table~\ref{tab:real_robot_results}.}
  \label{tab:standard_suite_breakdown}
  \begin{tabularx}{\linewidth}{@{}
    >{\raggedright\arraybackslash}m{0.255\linewidth}
    >{\raggedright\arraybackslash}X
    *{5}{>{\centering\arraybackslash}m{0.059\linewidth}}
    >{\centering\arraybackslash}m{0.065\linewidth}@{}}
    \toprule
    \multicolumn{2}{c}{\textbf{Evaluation details}} &
    \multicolumn{2}{c}{\textbf{Released policies}} &
    \multicolumn{4}{c}{\textbf{\name{} initialization}} \\
    \cmidrule(r){1-2}\cmidrule(lr){3-4}\cmidrule(l){5-8}
    \textbf{Task} & \textbf{Evaluation setup} & \textbf{$\pi_{0.5}$} &
    \textbf{\shortstack{Dream\\Zero}} & \textbf{\shortstack{From\\scratch}} &
    \textbf{VA} & \textbf{V$+$VA} & \textbf{SV$+$VA} \\
    \midrule
    \textit{Stack bowls and place on a plate} & pink on blue; blue on pink
      & 5/10 & 6/10 & 2/10 & 2/10 & 7/10 & 8/10 \\
    \textit{Pick up the instructed cup} & left cup; right cup
      & 10/10 & 7/10 & 3/10 & 5/10 & 8/10 & 10/10 \\
    \textit{Push down the toaster switch} & push down
      & 5/10 & 6/10 & 9/10 & 8/10 & 9/10 & 8/10 \\
    \textit{Close the microwave door} & initial door angle $75^\circ$
      & 10/10 & 7/10 & 0/10 & 9/10 & 8/10 & 9/10 \\
    \textit{Put all toys into the blue box} & random layout; 1000-step horizon
      & 25/30 & 13/30 & 19/30 & 16/30 & 28/30 & 28/30 \\
    \textit{Move bread from the toaster to a plate} & left-to-right; right-to-left
      & 7/10 & 2/10 & 4/10 & 4/10 & 10/10 & 9/10 \\
    \textit{Open the air-fryer drawer} & pull the drawer open
      & 0/10 & 10/10 & 6/10 & 9/10 & 6/10 & 10/10 \\
    \textit{Stack the purple cube} & on pink; on green
      & 7/10 & 7/10 & 7/10 & 4/10 & 9/10 & 7/10 \\
    \midrule
    \rowcolor{rowblue}
    \textbf{Macro average (\%)} & \textbf{8 tasks}
      & \textbf{65.4} & \textbf{61.7} & \textbf{46.7} & \textbf{57.9} & \textbf{82.9} & \textbf{87.9} \\
    \bottomrule
  \end{tabularx}
\end{table}

\subsection{Visually Perturbed Suite Results}
\label{app:perturbed_suite_results}

Table~\ref{tab:perturbed_suite_breakdown} reports results under controlled
visual shifts. The bread and bowl tasks combine an unseen scene, a cluttered
background, and unseen distractor objects. The appliance tasks introduce
dynamic illumination and background clutter; the toaster is also evaluated at
a different location.

\begin{table}[!htbp]
  \centering
  \small
  \setlength{\tabcolsep}{2.5pt}
  \renewcommand{\arraystretch}{1.12}
  \renewcommand{\tabularxcolumn}[1]{m{#1}}
  \caption{\textbf{Visually Perturbed-suite evaluation details.} Results for
  $\pi_{0.5}$, DreamZero, and the four \name{} variants. Each entry reports
  successful rollouts over ten attempts. Here, VA means multi-embodiment
  video-action pre-training; V$+$VA means video pre-training followed by VA;
  and SV$+$VA means scaled video pre-training followed by VA.}
  \label{tab:perturbed_suite_breakdown}
  \begin{tabularx}{\linewidth}{@{}
    >{\raggedright\arraybackslash}m{0.255\linewidth}
    >{\raggedright\arraybackslash}X
    *{5}{>{\centering\arraybackslash}m{0.059\linewidth}}
    >{\centering\arraybackslash}m{0.065\linewidth}@{}}
    \toprule
    \multicolumn{2}{c}{\textbf{Evaluation details}} &
    \multicolumn{2}{c}{\textbf{Released policies}} &
    \multicolumn{4}{c}{\textbf{\name{} initialization}} \\
    \cmidrule(r){1-2}\cmidrule(lr){3-4}\cmidrule(l){5-8}
    \textbf{Task} & \textbf{Visual shift} & \textbf{$\pi_{0.5}$} &
    \textbf{\shortstack{Dream\\Zero}} & \textbf{\shortstack{From\\scratch}} &
    \textbf{VA} & \textbf{V$+$VA} & \textbf{SV$+$VA} \\
    \midrule
    \textit{Place two pieces of bread on a plate} & unseen scene; cluttered background; unseen objects
      & 4/10 & 0/10 & 1/10 & 3/10 & 5/10 & 7/10 \\
    \textit{Stack two bowls} & unseen scene; cluttered background; unseen objects
      & 6/10 & 3/10 & 1/10 & 1/10 & 3/10 & 5/10 \\
    \textit{Open the air-fryer drawer} & dynamic illumination; background clutter
      & 1/10 & 7/10 & 3/10 & 5/10 & 3/10 & 5/10 \\
    \textit{Push down the toaster switch} & dynamic illumination; different location
      & 2/10 & 5/10 & 1/10 & 0/10 & 3/10 & 6/10 \\
    \midrule
    \rowcolor{rowblue}
    \textbf{Macro average (\%)} & \textbf{4 tasks}
      & \textbf{32.5} & \textbf{37.5} & \textbf{15.0} & \textbf{22.5} &
      \textbf{35.0} & \textbf{57.5} \\
    \bottomrule
  \end{tabularx}
\end{table}

\subsection{Perturbation Controls and Failure Analysis}
\label{app:real_robot_ood}

The appliance tasks vary illumination through glare and flashing lights and add
nearby objects as visual distractors. The bread and bowl tasks replace the
tabletop appearance with a tablecloth and introduce many unseen surrounding
objects. The Visually Perturbed score is reported separately from the Standard score to
isolate robustness under these controlled shifts. Failure analysis distinguishes
instruction or perception errors, grasp and contact failures, accumulated pose
error, and failures to recover after partial execution errors.

\Needspace{6\baselineskip}
\section{Additional Simulation Experiment Details}
\label{app:simulation_details}

This appendix provides supplementary experimental details and analyses for the
simulation benchmarks evaluated in Section~\ref{sec:benchmark_results}.

\subsection{LIBERO-Plus Detailed Analysis}
\label{app:liberoplus_details}

\paragraph{Evaluation scope and aggregation.}
We evaluate both protocols on the same 10,030 episodes covering four suites and
seven perturbation categories. Category scores pool episodes across suites, and
their unweighted mean gives the category-macro result. For Robot Initial States,
we follow the official protocol without restoring the perturbed configuration.

\begin{table}[!htbp]
  \centering
  \small
  \setlength{\tabcolsep}{4.5pt}
  \renewcommand{\arraystretch}{1.12}
  \caption{\textbf{\name{} performance across the four underlying LIBERO task
  suites.} Values are episode-level success rates (\%); Overall is weighted by the number of episodes and therefore
  differs from the category-macro score in
  Table~\ref{tab:libero-plus-results}.}
  \label{tab:libero-plus-suites}
  \begin{tabular}{l|cccc|c}
    \toprule
    \textbf{Protocol} & \textbf{Spatial} & \textbf{Object} & \textbf{Goal} &
    \textbf{LIBERO-10} & \textbf{Overall} \\
    \midrule
    Zero-shot & 87.8 & 88.4 & 81.4 & 86.5 & 86.0 \\
    SFT & 92.8 & 92.1 & 87.1 & 94.0 & 91.5 \\
    $\Delta$ & +5.0 & +3.7 & +5.8 & +7.5 & +5.5 \\
    \bottomrule
  \end{tabular}
\end{table}

Table~\ref{tab:libero-plus-suites} shows that SFT improves all four suites
(+3.7 to +7.5 points), with the largest gain on LIBERO-10. Goal remains the
hardest suite at 87.1\%, so aggregate robustness gains do not eliminate the
difficulty of goal-conditioned task execution.

\begin{table}[!htbp]
  \centering
  \small
  \setlength{\tabcolsep}{4.5pt}
  \renewcommand{\arraystretch}{1.12}
  \caption{\textbf{Fine-grained effect of LIBERO-Plus supervision.} Each entry
  is the SFT success rate minus the zero-shot success rate in percentage
  points, computed over the same episodes. Positive values denote gains.}
  \label{tab:libero-plus-suite-category-delta}
  \begin{tabular}{l|ccccccc}
    \toprule
    \textbf{Suite} & \textbf{Camera} & \textbf{Robot} & \textbf{Language} &
    \textbf{Light} & \textbf{Background} & \textbf{Noise} & \textbf{Layout} \\
    \midrule
    Spatial & +42.3 & -6.9 & -1.8 & -1.0 & +20.2 & -5.4 & -9.6 \\
    Object & +28.8 & -16.8 & -3.1 & +1.0 & +2.4 & +7.1 & +4.5 \\
    Goal & +32.1 & -1.5 & -5.6 & +3.2 & +3.6 & +7.1 & +0.5 \\
    LIBERO-10 & +45.8 & -6.1 & 0.0 & 0.0 & +4.8 & +3.6 & -3.2 \\
    \bottomrule
  \end{tabular}
\end{table}

Table~\ref{tab:libero-plus-suite-category-delta} shows consistent camera gains
across all suites, largest on LIBERO-10 (+45.8 points), and background gains
throughout. Robot Initial States decreases in every suite, most on Object
(-16.8 points), while Layout improves on Object and Goal but declines on Spatial
and LIBERO-10. These trade-offs explain how SFT raises every suite-level
aggregate despite several category-level regressions. Figure~\ref{fig:liberoplus_perturbations_all}
shows initial observations for all seven perturbations.

\noindent\begin{minipage}{\linewidth}
  \centering
  \includegraphics[width=0.66\linewidth]{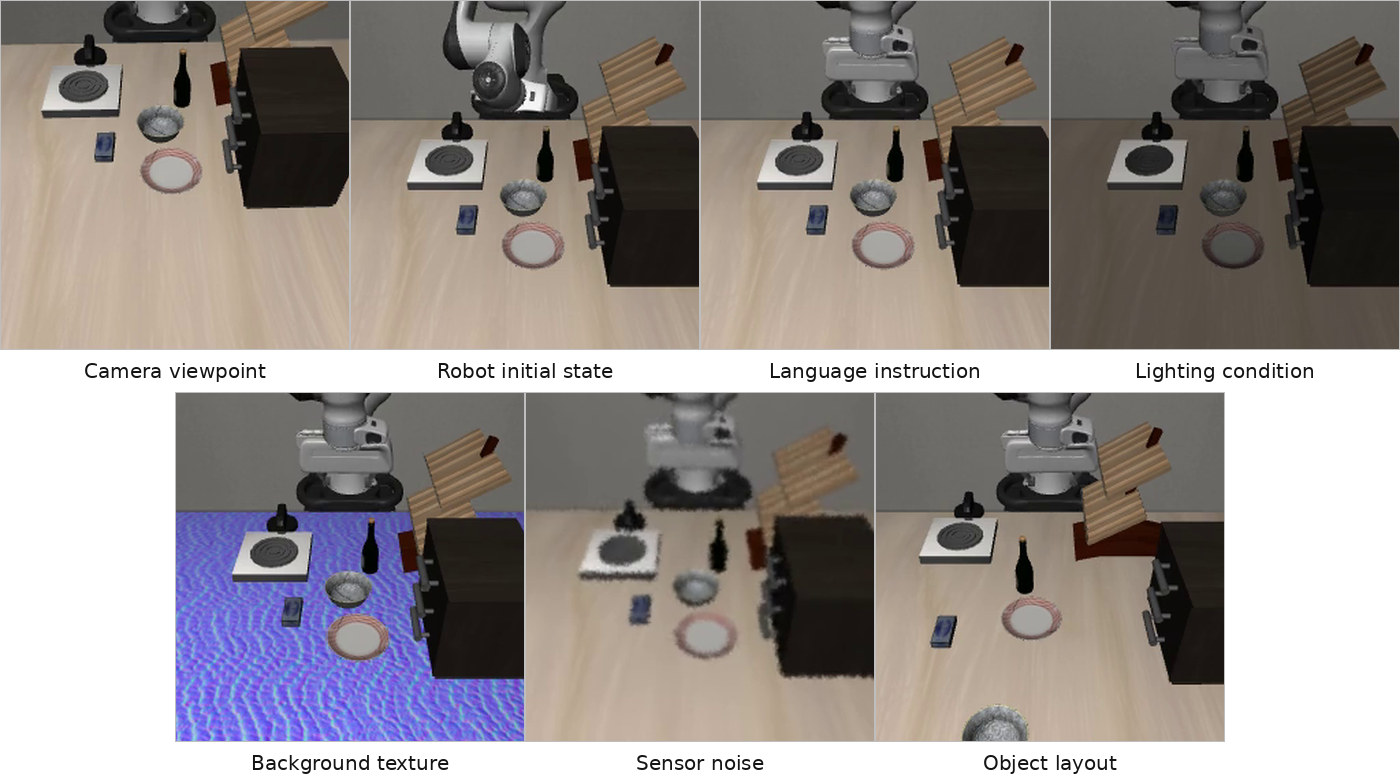}
  \captionsetup{hypcap=false}
  \captionof{figure}{\textbf{The seven LIBERO-Plus perturbation categories.}
  Initial observations are shown for the same task in our official no-restore
  SFT evaluation. Language perturbations modify the instruction; quantitative results use all episodes
  (Table~\ref{tab:libero-plus-results}).}
  \label{fig:liberoplus_perturbations_all}
\end{minipage}

\subsection{RoboTwin 2.0 Per-Task Results}
\label{app:robotwin_task_results}
Detailed per-task success rates on RoboTwin~2.0 are reported in
Table~\ref{tab:robotwin_task_results}.
\begingroup
\small
\setlength{\tabcolsep}{4pt}
\renewcommand{\arraystretch}{1.04}
\begin{longtable}{@{}>{\raggedright\arraybackslash}p{0.40\linewidth}*{3}{>{\centering\arraybackslash}p{0.16\linewidth}}@{}}
\caption{Per-task success rates (\%) on RoboTwin~2.0. Each task is evaluated over 100 episodes in both Clean and Randomized settings; Average is their arithmetic mean.}
\label{tab:robotwin_task_results}\\
\toprule
Task & Clean & Randomized & Average \\
\midrule
\endfirsthead

\multicolumn{4}{c}{\tablename~\thetable{} continued} \\
\toprule
Task & Clean & Randomized & Average \\
\midrule
\endhead

\midrule
\multicolumn{4}{r}{Continued on next page} \\
\endfoot

\bottomrule
\endlastfoot

\texttt{adjust\_bottle} & 100.0 & 100.0 & 100.0 \\
\texttt{beat\_block\_hammer} & 98.0 & 98.0 & 98.0 \\
\texttt{blocks\_ranking\_rgb} & 100.0 & 100.0 & 100.0 \\
\texttt{blocks\_ranking\_size} & 94.0 & 92.0 & 93.0 \\
\texttt{click\_alarmclock} & 92.0 & 83.0 & 87.5 \\
\texttt{click\_bell} & 100.0 & 99.0 & 99.5 \\
\texttt{dump\_bin\_bigbin} & 95.0 & 92.0 & 93.5 \\
\texttt{grab\_roller} & 97.0 & 97.0 & 97.0 \\
\texttt{handover\_block} & 99.0 & 97.0 & 98.0 \\
\texttt{handover\_mic} & 100.0 & 98.0 & 99.0 \\
\texttt{hanging\_mug} & 67.0 & 80.0 & 73.5 \\
\texttt{lift\_pot} & 100.0 & 99.0 & 99.5 \\
\texttt{move\_can\_pot} & 100.0 & 98.0 & 99.0 \\
\texttt{move\_pillbottle\_pad} & 100.0 & 99.0 & 99.5 \\
\texttt{move\_playingcard\_away} & 100.0 & 97.0 & 98.5 \\
\texttt{move\_stapler\_pad} & 90.0 & 83.0 & 86.5 \\
\texttt{open\_laptop} & 98.0 & 100.0 & 99.0 \\
\texttt{open\_microwave} & 79.0 & 87.0 & 83.0 \\
\texttt{pick\_diverse\_bottles} & 97.0 & 93.0 & 95.0 \\
\texttt{pick\_dual\_bottles} & 100.0 & 100.0 & 100.0 \\
\texttt{place\_a2b\_left} & 99.0 & 98.0 & 98.5 \\
\texttt{place\_a2b\_right} & 95.0 & 92.0 & 93.5 \\
\texttt{place\_bread\_basket} & 100.0 & 99.0 & 99.5 \\
\texttt{place\_bread\_skillet} & 98.0 & 95.0 & 96.5 \\
\texttt{place\_burger\_fries} & 97.0 & 99.0 & 98.0 \\
\texttt{place\_can\_basket} & 96.0 & 91.0 & 93.5 \\
\texttt{place\_cans\_plasticbox} & 100.0 & 100.0 & 100.0 \\
\texttt{place\_container\_plate} & 99.0 & 99.0 & 99.0 \\
\texttt{place\_dual\_shoes} & 89.0 & 79.0 & 84.0 \\
\texttt{place\_empty\_cup} & 100.0 & 100.0 & 100.0 \\
\texttt{place\_fan} & 100.0 & 97.0 & 98.5 \\
\texttt{place\_mouse\_pad} & 56.0 & 66.0 & 61.0 \\
\texttt{place\_object\_basket} & 96.0 & 96.0 & 96.0 \\
\texttt{place\_object\_scale} & 99.0 & 92.0 & 95.5 \\
\texttt{place\_object\_stand} & 98.0 & 98.0 & 98.0 \\
\texttt{place\_phone\_stand} & 97.0 & 99.0 & 98.0 \\
\texttt{place\_shoe} & 92.0 & 98.0 & 95.0 \\
\texttt{press\_stapler} & 83.0 & 85.0 & 84.0 \\
\texttt{put\_bottles\_dustbin} & 100.0 & 98.0 & 99.0 \\
\texttt{put\_object\_cabinet} & 94.0 & 90.0 & 92.0 \\
\texttt{rotate\_qrcode} & 88.0 & 92.0 & 90.0 \\
\texttt{scan\_object} & 95.0 & 97.0 & 96.0 \\
\texttt{shake\_bottle} & 100.0 & 100.0 & 100.0 \\
\texttt{shake\_bottle\_horizontally} & 100.0 & 99.0 & 99.5 \\
\texttt{stack\_blocks\_three} & 100.0 & 99.0 & 99.5 \\
\texttt{stack\_blocks\_two} & 100.0 & 100.0 & 100.0 \\
\texttt{stack\_bowls\_three} & 97.0 & 99.0 & 98.0 \\
\texttt{stack\_bowls\_two} & 99.0 & 97.0 & 98.0 \\
\texttt{stamp\_seal} & 98.0 & 99.0 & 98.5 \\
\texttt{turn\_switch} & 65.0 & 71.0 & 68.0 \\
\midrule
\rowcolor{rowblue}
\textbf{Overall} & \textbf{94.7} & \textbf{94.3} & \textbf{94.5} \\
\end{longtable}
\endgroup

\FloatBarrier
\newpage
\subsection{RoboCasa365 Per-Task Results}
\label{app:robocasa_task_results}
Detailed per-task success rates on RoboCasa365 are reported in Table~\ref{tab:robocasa_task_results}.
\begin{center}
\small
\setlength{\tabcolsep}{4pt}
\renewcommand{\arraystretch}{0.9}

\captionof{table}{Per-task success rates (\%) on RoboCasa365.
Each task is evaluated over 50 episodes. Average reports the task-level
mean within each split, and Overall is averaged across all 50 tasks.}
\label{tab:robocasa_task_results}

% \begin{center}
% \small
% \setlength{\tabcolsep}{5pt}
% \renewcommand{\arraystretch}{0.88}
\label{tab:robocasa_task_results}
\begin{tabular}{@{}>{\centering\arraybackslash}p{0.18\linewidth}>{\raggedright\arraybackslash}p{0.58\linewidth}>{\centering\arraybackslash}p{0.14\linewidth}@{}}
\toprule
\textbf{Split} & \textbf{Task} & \textbf{Success} \\
\midrule
\multirow{18}{*}{\textit{Atomic-Seen}} & \texttt{CloseBlenderLid} & 42.0 \\
 & \texttt{CloseFridge} & 96.0 \\
 & \texttt{CloseToasterOvenDoor} & 96.0 \\
 & \texttt{CoffeeSetupMug} & 44.0 \\
 & \texttt{NavigateKitchen} & 84.0 \\
 & \texttt{OpenCabinet} & 92.0 \\
 & \texttt{OpenDrawer} & 86.0 \\
 & \texttt{OpenStandMixerHead} & 88.0 \\
 & \texttt{PickPlaceCounterToCabinet} & 74.0 \\
 & \texttt{PickPlaceCounterToStove} & 76.0 \\
 & \texttt{PickPlaceDrawerToCounter} & 88.0 \\
 & \texttt{PickPlaceSinkToCounter} & 82.0 \\
 & \texttt{PickPlaceToasterToCounter} & 78.0 \\
 & \texttt{SlideDishwasherRack} & 88.0 \\
 & \texttt{TurnOffStove} & 46.0 \\
 & \texttt{TurnOnElectricKettle} & 90.0 \\
 & \texttt{TurnOnMicrowave} & 80.0 \\
 & \texttt{TurnOnSinkFaucet} & 76.0 \\
\cmidrule(l){2-3}
& \textbf{Average (\%) $\uparrow$} & \textbf{78.1} \\
\midrule
\multirow{16}{*}{\textit{Composite-Seen}} & \texttt{DeliverStraw} & 32.0 \\
 & \texttt{GetToastedBread} & 28.0 \\
 & \texttt{KettleBoiling} & 50.0 \\
 & \texttt{LoadDishwasher} & 54.0 \\
 & \texttt{PackIdenticalLunches} & 6.0 \\
 & \texttt{PreSoakPan} & 78.0 \\
 & \texttt{PrepareCoffee} & 22.0 \\
 & \texttt{RinseSinkBasin} & 66.0 \\
 & \texttt{ScrubCuttingBoard} & 90.0 \\
 & \texttt{SearingMeat} & 16.0 \\
 & \texttt{SetUpCuttingStation} & 62.0 \\
 & \texttt{StackBowlsCabinet} & 72.0 \\
 & \texttt{SteamInMicrowave} & 66.0 \\
 & \texttt{StirVegetables} & 24.0 \\
 & \texttt{StoreLeftoversInBowl} & 82.0 \\
 & \texttt{WashLettuce} & 58.0 \\
\cmidrule(l){2-3}
& \textbf{Average (\%) $\uparrow$} & \textbf{50.4} \\
\midrule
\multirow{16}{*}{\textit{Composite-Unseen}} & \texttt{ArrangeBreadBasket} & 10.0 \\
 & \texttt{ArrangeTea} & 2.0 \\
 & \texttt{BreadSelection} & 36.0 \\
 & \texttt{CategorizeCondiments} & 10.0 \\
 & \texttt{CuttingToolSelection} & 46.0 \\
 & \texttt{GarnishPancake} & 20.0 \\
 & \texttt{GatherTableware} & 0.0 \\
 & \texttt{HeatKebabSandwich} & 0.0 \\
 & \texttt{MakeIceLemonade} & 8.0 \\
 & \texttt{PanTransfer} & 0.0 \\
 & \texttt{PortionHotDogs} & 10.0 \\
 & \texttt{RecycleBottlesByType} & 6.0 \\
 & \texttt{SeparateFreezerRack} & 14.0 \\
 & \texttt{WaffleReheat} & 54.0 \\
 & \texttt{WashFruitColander} & 36.0 \\
 & \texttt{WeighIngredients} & 12.0 \\
\cmidrule(l){2-3}
& \textbf{Average (\%) $\uparrow$} & \textbf{16.5} \\
\midrule
\rowcolor{rowblue}
\multicolumn{2}{l}{\textbf{Overall Average (\%) $\uparrow$}} & \textbf{49.5} \\
\bottomrule
\end{tabular}
\end{center}

\FloatBarrier

\section{Attention Masks}
\label{app:attention_details}

\begin{figure}[htbp]
  \centering
  \includegraphics[width=0.98\textwidth]{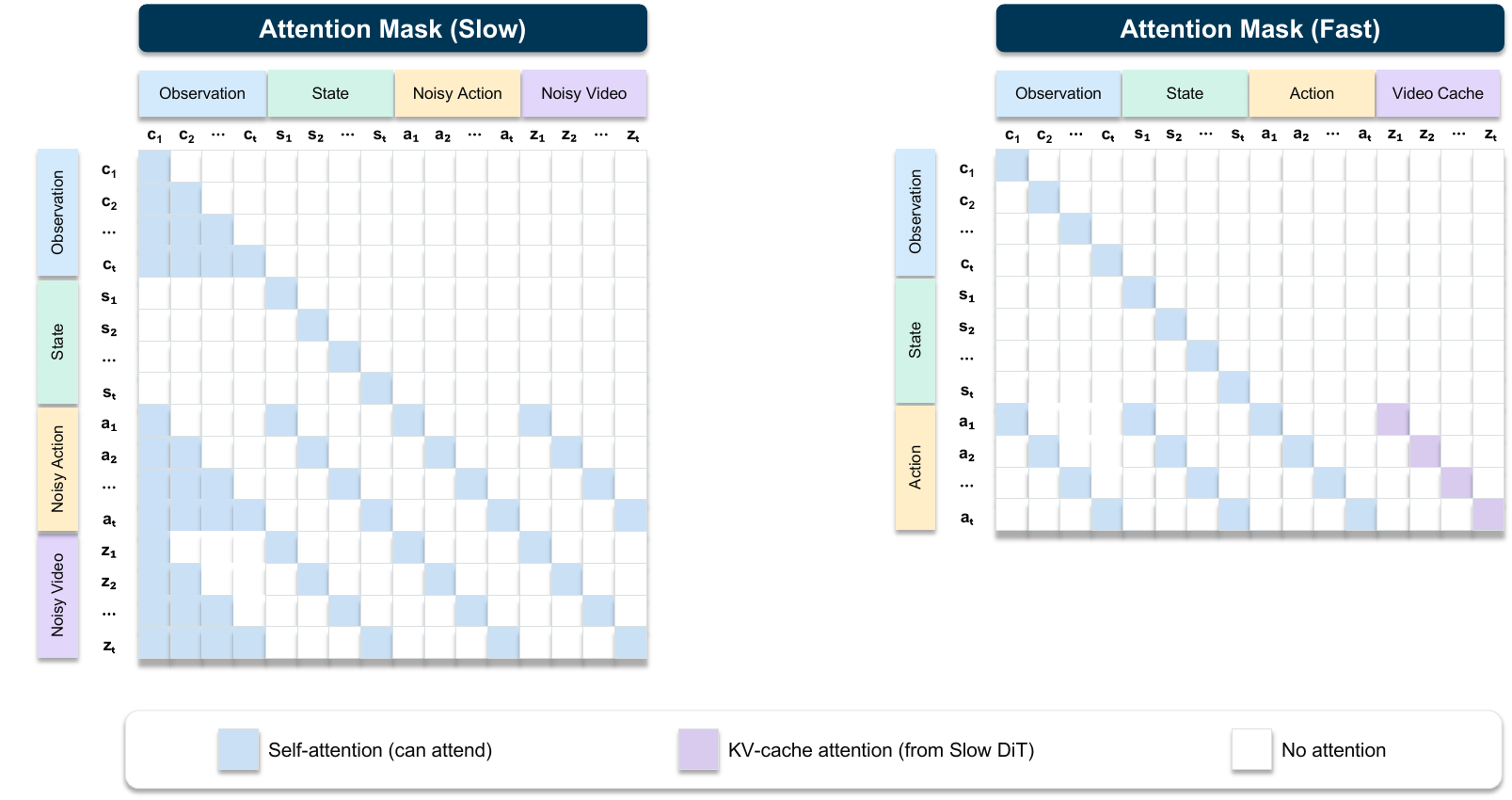}
  \caption{\textbf{Attention masks of the Slow and Fast DiTs.}
  Blue cells denote K/V blocks produced within the same branch, purple cells
  denote video K/V features imported from the Slow DiT, and white cells denote
  blocked dependencies. Rows are queries and columns are keys and values.
  In the Slow mask, $c_i$, $z_i$, $s_i$, and $a_i$ denote the clean
  video context, noised future-video latent, proprioceptive state, and noised
  action chunk at step $i$. In the Fast mask, $c_i$, $s_i$, and $a_i$ denote
  the updated observation, state, and prefix-conditioned action chunk at step
  $i$, while $\mathcal{K}_{\mathrm{slow}}$ is the read-only Slow video K/V
  cache. Language condition $\ell$ is supplied through
  cross-attention and omitted from the self-attention diagram for clarity.}
  \label{fig:dual_attention_mask}
\end{figure}

Figure~\ref{fig:dual_attention_mask} summarizes the self-attention dependencies
of the Slow and Fast DiTs. We use $i\in\{1,\ldots,t\}$ as the temporal index
throughout.

\paragraph{Slow DiT.}
\label{app:slow_attention}

The Slow DiT uses block-causal attention over trajectory segments. Clean
context tokens attend only to the current and preceding clean context. For each
transition, the noised video and action tokens jointly attend to the preceding
clean context, the aligned state, and one another within the same block. They
cannot access target tokens from other blocks or the clean outcome of the
current or future transitions. This mask enables teacher-forced training of
multiple transitions in parallel without leaking future observations. During
video-only pre-training, the action and state streams are omitted; the
view-validity mask additionally blocks padded camera regions.

\paragraph{Fast DiT.}
\label{app:fast_attention}

The Fast DiT predicts actions without a future-video query stream. At step $i$,
each action query attends to the updated observation $c_i$, current state
$s_i$, the complete prefix-conditioned action chunk $a_i$, and the read-only
Slow video K/V cache
$\mathcal{K}_{\mathrm{slow}}$. Only action queries fuse these sources;
observation and state queries remain local. Because Slow is frozen during Fast
training, its cache provides conditioning without receiving gradients.

Attention within the action chunk is non-causal, so every action token can
attend to every other valid token in the same request. 
Different Fast requests remain independent, and
each request jointly denoises one action horizon from the latest observation
and state.

\end{document}